%% file: main.tex
\pdfoutput=1
\PassOptionsToPackage{table}{xcolor}
\documentclass[hidelinks,onefignum,onetabnum]{siamart251216}
\usepackage{graphicx}
\usepackage{amssymb}
\usepackage{booktabs}
\usepackage{multirow}
\usepackage{placeins}
\newcommand{\R}{\mathbb{R}}
\newsiamremark{example}{Example}
\crefname{example}{Example}{Examples}
\headers{Regularization for Flow Matching}{L. Ruthotto and E. Treister}

\title{Explicit Regularization for Flow Matching\thanks{\funding{This work was supported in part by the US--Israeli NSF--BSF award 2411264. Lars Ruthotto's work is supported in part by NSF award DMS-2038118.}}}
\author{Lars Ruthotto\thanks{Departments of Mathematics and Computer Science, Emory University, Atlanta, GA, USA (\email{lruthotto@emory.edu}).}
\and Eran Treister\thanks{Faculty of Computer and Information Science, Ben-Gurion University of the Negev, Beer-Sheva, Israel (\email{erant@bgu.ac.il}).}}
\ifpdf
\hypersetup{
  pdftitle={Explicit Regularization for Flow Matching},
  pdfauthor={Lars Ruthotto and Eran Treister}
}
\fi

\newsavebox{\tablebox}
\newcommand{\fittable}[1]{\sbox{\tablebox}{\footnotesize\input{#1}}%
  \ifdim\wd\tablebox>\textwidth\resizebox{\textwidth}{!}{\usebox{\tablebox}}\else\usebox{\tablebox}\fi}
\begin{document}

\maketitle
\begin{abstract}
Learning the underlying probability distribution from samples is a core problem in statistics and applied mathematics.
Generative models based on diffusion and flow matching learn a velocity field whose flow transports a known distribution to the data distribution.
They produce novel, realistic samples, although the empirical learning problem is ill posed near the data distribution.
Its minimizer transports the samples onto the training data, and small perturbations of the data change the velocity field strongly near the end of the flow.
The observed generalization is often attributed to implicit regularization by the network parameterization, early stopping of the training, and the time integration during sampling.
Explicit regularization has advanced the theory and computation of many inverse problems and has recently been proposed to prevent memorization in generative models.
With a regularizer on the magnitude of the velocity, however, the field approaches the memorizing velocity unless the regularization parameter grows to infinity fast enough.
We show that penalizing the derivatives of order $k$ of the velocity renders the $D$-dimensional empirical problem well posed uniformly in time if $2k > D$ and the training points are in general position.
Our experiments in one and two dimensions discretize the velocity with splines and use direct solvers for the regularized least-squares problems to exclude optimization error.
Refining the discretization in space and time changes the error of the Hessian-regularized flows only marginally, whereas without the regularizer the error tends to that of memorization.
These flows generalize, with errors comparable to or lower than those of well-tuned kernel density estimates.
\end{abstract}



\section{Introduction}\label{sec:intro}

Creating a computational model that generates novel samples or estimates the underlying density from independent and identically distributed samples $x_1, \ldots, x_N \in \R^D$ of a distribution $\pi$ is a core challenge in statistics and applied mathematics.
Generative models produce realistic natural images~\cite{Ho2020,Song2021SDE} and probabilistic weather forecasts~\cite{price2025gencast}.

Continuous-time generative models represent $\pi$ as the push-forward of a tractable latent distribution under the flow of a learned velocity field.
The marginals of the flow solve the continuity equation, with the latent at the initial and the data distribution at the terminal time.
Diffusion and score-based models~\cite{sohlDickstein2015diffusion,Ho2020,Song2021SDE} add noise to the data and learn the score, the gradient of the logarithm of the density of the noisy data, to reverse the process; flow matching regresses onto the velocity directly~\cite{Lipman2023,Liu2023RectifiedFlow}, and stochastic interpolants contain both approaches~\cite{AlbergoBoffi2023Unifying}.
Flow matching and rectified flows have been used to train state-of-the-art image and video generators~\cite{esser2024scaling,polyak2024moviegen}.

\begin{figure}[tbp]
\centering
\includegraphics[width=\textwidth]{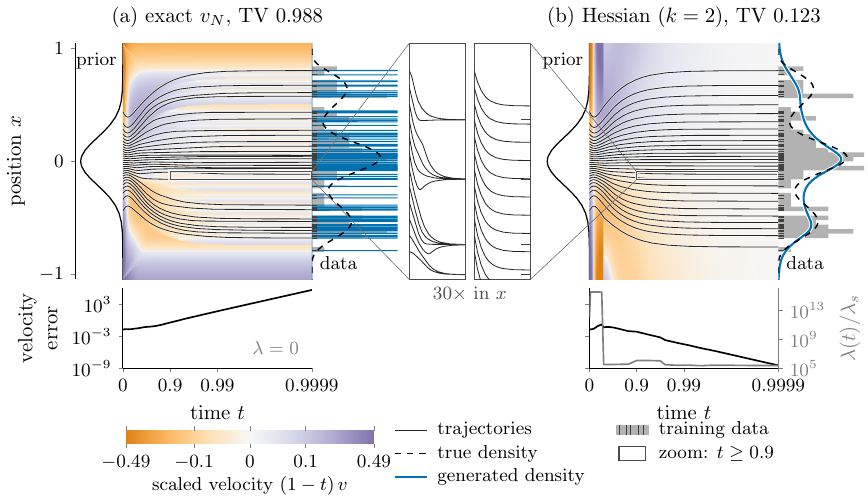}
\caption{Flows of (a) the empirical velocity $v_N$ and (b) the regularized velocity with the Hessian regularizer and hold-out $\lambda(t)$ for the trimodal target with $N = 100$.
Images show the velocity, scaled by $1-t$ and plotted on a time axis that stretches late times, with trajectories; the side panels show the generated density, the true density (dashed) and the training data; the strips show the velocity error, the mean squared distance to the population velocity, and, in (b), the selected $\lambda(t)$ relative to the scale $\lambda_s$ of \cref{sec:solver}. The narrow panels between the columns magnify $t \ge 0.9$ thirtyfold in $x$ near a group of training points: the trajectories of $v_N$ end on the training points, those of the regularized flow do not. Panel titles give the total variation error. Training set with the median hold-out error.}
\label{fig:hero}
\end{figure}

Minimizing the empirical flow matching loss over all velocity fields leads to a flow that memorizes the training data.
Let $\pi_N = \frac{1}{N}\sum_{i=1}^N \delta_{x_i}$ be the empirical measure of the training data, with $\delta_{x_i}$ the Dirac measure at $x_i$.
With $\pi_N$ in place of the data distribution, the empirical velocity $v_N$ has a closed form~\cite[Prop.~1]{bertrand2025closedFormFM}.
As $t \to 1$, the flow of $v_N$ moves almost every latent sample onto a training point and generates $\pi_N$ (\cref{fig:hero}a), so it memorizes the training data~\cite[Thm.~4.6]{gao2024fmMemorize}.
The closed-form score of diffusion models memorizes in the same way; see, e.g.,~\cite{scarvelis2025closedform,baptista2025memorization}.
Hence an over-parameterized network that represents the velocity and is trained long enough approaches the memorizing solution~\cite{bonnaire2025implicit}.
This resembles the semi-convergence of iterative methods for ill-posed problems, whose iterates first approach the exact solution and then move away from it~\cite[Sec.~6.1]{englRegularization1996}.

Image generators reproduce some of their training images, among them photographs of individual people, which compromises the privacy of the training data~\cite[p.~1]{carlini2023extracting}.
Deduplication of the training data reduces these copies~\cite[App.~E.2--E.3]{esser2024scaling}.
That their other samples are not copies of the training data is attributed to implicit regularization, which includes the inductive bias of the network~\cite{kamb2025analytic,kadkhodaie2024generalization}, iterative regularization from early stopping of the training~\cite{bonnaire2025implicit}, and large learning rates~\cite{wu2025largelr}.
Networks trained on the closed-form target generalize as well as networks trained on the stochastic target~\cite{bertrand2025closedFormFM}, so the stochasticity of the target is not the source.

In this paper, we interpret memorization as a symptom of the ill-posedness of the empirical flow matching problem.
A problem is well posed in the sense of Hadamard if a solution exists, is unique, and depends continuously on the data; see, e.g.,~\cite[p.~31]{englRegularization1996}.
For flow matching, the data are the empirical measure $\pi_N$, compared in the Wasserstein-1 distance, and the unknown is the velocity field at each time, compared in a norm on fields.
At each time $t < 1$, the empirical problem has the unique minimizer $v_N$, but, as we show below, its dependence on the training points degenerates as $t \to 1$.
Near a training point $x_i$, the field is approximately $(x_i - x)/(1-t)$, so moving $x_i$ by $\varepsilon$ changes it by $\varepsilon/(1-t)$.
The empirical problem is thus stable at each fixed time but not uniformly in time.

Explicit regularization has recently been proposed to avoid memorization.
Posing score matching with the empirical measure as an inverse problem and adding a Tikhonov regularizer on the norm of the score whose weight grows as the noise vanishes can avoid memorization by ensuring the regularized empirical score stays bounded~\cite[Sec.~5.1]{baptista2025memorization}.
This smallness regularizer, however, shrinks the score toward zero, and with a zero score a variance-exploding sampler leaves the samples at its reference distribution; for flow matching, the minimizer with this regularizer either approaches the memorizing velocity or stays bounded by a constant that is small only for large weights (\cref{ex:tikhonov}).
Also, without spatial coupling, the regularizer does not spread the information of a training point into its neighborhood.
Derivative regularizers on the score appear earlier, in score matching for energy-based models~\cite{kingma2010regularized}.
In one dimension, minimizing the total variation of the derivative of the score, subject to a bound on the score-matching loss, has been proposed to explain the implicit bias of networks, not as an estimator~\cite{chen2026scoresmoothing}.

Other works smooth the solution of the empirical problem without addressing its ill-posedness, for example by Gaussian convolution~\cite{gabriel2025kernelsmoothed}, by averaging the softmax weights over perturbations of the input~\cite{scarvelis2025closedform}, or by tempered weights~\cite{zhou2026smoothing}.
In these works, the smoothed field is not the minimizer of a variational problem, and the amount of smoothing is set by hand~\cite{scarvelis2025closedform} or tuned against the true distribution~\cite{gabriel2025kernelsmoothed}.
None of these works analyzes the stability of the regularized problem with respect to the training data.

Our main contribution is a well-posed variational formulation of empirical flow matching that penalizes the $k$th-order derivatives of the velocity. In particular,
\begin{itemize}
\item our analysis of the $D$-dimensional flow matching problem shows that for $2k > D$ and training points in general position, the problem for the velocity is well posed uniformly in $t$ (\cref{sec:method});
\item our numerical experiments use a spline discretization in $D \le 2$, solved by sparse factorization, with the parameter selected by a hold-out rule. Without the regularizer, the error tends to that of memorization under mesh refinement, and with the Hessian regularizer the error changes by less than $0.01$, so the explicit regularizer, not the discretization, causes the generalization (\cref{sec:numerics,sec:discretization});
\item our experiments on six test densities in one and two dimensions show that the flow with the Hessian regularizer learns the target distribution, with errors far below those of memorization, whereas the error of the flow with the smallness regularizer grows toward that of memorization under refinement; kernel density estimates (KDEs) with established data-driven bandwidths serve as a demanding baseline (\cref{sec:experiments}).
\end{itemize}

\Cref{sec:background} reviews flow matching.
\Cref{sec:method} formulates and analyzes the regularized problem, \cref{sec:numerics} discretizes it in one and two dimensions, \cref{sec:experiments} presents the experiments, and \cref{sec:discussion} summarizes the results and states open problems.
The code that reproduces all experiments, figures and tables is available at \url{https://github.com/EmoryMLIP/2026-RegularizedFlowMatching}.

\section{Flow Matching}\label{sec:background}\label{sec:continuous}
We seek to approximate the data distribution $\pi$ on $\R^D$ as a push-forward of a tractable latent distribution $\gamma$ with density $K$.
We write $K_\sigma(x) = \sigma^{-D} K(x/\sigma)$ for the kernel scaled by $\sigma > 0$.
Flow matching admits any latent distribution that can be sampled.
To simplify our analysis, we use the standard Gaussian conditioned on a box (\cref{sec:formulation}), as it has compact support.
Let the density of $\pi$, which we also denote by $\pi$, be sufficiently regular.

Flow matching applies this push-forward incrementally in artificial time~\cite{Lipman2023,Liu2023RectifiedFlow,Albergo2023Interpolants}.
The sampling process can thus be modeled by the following boundary value problem for the continuity equation,
\begin{equation}\label{eq:continuity}
\partial_t \rho(x,t) + \nabla \cdot \left(\rho(x,t)\, v(x,t)\right) = 0, \qquad \rho(\cdot,0) = \gamma, \quad \rho(\cdot,1) = \pi,
\end{equation}
for a density path $\rho$ and a velocity $v \colon \mathbb{R}^D \times [0,1] \to \mathbb{R}^D$.
The boundary conditions do not determine $\rho$ and $v$ uniquely.
For any sufficiently regular pair $(\rho,v)$ that solves~\cref{eq:continuity}, the ordinary differential equation (ODE) $\dot{x}(t) = v(x(t),t)$ with $x(0) \sim \gamma$ yields $x(1) \sim \pi$, and hence is a valid generator.

Flow matching constructs the probability path $\rho$ by coupling points from the latent and data distributions and then learns the resulting velocity field to satisfy~\cref{eq:continuity}.
Using independent coupling between $z \sim \gamma$ and $y \sim \pi$, the law $\rho(\cdot,t)$ of the interpolant $x_t = (1-t)z + t y$ satisfies~\cref{eq:continuity} with the velocity given by the conditional expectation~\cite{Lipman2023,Liu2023RectifiedFlow,Albergo2023Interpolants}
\begin{equation}\label{eq:velocity}
v_\pi(x,t) = \mathbb{E}\left[y - z \mid x_t = x\right] = \frac{\mathbb{E}\left[y \mid x_t = x\right] - x}{1-t}, \qquad t \in [0,1).
\end{equation}

Replacing $\pi$ by $\pi_N$ in~\cref{eq:velocity} gives the empirical velocity.
Since the density of $x_t$ given $y = x_i$ is $K_{1-t}(\cdot - t x_i)$, we obtain
\begin{equation}\label{eq:empirical}
v_N(x,t) = \frac{m_N(x,t) - x}{1-t}, \qquad m_N(x,t) = \sum_{i=1}^N w_i(x,t)\, x_i,
\end{equation}
with the weights
\begin{equation*}
      w_i(x,t) = \frac{K_{1-t}(x - t x_i)}{\sum_{j=1}^N K_{1-t}(x - t x_j)}.
\end{equation*}
For the standard Gaussian, the weights are the softmax of $-\|x - t x_i\|^2/(2(1-t)^2)$. That is, these exponentials are normalized to sum to one, so the training point whose contracted position $t x_i$ is nearest to $x$ receives the largest weight.
The weighted average $m_N(x,t)$ is the posterior mean of the data point $y \sim \pi_N$ given $x_t = x$, and it is the Nadaraya--Watson estimate~\cite[Sec.~1.5]{tsybakov2009introduction} of $\mathbb{E}[y \mid x_t = x]$ from the pairs $(t x_i, x_i)$ with kernel $K$ and bandwidth $1-t$.

The flow of $v_N$ transports $\gamma$ to $\rho_N(\cdot,t) = \frac{1}{N}\sum_{i=1}^N K_{1-t}(\cdot - t x_i)$.
For points $x_1, \ldots, x_N$, the KDE with kernel $K$ and bandwidth $\sigma$ is $\frac{1}{N}\sum_{i=1}^N K_\sigma(\cdot - x_i)$; see, e.g.,~\cite[Sec.~1.2]{tsybakov2009introduction}.
Hence $x_t$ is distributed as the KDE of the contracted data $t x_1, \ldots, t x_N$ with bandwidth $1-t$, and $x_t/t$ as the KDE of the training data with kernel $K$ and bandwidth $\sigma = (1-t)/t$~\cite[Prop.~1]{smola2026supportconditioned}.
In general, for a latent $z = c + s z_0$ with $z_0 \sim \gamma$, the variable $(x_t - (1-t)c)/t = y + \frac{s(1-t)}{t} z_0$ is distributed as the KDE of the training data with kernel $K$ and bandwidth $s(1-t)/t$.
With a positive terminal noise level, empirical flow matching therefore inherits the Wasserstein-1 rate of the KDE, which matches the minimax lower bound up to logarithmic factors~\cite{kunkel2026minimax}, and an accurate empirical score yields a KDE~\cite{li2024goodscore}.

As $t \to 1$, the weights $w_i$ concentrate on the nearest training point and $\rho_N(\cdot,t)$ converges to $\pi_N$, so the flow of $v_N$ memorizes the training data.
We say that a flow generalizes when its generated law is close to $\pi$ rather than to $\pi_N$, and we measure this by the total variation distance to $\pi$, which equals one for a memorizing flow, since $\pi_N$ is concentrated on the $N$ training points, a set of $\pi$-measure zero (\cref{sec:setup}).

In practice, a network $v_\theta$ is fit by minimizing the empirical flow matching loss
\begin{equation}\label{eq:efm}
\mathcal{L}_N(v) = \frac{1}{2}\int_0^1 \ell_t(v(\cdot,t))\, dt, \quad \ell_t(u) = \frac{1}{N}\sum_{i=1}^N \mathbb{E}_{z \sim \gamma}\left\|u\big((1-t)z + t x_i\big) - (x_i - z)\right\|^2,
\end{equation}
over the parameters $\theta$ of $v = v_\theta$, where $\ell_t$ is the loss at time $t$ of a field $u \colon \R^D \to \R^D$. Stochastic gradient methods minimize the loss, with the integral and the expectation estimated by sampling $t$, $z$, and mini-batches of training points~\cite{Lipman2023,Liu2023RectifiedFlow}. The empirical velocity~\cref{eq:empirical} is the minimizer of~\cref{eq:efm} over all fields; with $\pi$ in place of $\pi_N$, the minimizer is the population velocity $v_\pi$ of~\cref{eq:velocity}. In our experiments, we represent the velocity at each time by splines in space, which keeps the problem quadratic.
A spline space also regularizes, so we refine it until the error no longer depends on it to isolate the effect of the explicit regularization (\cref{sec:discretization}).

\section{Regularized Flow Matching}\label{sec:method}
We formulate the regularized flow matching problem in \cref{sec:formulation} and analyze it in \cref{sec:wellposed}.

\subsection{Variational Formulation}\label{sec:formulation}
We define the regularized velocity field on $\Omega \times [0,1]$, with the box $\Omega = [-L, L]^D$ of diameter $\operatorname{diam}(\Omega) = 2L\sqrt{D}$.
The latent is the standard Gaussian conditioned on $\Omega$, which cuts each coordinate at $L$ standard deviations and differs from $\mathcal{N}(0, I)$ in total variation by $1 - \operatorname{erf}(L/\sqrt{2})^D$.
The formulas of \cref{sec:continuous} hold with $K$ its density.
Since $\Omega$ is convex and contains the latent samples and the training points (assumptions (A1) and (A3) below), it contains $x_t$ at every $t$.

We denote by $H^k(\Omega)^D$ the Sobolev space of fields on the interior of $\Omega$.
For $u \in H^k(\Omega)^D$, the tensor $\nabla^k u = (\nabla^k u_1, \ldots, \nabla^k u_D)$ collects the derivatives of order $k$ of the components.
We write $|u|_{H^k}^2 = \int_\Omega \|\nabla^k u\|^2\, dx$, with the Frobenius norm, and $\|u\|_{H^k}^2 = \sum_{j=0}^k |u|_{H^j}^2$, and define both on a subdomain in the same way.
For a probability density $\rho$ on $\Omega$, we write $\|u\|_{L^2(\rho)}^2 = \int_\Omega \|u\|^2 \rho\, dx$.

Our analysis and numerical experiments rest on the following assumptions:
\begin{enumerate}
  \item[(A1)] The latent distribution $\gamma$ is supported in $\Omega$ and has a bounded density $K$, as does the Gaussian conditioned on $\Omega$.
  \item[(A2)] The samples are pairwise distinct.
  \item[(A3)] The training points and their perturbations in \cref{sec:wellposed} lie in $\Omega$.
\end{enumerate}

The field $v \colon \Omega \times [0,1] \to \R^D$ is obtained by solving the variational problem
\begin{equation}\label{eq:regfm}
\min_v \int_0^1 F_t\big(v(\cdot,t)\big)\, dt, \qquad F_t(u) = \frac{1}{2}\|u - v_N\|_{L^2(\rho_N)}^2 + \frac{\lambda(t)}{2}|u|_{H^k}^2 + \frac{\mu}{2}\|u\|_{L^2(\Omega)}^2,
\end{equation}
where $u \in H^k(\Omega)^D$, $v_N = v_N(\cdot,t)$ and $\rho_N = \rho_N(\cdot,t)$, $\lambda(t)>0$ is a time-dependent regularization parameter, $k \ge 0$ is the order of the regularizer, and $\mu \ge 0$ is the fixed weight of the mass term.
Problem~\cref{eq:regfm} decouples across time, so the regularized field at time $t$ is the minimizer $v_t$ of $F_t$; we do not study its regularity in $t$.

The data term in~\cref{eq:regfm} is the empirical flow matching loss~\cref{eq:efm} up to a constant that does not depend on the field, the empirical counterpart of the equivalence of the conditional and the marginal flow matching losses~\cite[Thm.~2]{Lipman2023}, which holds for the empirical law~\cite[Thm.~2.2]{kunkel2026minimax}.
Indeed, for each $t \in [0,1)$,
\begin{equation}\label{eq:dist-vN}
\ell_t(v) =  \|v - v_N\|^2_{L^2(\rho_N(\cdot,t))}  + \ell_t(v_N).
\end{equation}
To see this, let $y \sim \pi_N$ and $z \sim \gamma$ be independent, so that $\ell_t(v) = \mathbb{E}\|v(x_t) - (y - z)\|^2$, and $x_t \in \Omega$ by (A1) and (A3).
The empirical velocity $v_N(x_t,t) = \mathbb{E}[y - z \mid x_t]$ is the orthogonal projection of $y - z$ onto the random variables $g(x_t)$ with $\mathbb{E}\|g(x_t)\|^2 < \infty$, so $\mathbb{E}[(y - z - v_N(x_t,t)) \cdot (v(x_t) - v_N(x_t,t))] = 0$.
Expanding the square therefore gives~\cref{eq:dist-vN}, whose last term $\ell_t(v_N)$ is the minimum of $\ell_t$.
On $H^k(\Omega)^D$, $\|\cdot\|_{L^2(\rho_N(\cdot,t))}$ is only a seminorm, since by (A1) $\rho_N(\cdot,t)$ vanishes outside the boxes $\Omega_i(t) = t x_i + (1-t)\Omega$ of side length $2L(1-t)$, which support the kernels $K_{1-t}(\cdot - t x_i)$.

The regularizer defines a Tikhonov regularization, in standard form for $k = 0$ and in general form, with $\nabla^k$ as the regularization operator, for $k \ge 1$; see, e.g.,~\cite[Sec.~9.2]{englRegularization1996}.
For $k = 2$ and $D = 2$, it is the thin-plate energy~\cite[Eq.~(2.4.1)]{wahba1990spline}.
We call the regularizers of order $k = 0$, $1$ and $2$ the smallness, gradient and Hessian regularizers, respectively.
The mass term has the form of the smallness regularizer, but its weight is fixed and small, while $\lambda(t)$ is selected from the data.

The null space of $\nabla^k$ consists of the fields whose components are polynomials of degree less than $k$.
It is trivial for the smallness regularizer, consists of the constant fields for the gradient regularizer, and of the affine fields for the Hessian regularizer; only the last contains the limit $x - \mathbb{E}\,z$, which is $x$ for a centered latent, of the population velocity $v_\pi$ as $t \to 1$ at points $x$ where $\pi$ is continuous and positive.
There, the conditional density of $z$ given $x_t = x$ is proportional to $K(z)\,\pi((x - (1-t)z)/t)$, which tends to $K(z)\,\pi(x)$; since $\pi$ is bounded near $x$ by continuity and $K$ has bounded support, dominated convergence gives $v_\pi(x,t) = \mathbb{E}[y - z \mid x_t = x] \to x - \mathbb{E}\,z$.

\subsection{Well-Posedness}\label{sec:wellposed}
We show that the problem has a unique solution at each time and that, for $2k > D$, the solution is stable in the data uniformly in time. We also show which issues arise without sufficient smoothness.
The numerical experiments of \cref{sec:experiments} complement this qualitative analysis and quantify the size of the effects.
For general $k$, we call the training points unisolvent, or in general position, if the only polynomial of degree less than $k$ that vanishes at all of them is zero; see, e.g.,~\cite[Sec.~2.4, p.~34]{wahba1990spline}.
Every set of points is unisolvent for $k \le 1$, and for $k = 2$ the points are unisolvent if they do not all lie on one hyperplane, since a nonzero affine function vanishes only on a hyperplane; this needs at least $D + 1$ points.

We first show that the data term and the regularizer together control the full Sobolev norm at each fixed time, with a constant that may degenerate as $t \to 1$.
\begin{lemma}[Coercivity]\label{lem:coercive}
Let $k \ge 0$ and let (A1) and (A3) hold.
For every $t \in [0,1)$, there exists a constant $c_t>0$ with
\begin{equation}\label{eq:coercive}
\|v\|_{L^2(\rho_N(\cdot,t))}^2 + |v|_{H^k}^2 \ge c_t\, \|v\|_{H^k}^2, \qquad v \in H^k(\Omega)^D.
\end{equation}
\end{lemma}
\begin{proof}
Fix $t \in [0,1)$ and write $Q(v)$ for the left-hand side of~\cref{eq:coercive}.
For $k = 0$, $|v|_{H^0} = \|v\|_{H^0}$, so $Q(v) \ge \|v\|_{H^0}^2$ and $c_t \ge 1$.
Let $k \ge 1$.
Since $Q \ge 0$, the best constant $c_t = \inf_{v \ne 0} Q(v)/\|v\|_{H^k}^2$ is nonnegative, and we show $c_t > 0$ by contradiction.
If $c_t = 0$, there are fields of unit norm with $Q$ arbitrarily small; by compactness, a subsequence converges to a field $p$ with $\|p\|_{H^k} = 1$ and $Q(p) = 0$, and we show that no such field exists.

Suppose $c_t = 0$.
Then, for each $j \in \mathbb{N}$, there is $w_j \in H^k(\Omega)^D$ with $Q(w_j) < \|w_j\|_{H^k}^2/j$, and $w_j \ne 0$, since $w_j = 0$ would give $0 < 0$.
Both sides of this inequality are homogeneous of degree two in $w_j$, so $v_j = w_j/\|w_j\|_{H^k}$ satisfies
\begin{equation*}
\|v_j\|_{H^k} = 1, \qquad Q(v_j) < 1/j.
\end{equation*}

The sequence $(v_j)$ is bounded in $H^k(\Omega)^D$, and $H^k(\Omega)$ embeds compactly into $H^{k-1}(\Omega)$~\cite[Thm.~6.3]{adams2003sobolev}.
Hence a subsequence, which we still denote by $(v_j)$, converges in $H^{k-1}(\Omega)^D$.
The regularizer is part of $Q$, so $|v_j|_{H^k}^2 < 1/j$.
Since $\|\cdot\|_{H^k}^2 = \|\cdot\|_{H^{k-1}}^2 + |\cdot|_{H^k}^2$ and $|a - b|_{H^k}^2 \le 2|a|_{H^k}^2 + 2|b|_{H^k}^2$,
\begin{equation*}
\|v_j - v_l\|_{H^k}^2 \le \|v_j - v_l\|_{H^{k-1}}^2 + 2/j + 2/l,
\end{equation*}
and the first term tends to zero because $(v_j)$ converges in $H^{k-1}(\Omega)^D$, so $(v_j)$ is a Cauchy sequence in $H^k(\Omega)^D$ and converges there to some $p$.
By continuity of the norm and the seminorm, $\|p\|_{H^k} = 1$ and $|p|_{H^k} = 0$.
All weak derivatives of order $k$ of the components of $p$ therefore vanish on the connected set $\Omega$.
A function with zero weak gradient on a connected open set is constant, and applying this $k$ times shows that the components of $p$ are polynomials of degree less than $k$.

It remains to show $p = 0$.
The data term is part of $Q$, so $\|v_j\|_{L^2(\rho_N(\cdot,t))}^2 < 1/j$.
Since $\rho_N(\cdot,t)$ is the average of the kernels $K_{1-t}(\cdot - t x_i)$, each bounded by $(1-t)^{-D}\|K\|_\infty$ by (A1), the constant $M_t = (1-t)^{-D/2}\|K\|_\infty^{1/2}$ satisfies
\begin{equation}\label{eq:Mt}
\|g\|_{L^2(\rho_N(\cdot,t))} \le M_t \|g\|_{L^2(\Omega)} \le M_t \|g\|_{H^k}, \qquad g \in H^k(\Omega)^D.
\end{equation}
The triangle inequality for the seminorm of $L^2(\rho_N(\cdot,t))$ gives
\begin{equation*}
\|p\|_{L^2(\rho_N(\cdot,t))} \le \|v_j\|_{L^2(\rho_N(\cdot,t))} + \|v_j - p\|_{L^2(\rho_N(\cdot,t))} < j^{-1/2} + M_t \|v_j - p\|_{H^k},
\end{equation*}
and both terms tend to zero, so $\|p\|_{L^2(\rho_N(\cdot,t))} = 0$.
Hence $p$ vanishes almost everywhere on the set where $\rho_N(\cdot,t) > 0$, which has positive Lebesgue measure, since $\rho_N(\cdot,t)$ integrates to one.
A nonzero polynomial vanishes only on a set of measure zero, so $p = 0$, which contradicts $\|p\|_{H^k} = 1$.
\end{proof}

The result above is the complementation condition of general-form Tikhonov regularization~\cite[Sec.~8.1, Eq.~(8.3)]{englRegularization1996}, with $T$ denoting the inclusion $Tv = v$ of $H^k(\Omega)^D$ into $L^2(\rho_N(\cdot,t))$, which is bounded by~\cref{eq:Mt}, and $\nabla^k$ as the regularization operator.
The largest constant $c_t$ of \cref{lem:coercive} may tend to zero as $t \to 1$, where $\rho_N(\cdot,t)$ concentrates at the training points.

For $2k > D$ and $0 < \nu \le 1$ with $\nu < k - D/2$, $H^k(\Omega)$ embeds into the H\"older space $C^{0,\nu}(\overline\Omega)$, since the box $\Omega$ satisfies the strong local Lipschitz condition~\cite[Thm.~4.12]{adams2003sobolev}.
We identify each $g \in H^k(\Omega)^D$ with its continuous representative, so that point values are defined, and apply the embedding componentwise.
This gives a constant $c_H$, depending only on $k$, $D$, the box size $L$ and $\nu$, with
\begin{equation}\label{eq:holder}
\|g(a)\| \le c_H\|g\|_{H^k}, \quad \|g(a) - g(b)\| \le c_H\|g\|_{H^k}\,\|a - b\|^\nu, \quad a, b \in \Omega,\ g \in H^k(\Omega)^D,
\end{equation}
since the squared $H^k$ norm of a field is the sum of the squared $H^k$ norms of its components.
In particular, point evaluation is bounded on $H^k(\Omega)$, the classical reason why thin-plate and smoothing splines require $2k > D$~\cite[Sec.~2.4, p.~31]{wahba1990spline}.

For perturbed training data $x_1', \ldots, x_N' \in \Omega$, we write $\pi_N'$ for their empirical measure and $\rho_N'(\cdot,t)$ for the density of~\cref{sec:continuous} formed from them, with $\rho_N'(\cdot,1) = \pi_N'$, so that $\|v\|_{L^2(\rho_N'(\cdot,1))}^2 = \frac{1}{N}\sum_{i=1}^N \|v(x_i')\|^2$.
For $2k > D$, the next lemma bounds the constant of \cref{lem:coercive} below by one constant for all times, including $t = 1$, and for slightly moved training points.

\begin{lemma}[Uniform coercivity]\label{lem:uniform}
Let $2k > D$, (A1) and (A3) hold, and let the training points $x_1, \ldots, x_N$ be unisolvent.
Then there are $c_\star > 0$ and $\delta_0 > 0$ such that
\begin{equation}\label{eq:coercive-uniform}
\|v\|_{L^2(\rho_N'(\cdot,t))}^2 + |v|_{H^k}^2 \ge c_\star\, \|v\|_{H^k}^2, \qquad v \in H^k(\Omega)^D,
\end{equation}
for every $t \in [0,1]$ and all points $x_1', \ldots, x_N' \in \Omega$ with $\|x_i' - x_i\| \le \delta_0$ for $i = 1, \ldots, N$.
\end{lemma}
\begin{proof}
We show by contradiction that one pair $(c_\star, \delta_0)$ serves all $t \in [0,1]$ and all perturbed points.
Since $2k > D$ and $D \ge 1$, $k \ge 1$.
Write $Q_{t,X'}(v)$ for the left-hand side of~\cref{eq:coercive-uniform}, with $\rho_N'$ formed from $X' = (x_1', \ldots, x_N')$.

Suppose that no such pair exists.
Then, for each $j \in \mathbb{N}$, the pair $c_\star = \delta_0 = 1/j$ fails: \cref{eq:coercive-uniform} with $c_\star = 1/j$ is violated at some time $t_j \in [0,1]$, for some points $X^{(j)} = (x_1^{(j)}, \ldots, x_N^{(j)})$ in $\Omega$ with $\max_i \|x_i^{(j)} - x_i\| \le 1/j$, and by some $w_j \in H^k(\Omega)^D$, that is, $Q_{t_j,X^{(j)}}(w_j) < \|w_j\|_{H^k}^2/j$.
The density $\rho_j = \rho_N'(\cdot,t_j)$ in $Q_{t_j,X^{(j)}}$ is formed from the points $X^{(j)}$, not from the training points.
As in the proof of \cref{lem:coercive}, $w_j \ne 0$, and $v_j = w_j/\|w_j\|_{H^k}$ satisfies $\|v_j\|_{H^k} = 1$ and $Q_{t_j,X^{(j)}}(v_j) < 1/j$ by homogeneity of degree two.
Since $[0,1]$ is compact, we may pass to a subsequence with $t_j \to t^* \in [0,1]$; by construction, $x_i^{(j)} \to x_i$ for every $i$.

The regularizer is part of $Q_{t_j,X^{(j)}}$, so $|v_j|_{H^k}^2 < 1/j$.
The compactness argument of the proof of \cref{lem:coercive} uses only this bound and $\|v_j\|_{H^k} = 1$, so a further subsequence converges in $H^k(\Omega)^D$ to a field $p$ whose components are polynomials of degree less than $k$, with $\|p\|_{H^k} = 1$.

We show that the data term forces $\|p\|_{L^2(\rho_j)} \to 0$, although the measure $\rho_j$ changes with $j$.
Each $\rho_j$ is a probability measure on $\Omega$: for $t_j < 1$ a probability density, since the points $X^{(j)}$ lie in $\Omega$ by (A3), and for $t_j = 1$ the empirical measure of $X^{(j)}$.
Hence~\cref{eq:holder} gives $\|g\|_{L^2(\rho_j)} \le c_H \|g\|_{H^k}$ for every $j$, with a constant that depends neither on $t_j$ nor on $X^{(j)}$.
The data term is part of $Q_{t_j,X^{(j)}}$, so $\|v_j\|_{L^2(\rho_j)}^2 < 1/j$, and the triangle inequality for the seminorm of $L^2(\rho_j)$ gives
\begin{equation*}
\|p\|_{L^2(\rho_j)} \le \|v_j\|_{L^2(\rho_j)} + \|v_j - p\|_{L^2(\rho_j)} < j^{-1/2} + c_H \|v_j - p\|_{H^k} \to 0.
\end{equation*}
The bound~\cref{eq:Mt} cannot replace the embedding here, since it degenerates as $t_j \to 1$.

The kernel $K_{1-t}(\cdot - t x_i')$ is the density of $(1-t)z + t x_i'$ for $z \sim \gamma$, and for $t = 1$ this point is $x_i'$, so for every $j$
\begin{equation*}
\|p\|_{L^2(\rho_j)}^2 = \frac{1}{N}\sum_{i=1}^N \mathbb{E}\,\big\|p\big((1-t_j)z + t_j x_i^{(j)}\big)\big\|^2.
\end{equation*}
For every $z \in \Omega$, the argument of $p$ converges to $(1-t^*)z + t^* x_i$, and $p$ is continuous, so the integrand converges.
It is bounded by $\max_\Omega \|p\|^2$, which is finite because $p$ is a polynomial and $\Omega$ is compact, so dominated convergence gives
\begin{equation*}
0 = \lim_{j \to \infty} \|p\|_{L^2(\rho_j)}^2 = \frac{1}{N}\sum_{i=1}^N \mathbb{E}\,\big\|p\big((1-t^*)z + t^* x_i\big)\big\|^2 = \|p\|_{L^2(\rho_N(\cdot,t^*))}^2,
\end{equation*}
where $\rho_N(\cdot,t^*)$ is formed from the training points.
The perturbed points have disappeared from the limit, and only the training points need to be unisolvent.

If $t^* < 1$, $p$ vanishes almost everywhere on the set where $\rho_N(\cdot,t^*) > 0$, which has positive measure, so $p = 0$, as in the proof of \cref{lem:coercive}.
If $t^* = 1$, the limit is $\frac{1}{N}\sum_{i=1}^N \|p(x_i)\|^2 = 0$, so each component of $p$ is a polynomial of degree less than $k$ that vanishes at every training point, and unisolvency gives $p = 0$.
In both cases $p = 0$, which contradicts $\|p\|_{H^k} = 1$.
\end{proof}

At each time, the problem has exactly one solution, bounded in $H^k$, and near-minimizers lie close to it.
\begin{theorem}[Existence and uniqueness]\label{thm:existence}
Let $k \ge 0$ and $\mu \ge 0$, let (A1) and (A3) hold, and let $\lambda(t) \ge \lambda_0 > 0$ for every $t \in [0,1)$.
For every $t \in [0,1)$, $F_t$ in \cref{eq:regfm} has a unique minimizer $v_t \in H^k(\Omega)^D$, and
\begin{equation*}
\|v_t\|_{H^k}^2 \le \frac{\operatorname{diam}(\Omega)^2}{\kappa_t}, \qquad \|w - v_t\|_{H^k}^2 \le \frac{2}{\kappa_t}\big(F_t(w) - F_t(v_t)\big), \quad w \in H^k(\Omega)^D,
\end{equation*}
with $\kappa_t = \min(1, \lambda_0)\, c_t$ and $c_t$ of \cref{lem:coercive}.
\end{theorem}
\begin{proof}
Fix $t \in [0,1)$.
We verify, on $V = H^k(\Omega)^D$, the assumptions of the existence theorem for elliptic quadratic minimization problems~\cite[Thm.~1.1.1]{ciarlet2002finite}.
Expanding $\|v - v_N\|^2 = \|v\|^2 - 2\, v \cdot v_N + \|v_N\|^2$ gives
\begin{equation*}
F_t(v) = \frac{1}{2} a_t(v,v) - f_t(v) + \frac{1}{2}\|v_N\|_{L^2(\rho_N)}^2,
\end{equation*}
with the symmetric bilinear form and the linear functional
\begin{align*}
a_t(v,w) &= \int_\Omega v \cdot w\, \rho_N\, dx + \lambda(t) \int_\Omega \nabla^k v : \nabla^k w\, dx + \mu \int_\Omega v \cdot w\, dx,\\
f_t(w) &= \int_\Omega v_N \cdot w\, \rho_N\, dx.
\end{align*}
The constant $\frac{1}{2}\|v_N\|_{L^2(\rho_N)}^2$ does not change the minimizer.

The assumptions (A1) and (A3) place $y$ and $z$ in $\Omega$, so $\|v_N(x,t)\| = \|\mathbb{E}[y - z \mid x_t = x]\| \le \operatorname{diam}(\Omega)$ wherever $\rho_N(x,t) > 0$.
Since $\rho_N(\cdot,t)$ is a probability density, $\|v_N\|_{L^2(\rho_N)} \le \operatorname{diam}(\Omega)$.
The Cauchy--Schwarz inequality in $L^2(\rho_N)$ and in $L^2(\Omega)$, together with~\cref{eq:Mt}, gives
\begin{equation*}
|a_t(v,w)| \le \big(M_t^2 + \lambda(t) + \mu\big)\|v\|_{H^k}\|w\|_{H^k}, \qquad |f_t(w)| \le \operatorname{diam}(\Omega)\, M_t\, \|w\|_{H^k},
\end{equation*}
so $a_t$ and $f_t$ are bounded.
Since $\lambda(t) \ge \lambda_0$ and $\mu \ge 0$, \cref{lem:coercive} gives
\begin{equation*}
a_t(u,u) \ge \min(1,\lambda_0)\big(\|u\|_{L^2(\rho_N)}^2 + |u|_{H^k}^2\big) \ge \kappa_t \|u\|_{H^k}^2.
\end{equation*}
The assumptions of the existence theorem are thus satisfied, and $F_t$ has a unique minimizer $v_t$.

For $h \in V$ and $\xi \in \R$, bilinearity and symmetry of $a_t$ give
\begin{equation*}
F_t(v_t + \xi h) = F_t(v_t) + \xi\big(a_t(v_t,h) - f_t(h)\big) + \frac{\xi^2}{2}\, a_t(h,h).
\end{equation*}
The right-hand side is minimal at $\xi = 0$, so the coefficient of $\xi$ vanishes, and $a_t(v_t,h) = f_t(h)$ for every $h \in V$.
With $\xi = 1$ and $h = w - v_t$, $F_t(w) - F_t(v_t) = \frac{1}{2}a_t(w - v_t, w - v_t) \ge \frac{\kappa_t}{2}\|w - v_t\|_{H^k}^2$, which is the second bound.
Every term of $F_t$ is nonnegative, so $F_t(v_t) \ge 0$, and $F_t(0) = \frac{1}{2}\|v_N\|_{L^2(\rho_N)}^2 \le \frac{1}{2}\operatorname{diam}(\Omega)^2$.
The second bound with $w = 0$ then gives $\|v_t\|_{H^k}^2 \le \frac{2}{\kappa_t}\big(F_t(0) - F_t(v_t)\big) \le \operatorname{diam}(\Omega)^2/\kappa_t$, which is the first.
\end{proof}
For $2k > D$ and unisolvent training points, the bounds in \cref{thm:existence} hold uniformly in time, which enables our stability and late-time results. \Cref{lem:uniform} gives $c_t \ge c_\star$ for every $t$, so
\begin{equation}\label{eq:R}
\kappa_t \ge \kappa = \min(1, \lambda_0)\, c_\star, \qquad \|v_t\|_{H^k} \le R = \operatorname{diam}(\Omega)/\sqrt{\kappa}, \qquad t \in [0,1).
\end{equation}

For $\mu > 0$, we may instead drop the data term in the proof of \cref{thm:existence} and obtain a bound for every set of training points.
The interpolation inequality bounds each seminorm $|u|_{H^j}$ with $0 < j < k$ by $\|u\|_{L^2(\Omega)}$ and $|u|_{H^k}$~\cite[Thm.~5.2]{adams2003sobolev}.
Summing over $j$ gives $\|u\|_{H^k}^2 \le c_I\big(\|u\|_{L^2(\Omega)}^2 + |u|_{H^k}^2\big)$ with $c_I$ depending only on $k$, $D$ and $L$, so $a_t(u,u) \ge \min(\lambda_0, \mu)\, \|u\|_{H^k}^2 / c_I$, and~\cref{eq:R} holds for every set of training points with $\kappa = \min(\lambda_0, \mu)/c_I$.
The results below that assume unisolvent points therefore also hold, with this $\kappa$, for every set of training points if $\mu > 0$.

As $t \to 1$, the support of $\rho_N(\cdot,t)$ splits into disjoint boxes around the training points, and on each box the empirical velocity is affine with slope $-1/(1-t)$.
For $y = x_i$, the interpolant $x_t = t x_i + (1-t)z$ lies in $\Omega_i(t)$, and each point of $\Omega_i(t)$ is of this form for exactly one latent sample $z \in \Omega$.
We set $\tilde v_i(z) = v(t x_i + (1-t)z)$, which depends on $t$.
This change of variables maps the shrinking box onto the fixed box $\Omega$ and moves the dependence on $t$ from the domain into the coefficients.
The scaling of the data term and the regularizer on the boxes indicates the limiting behavior.
A field that matches the target on each box and vanishes between the boxes has a zero data term, and its regularizer is that of a fixed field on $\Omega$ scaled by $\lambda(t)(1-t)^{D-2k}$.
The same exponent decides whether point evaluation is bounded: a bump of unit height and width $\varepsilon$ has a squared seminorm of order $\varepsilon^{D-2k}$.
For bounded $\lambda$, this factor tends to zero for $2k < D$, so following the targets costs nothing in the limit, and it grows without bound for $2k > D$.
\Cref{lem:boxes} makes this scaling exact, and the results after it prove both cases. The lemma also states when the dilated boxes $t x_i + 2(1-t)\Omega$, with double side length of $\Omega_i$, are pairwise disjoint.

\begin{lemma}[Late-time localization]\label{lem:boxes}
Let (A1)--(A3) hold and let $t_N = 4L/(4L + \min_{i \ne j}\|x_i - x_j\|_\infty)$.
For every $t \in (t_N, 1)$, the following hold.
\begin{enumerate}
\item[(a)] The boxes $\Omega_i(t)$ lie in $\Omega$, the dilated boxes $t x_i + 2(1-t)\Omega$ are pairwise disjoint, and $\rho_N(\cdot,t)$ vanishes outside $\bigcup_i \Omega_i(t)$.
\item[(b)] For $x = t x_i + (1-t)z \in \Omega_i(t)$, $\rho_N(x,t) = K(z)/(N(1-t)^D)$, and $v_N(x,t) = x_i - z$ wherever $\rho_N(x,t) > 0$, so $\ell_t(v_N) = 0$.
\item[(c)] For $v \in H^k(\Omega)^D$,
\begin{gather*}
\int_{\Omega_i(t)} \|v - v_N\|^2 \rho_N\, dx = \frac{1}{N}\int_\Omega \|\tilde v_i(z) - (x_i - z)\|^2 K(z)\, dz, \\
|v|_{H^k(\Omega_i(t))}^2 = (1-t)^{D-2k}\, |\tilde v_i|_{H^k(\Omega)}^2, \qquad
\|v\|_{L^2(\Omega_i(t))}^2 = (1-t)^D\, \|\tilde v_i\|_{L^2(\Omega)}^2.
\end{gather*}
\end{enumerate}
\end{lemma}
\begin{proof}
By (A1), $K_{1-t}(\cdot - t x_i)$ is supported in $\Omega_i(t)$, which lies in $\Omega$ by (A3) and the convexity of $\Omega$, so $\rho_N(\cdot,t)$ vanishes outside the union of these boxes.
The dilated boxes have the centers $t x_i$ and the half-width $2L(1-t)$, so they are disjoint if $4L(1-t) < t \min_{i \ne j}\|x_i - x_j\|_\infty$, that is, if $t > t_N$, and $t_N < 1$ by (A2).
On $\Omega_i(t)$ only the $i$th kernel is nonzero, so the weight $w_i$ in~\cref{eq:empirical} equals one, $\rho_N(x,t) = K(z)/(N(1-t)^D)$, and $v_N(x,t) = (x_i - x)/(1-t) = x_i - z$.
This is $y - z$ for $y = x_i$, so the integrand of $\ell_t(v_N)$ vanishes, which proves (b).
Part (c) follows from $dx = (1-t)^D dz$ and the chain rule, by which each derivative of order $k$ of $\tilde v_i$ is $(1-t)^k$ times the corresponding derivative of $v$.
\end{proof}

For $2k < D$, matching the target on each box does not result in a large regularizer, so the regularized field approaches the memorizing field.
\begin{proposition}[Localized fields]\label{prop:localized}
Let $k \ge 0$, let (A1)--(A3) hold, let $\lambda(t) \ge \lambda_0 > 0$ for every $t\in[0,1)$, and let $v_t$, $t \in [0,1)$, be the minimizers of $F_t$ given by \cref{thm:existence}.
For $t \in (t_N, 1)$,
\begin{equation*}
\int_\Omega \|v_t - v_N\|^2 \rho_N\, dx \le C N \big(\lambda(t)(1-t)^{D-2k} + \mu (1-t)^D\big),
\end{equation*}
where $C$ depends only on $k$, $D$ and $L$.
In particular, for fixed $N$, if $\lambda(t)(1-t)^{D-2k} \to 0$ as $t \to 1$, as for a bounded $\lambda$ when $2k < D$, then $v_t$ approaches $v_N$ in $L^2(\rho_N(\cdot,t))$.
\end{proposition}
\begin{proof}
Fix a cutoff function $\varphi \in C_c^\infty(\R^D)$, depending only on $D$ and $L$, that equals one on $\Omega$ and vanishes outside $2\Omega$, let $\psi_i(z) = \varphi(z)(x_i - z)$, and let $u$ be the restriction to $\Omega$ of $\sum_i \psi_i\big((x - t x_i)/(1-t)\big)$.
The $i$th summand vanishes outside the $i$th dilated box, and these boxes are disjoint by \cref{lem:boxes}(a), so $u = v_N$ on the support of $\rho_N(\cdot,t)$ by \cref{lem:boxes}(b), and the data term of $u$ vanishes.
The summands have disjoint supports, so their seminorms add.
A dilated box extends beyond $\Omega$ if $x_i$ lies within $L(1-t)/t$ of the boundary, and restricting to $\Omega$ does not increase the seminorms; with the change of variables of \cref{lem:boxes}(c), applied on the dilated boxes, this gives
\begin{equation*}
F_t(u) \le \frac{1}{2} \sum_{i=1}^N \Big(\lambda(t)(1-t)^{D-2k}\, |\psi_i|_{H^k(2\Omega)}^2 + \mu (1-t)^D\, \|\psi_i\|_{L^2(2\Omega)}^2\Big).
\end{equation*}
The data term is at most $2F_t$, and $v_t$ minimizes $F_t$, so
\begin{equation*}
\int_\Omega \|v_t - v_N\|^2 \rho_N\, dx \le 2F_t(v_t) \le 2F_t(u) \le C N \big(\lambda(t)(1-t)^{D-2k} + \mu (1-t)^D\big),
\end{equation*}
where $C$ bounds the norms of $\psi_i$ on $2\Omega$ uniformly in $i$, since $\psi_i$ depends linearly on $x_i$ and $\Omega$ is bounded.
\end{proof}

For the borderline case $2k = D$, the factor equals $\lambda(t)$, which does not vanish for $\lambda(t) \ge \lambda_0$, so the bound does not decide this case.
For $k \ge 2$, the affine targets lie in the null space of $|\cdot|_{H^k}$, so the seminorm of $u$ comes only from the transition layer between the boxes.
For $N = 1$, for instance, the affine field $(x_1 - x)/(1-t)$ equals $v_N$ on the support of $\rho_N(\cdot,t)$ and has zero seminorm, so for $\mu = 0$ it minimizes $F_t$ and its flow memorizes; one point is not unisolvent for $k \ge 2$.

For $2k > D$ and unisolvent training points, in contrast, \cref{eq:R,eq:holder} show that for $t \in (t_N, 1)$ the minimizer varies by at most $c_H R\,(2L\sqrt{D}(1-t))^\nu$ on each box $\Omega_i(t)$, while the target varies by $2L\sqrt{D}$.

The remaining results show that for $2k>D$, the field depends stably on the data (\cref{thm:stability}) and converges as $t\to1$ to the minimizer of a limit problem (\cref{thm:latetime}).
Both proofs rely on the following lemma, which shows that minimizers of two functionals that grow quadratically away from them are nearby if the functionals are nearby.
\begin{lemma}[Comparison of minimizers]\label{lem:compare}
Let $G$ and $G'$ be real-valued functionals on $H^k(\Omega)^D$ with minimizers $u$ and $u'$, and let $\beta > 0$ be such that $G(g) - G(u) \ge \beta \|g - u\|_{H^k}^2$ and $G'(g) - G'(u') \ge \beta \|g - u'\|_{H^k}^2$ for every $g \in H^k(\Omega)^D$.
Then
\begin{equation*}
2\beta \|u' - u\|_{H^k}^2 \le \big(G(u') - G'(u')\big) - \big(G(u) - G'(u)\big).
\end{equation*}
\end{lemma}
\begin{proof}
The first inequality with $g = u'$ and the second with $g = u$ give $G(u') - G(u) \ge \beta\|u' - u\|_{H^k}^2$ and $G'(u) - G'(u') \ge \beta\|u' - u\|_{H^k}^2$.
Adding them and grouping the four terms by their argument gives the claim.
\end{proof}

For $2k > D$, moving the training points slightly moves the regularized field slightly, with a constant that does not depend on time.
\begin{theorem}[Stability in the data]\label{thm:stability}
Let $2k > D$ and $0 < \nu \le 1$ with $\nu < k - D/2$, let (A1) and (A3) hold, and let $\lambda(t) \ge \lambda_0 > 0$ for every $t \in [0,1)$.
Let $\pi_N'$ be a second empirical measure that satisfies (A3), and let $v_t$ and $v_t'$, $t\in[0,1)$,  be the minimizers given by \cref{thm:existence} for $\pi_N$ and $\pi_N'$, for the same $\lambda$ and $\mu$.
Let the points of both measures be unisolvent, and let $\kappa$ be the smaller of the constants of~\cref{eq:R} for $\pi_N$ and $\pi_N'$.
Let $W_1(\pi_N,\pi_N') = \min \mathbb{E}\,\|y - y'\|$ over the couplings of $y \sim \pi_N$ and $y' \sim \pi_N'$.
Then the minimizers depend H\"older continuously on the data, uniformly in $t$,
\begin{equation*}
\sup_{t \in [0,1)} \|v_t - v_t'\|_{H^k} \le C\, W_1(\pi_N, \pi_N')^\nu,
\end{equation*}
where $C$ depends only on $k$, $D$, $L$, $\nu$ and $\kappa$.
\end{theorem}
\begin{proof}
Fix $t \in [0,1)$ and let $h = v_t' - v_t$.
By~\cref{eq:dist-vN}, $F_t(v) = \frac{1}{2}\ell_t(v) + \frac{\lambda(t)}{2}|v|_{H^k}^2 + \frac{\mu}{2}\|v\|_{L^2(\Omega)}^2 - \frac{1}{2}\ell_t(v_N)$, and likewise $F_t'$ with $\ell_t'$ and $v_N'$ for $\pi_N'$.
The second bound of \cref{thm:existence} holds for $F_t$ and for $F_t'$ with $\kappa_t \ge \kappa$, so \cref{lem:compare} applies with $G = F_t$, $G' = F_t'$ and $\beta = \kappa/2$.
The regularizers of $F_t$ and $F_t'$ are equal and cancel in $F_t - F_t'$, and the constants $\ell_t(v_N)$ and $\ell_t'(v_N')$ cancel between the two differences, so
\begin{equation*}
4\beta \|h\|_{H^k}^2 \le \big(\ell_t(v_t') - \ell_t(v_t)\big) - \big(\ell_t'(v_t') - \ell_t'(v_t)\big).
\end{equation*}

Let $(y, y')$ be an optimal coupling, which exists because both measures are finitely supported, let $z \sim \gamma$ be independent of it, and let $x_t = (1-t)z + ty$ and $x_t' = (1-t)z + ty'$.
Then $\ell_t(g) = \mathbb{E}\|g(x_t) - (y - z)\|^2$ and $\ell_t'(g) = \mathbb{E}\|g(x_t') - (y' - z)\|^2$.
With $S = v_t + v_t'$, $a = y - z$, $a' = y' - z$ and the identity $\|u_1 - u_2\|^2 - \|u_3 - u_2\|^2 = (u_1 - u_3)\cdot(u_1 + u_3 - 2u_2)$, the right-hand side equals $\mathbb{E}\big[h(x_t)\cdot(S(x_t) - 2a) - h(x_t')\cdot(S(x_t') - 2a')\big]$, and adding and subtracting $h(x_t')\cdot(S(x_t) - 2a)$ gives
\begin{equation*}
\mathbb{E}\Big[\big(h(x_t) - h(x_t')\big) \cdot \big(S(x_t) - 2(y - z)\big) + h(x_t') \cdot \big(S(x_t) - S(x_t') - 2(y - y')\big)\Big].
\end{equation*}

We bound the two terms with~\cref{eq:holder}, which gives $\|h(x_t) - h(x_t')\| \le c_H\|h\|_{H^k}\|x_t - x_t'\|^\nu$ and $\|h(x_t')\| \le c_H\|h\|_{H^k}$.
Since~\cref{eq:R} bounds $v_t$ and $v_t'$ by $R$, it also gives $\|S\| \le 2c_H R$ and $\|S(x_t) - S(x_t')\| \le 2c_H R\, \|x_t - x_t'\|^\nu$.
Moreover, $\|x_t - x_t'\| = t\|y - y'\| \le \|y - y'\|$, $\|y - z\| \le \operatorname{diam}(\Omega)$ and $\|y - y'\| \le \operatorname{diam}(\Omega)^{1-\nu}\|y - y'\|^\nu$.
The right-hand side is therefore at most
\begin{equation*}
c_H \|h\|_{H^k}\big(4 c_H R + 2\operatorname{diam}(\Omega) + 2\operatorname{diam}(\Omega)^{1-\nu}\big)\, \mathbb{E}\,\|y - y'\|^\nu.
\end{equation*}
Jensen's inequality for the concave function $s \mapsto s^\nu$ and the optimality of the coupling give $\mathbb{E}\,\|y - y'\|^\nu \le \big(\mathbb{E}\,\|y - y'\|\big)^\nu = W_1(\pi_N,\pi_N')^\nu$.
Dividing by $4\beta\|h\|_{H^k}$ gives the claim with a constant that does not depend on $t$.
\end{proof}
The proof never evaluates $v_N$, which changes by order $1/(1-t)$ when a training point moves (\cref{sec:intro}); it compares the losses $\ell_t$ and $\ell_t'$ through the coupling, so this factor does not enter.
In \cref{thm:stability}, $\pi_N'$ may have a different number of points.
For unisolvent points, \cref{lem:uniform} gives one $\kappa$ for all point sets whose points lie within $\delta_0$ of the corresponding training points, so moving one training point by $\delta \le \delta_0$ changes $W_1$ by at most $\delta/N$ and hence the field by at most $C(\delta/N)^\nu$.

For $2k > D$, the regularized field converges as $t \to 1$ to the minimizer of a limit problem at the training points, not to the memorizing field.
\begin{theorem}[Late-time limit]\label{thm:latetime}
Let $2k > D$ and $0 < \nu \le 1$ with $\nu < k - D/2$, let (A1) and (A3) hold, let $\lambda(t) \ge \lambda_0 > 0$ for every $t \in [0,1)$ with $\lambda(t) \to \bar\lambda$ as $t \to 1$, and let $v_t$ be the minimizers given by \cref{thm:existence}.
Let the training points be unisolvent, and let $\kappa$ and $R$ be the constants of~\cref{eq:R}.
Then $\bar F(v) = \frac{1}{2}\ell_1(v) + \frac{\bar\lambda}{2}|v|_{H^k}^2 + \frac{\mu}{2}\|v\|_{L^2(\Omega)}^2$, with $\ell_1$ of~\cref{eq:efm}, has a unique minimizer $\bar v \in H^k(\Omega)^D$, and
\begin{equation*}
\|v_t - \bar v\|_{H^k} \le C\big((1-t)^\nu + |\lambda(t) - \bar\lambda|\big), \qquad t \in [0,1),
\end{equation*}
where $C$ depends only on $k$, $D$, $L$, $\nu$ and $\kappa$.
If (A2) holds as well, then $\int_\Omega \|v_t - v_N\|^2 \rho_N\, dx \to \ell_1(\bar v) \ge \operatorname{tr}\operatorname{Cov}(\gamma)$ as $t \to 1$.
\end{theorem}
\begin{proof}
As in the proof of \cref{thm:existence}, we first verify the assumptions of the existence theorem for $\bar F$, with~\cref{eq:holder} in place of~\cref{eq:Mt}.
Expanding the square in $\ell_1$ gives $\bar F(v) = \frac{1}{2}\bar a(v,v) - \bar f(v) + \frac{1}{2N}\sum_{i=1}^N \mathbb{E}\,\|x_i - z\|^2$, with
\begin{align*}
\bar a(v,w) &= \frac{1}{N}\sum_{i=1}^N v(x_i)\cdot w(x_i) + \bar\lambda \int_\Omega \nabla^k v : \nabla^k w\, dx + \mu \int_\Omega v\cdot w\, dx,\\
\bar f(w) &= \frac{1}{N}\sum_{i=1}^N (x_i - \mathbb{E}\,z)\cdot w(x_i).
\end{align*}
By~\cref{eq:holder} and since $\|x_i - \mathbb{E}\,z\| \le \operatorname{diam}(\Omega)$, $|\bar a(v,w)| \le (c_H^2 + \bar\lambda + \mu)\|v\|_{H^k}\|w\|_{H^k}$ and $|\bar f(w)| \le \operatorname{diam}(\Omega)\, c_H \|w\|_{H^k}$.
Since $\bar\lambda \ge \lambda_0$, \cref{lem:uniform} at $t = 1$ with $x_i' = x_i$ gives $\bar a(u,u) \ge \kappa\|u\|_{H^k}^2$.
Hence $\bar F$ has a unique minimizer $\bar v$.
The last paragraph of the proof of \cref{thm:existence} gives $\bar F(g) - \bar F(\bar v) \ge \frac{\kappa}{2}\|g - \bar v\|_{H^k}^2$, and $\bar F \ge 0$ and $\bar F(0) \le \frac{1}{2}\operatorname{diam}(\Omega)^2$ give $\|\bar v\|_{H^k} \le R$.

Fix $t \in [0,1)$ and let $h = \bar v - v_t$.
\Cref{lem:compare} with $G = F_t$, $G' = \bar F$ and $\beta = \kappa/2$ applies, by the second bound of \cref{thm:existence} and the previous paragraph.
The difference $F_t - \bar F$ consists of $\frac{1}{2}(\ell_t - \ell_1)$, the regularizer difference $\frac{1}{2}(\lambda(t) - \bar\lambda)|\cdot|_{H^k}^2$ and a constant, so
\begin{equation*}
4\beta \|h\|_{H^k}^2 \le \big(\ell_t(\bar v) - \ell_t(v_t)\big) - \big(\ell_1(\bar v) - \ell_1(v_t)\big) + \big(\lambda(t) - \bar\lambda\big)\big(|\bar v|_{H^k}^2 - |v_t|_{H^k}^2\big).
\end{equation*}

For the first part, with $y$, $z$ and $x_t$ as in the proof of \cref{thm:stability}, $\ell_1(g) = \mathbb{E}\|g(y) - (y - z)\|^2$.
With $S = \bar v + v_t$ and the identity of the proof of \cref{thm:stability}, the first part equals
\begin{equation*}
\mathbb{E}\Big[\big(h(x_t) - h(y)\big) \cdot \big(S(x_t) - 2(y - z)\big) + h(y) \cdot \big(S(x_t) - S(y)\big)\Big].
\end{equation*}
Since $x_t - y = (1-t)(z - y)$ and $\|\bar v\|_{H^k}, \|v_t\|_{H^k} \le R$, \cref{eq:holder} shows that it is at most
\begin{equation*}
c_H \|h\|_{H^k}\big(4 c_H R + 2\operatorname{diam}(\Omega)\big)\operatorname{diam}(\Omega)^\nu (1-t)^\nu.
\end{equation*}
For the second part, $|\bar v|_{H^k}^2 - |v_t|_{H^k}^2$ is the bilinear form of the seminorm evaluated at $h$ and $\bar v + v_t$, so by the Cauchy--Schwarz inequality its absolute value is at most $2R\, \|h\|_{H^k}$.
Dividing by $4\beta\|h\|_{H^k}$ gives the bound.

Let (A2) hold and $t \in (t_N, 1)$.
Then $\ell_t(v_N) = 0$ by \cref{lem:boxes}(b), so~\cref{eq:dist-vN} gives $\int_\Omega \|v_t - v_N\|^2 \rho_N\, dx = \ell_t(v_t)$.
The identity above gives $\ell_t(v_t) - \ell_1(\bar v) = \mathbb{E}\big[\big(v_t(x_t) - \bar v(y)\big)\cdot\big(v_t(x_t) + \bar v(y) - 2(y - z)\big)\big]$.
The second factor is bounded by $2c_H R + 2\operatorname{diam}(\Omega)$, and the triangle inequality and~\cref{eq:holder} give
\begin{equation*}
\|v_t(x_t) - \bar v(y)\| \le c_H \|v_t - \bar v\|_{H^k} + c_H R\,\big((1-t)\operatorname{diam}(\Omega)\big)^\nu,
\end{equation*}
which tends to zero by the bound above, so $\ell_t(v_t) \to \ell_1(\bar v)$.
Finally, $z$ is independent of $y$, so the cross term in $\|\bar v(y) - y + \mathbb{E}\,z + (z - \mathbb{E}\,z)\|^2$ has expectation zero, and $\mathbb{E}\,\|z - \mathbb{E}\,z\|^2 = \operatorname{tr}\operatorname{Cov}(\gamma)$ gives
\begin{equation*}
\ell_1(\bar v) = \frac{1}{N}\sum_{i=1}^N \|\bar v(x_i) - x_i + \mathbb{E}\,z\|^2 + \operatorname{tr}\operatorname{Cov}(\gamma).
\end{equation*}
The sum is nonnegative, so $\ell_1(\bar v) \ge \operatorname{tr}\operatorname{Cov}(\gamma)$.
\end{proof}
By \cref{thm:latetime}, the regularized field does not follow the targets of slope $-1/(1-t)$, and the theorem covers the Hessian regularizer for $D \le 3$.
If $\pi_N'$ satisfies the hypotheses of \cref{thm:latetime} as well, letting $t \to 1$ in the triangle inequality through $v_t$ and $v_t'$ extends \cref{thm:stability} to the limits, $\|\bar v - \bar v'\|_{H^k} \le C\, W_1(\pi_N,\pi_N')^\nu$.

The decomposition at the end of the proof holds for every field in place of $\bar v$.
For the Hessian regularizer with $\mu = 0$, the field $x - \mathbb{E}\,z$ makes its sum vanish and has zero seminorm, so $\bar v = x - \mathbb{E}\,z$ for every unisolvent set of training points and every $\bar\lambda$, and $\ell_1(\bar v) = \operatorname{tr}\operatorname{Cov}(\gamma)$.
This limit is that of the population velocity (\cref{sec:formulation}), so the extension to $t = 1$ is trivial in this case.
The rate $(1-t)^\nu$ then also holds without the assumption $\lambda(t) \to \bar\lambda$, since $x - \mathbb{E}\,z$ also minimizes $\bar F$ with $\lambda(t)$ in place of $\bar\lambda$, and \cref{lem:compare} with this functional as $G'$ has no regularizer difference.

\Cref{ex:tikhonov} shows that the smallness regularizer neither keeps the field away from the memorizing velocity nor stabilizes the problem unless $\lambda(t)$ grows fast enough, in which case the flow stays near the latent distribution.

\begin{example}[Smallness regularizer]\label[example]{ex:tikhonov}
Let $k = 0$, so that the regularizer and the mass term combine to $\frac{1}{2}(\lambda(t) + \mu)\|v\|_{L^2(\Omega)}^2$, let (A1)--(A3) hold, and let $\lambda(t) > 0$.
The problem at time $t$ contains no derivatives and therefore decouples in $x$, and pointwise minimization gives
\begin{equation*}
v_t = \frac{\rho_N}{\rho_N + \lambda(t) + \mu}\, v_N,
\end{equation*}
the empirical velocity shrunk by a factor between zero and one, with $v_t = 0$ where $\rho_N = 0$, as for the regularized empirical score~\cite[Thm.~5.1]{baptista2025memorization}.
For $t \in (t_N, 1)$ and $x = t x_i + (1-t)z$ in the box $\Omega_i(t)$, \cref{lem:boxes}(b) gives $\rho_N = K(z)/(N(1-t)^D)$ and $v_N = x_i - z$, so
\begin{equation*}
v_t(x) = \frac{K(z)}{K(z) + N(\lambda(t) + \mu)(1-t)^D}\,(x_i - z),
\end{equation*}
and the factor depends on $t$ only through $q(t) = N(\lambda(t) + \mu)(1-t)^D$.
If $\lambda(t)(1-t)^D \to 0$ as $t \to 1$, for instance for a bounded $\lambda$, then $q(t) \to 0$, and $v_t(x)$ tends to $x_i - z$ for every $z$ with $K(z) > 0$, in agreement with \cref{prop:localized} for $k = 0$ when $\lambda$ is bounded below.
The regularized field thus approaches the empirical velocity, whose flow memorizes the training points; we do not show that the flow of $v_t$ memorizes.
If instead $\lambda(t)(1-t)^D \ge \Lambda > 0$ for all $t$, then $\rho_N \le (1-t)^{-D}\|K\|_\infty$ and $\|v_N\| \le \operatorname{diam}(\Omega)$ on the support of $\rho_N$ give
\begin{equation*}
\sup_{x \in \Omega,\, t \in [0,1)} \|v_t(x)\| \le \frac{\|K\|_\infty \operatorname{diam}(\Omega)}{\Lambda},
\end{equation*}
so every absolutely continuous solution of $\dot x = v_t(x)$ moves by at most this distance, and a law generated by such solutions lies within it of the latent distribution in $W_1$; we do not show that such solutions exist, since $v_t$ need not be continuous in $x$.
The bound includes weights that grow like $(1-t)^{-D}$, for which the factor on the boxes tends to a constant, and the bound is small only for large $\Lambda$.

Next, for $\lambda(t)(1-t)^D \to 0$, we move one training point.
Let $x_1^\delta = x_1 + \delta e$ with $\delta > 0$ and a unit vector $e$, such that $x_1^\delta$ lies in $\Omega$ and differs from the other training points, and let $v_t^\delta$ be the minimizer, for the same $\lambda$ and $\mu$, for the perturbed data $\pi_N^\delta$.
Then $W_1(\pi_N, \pi_N^\delta) \le \delta/N$, since only $x_1$ moves.
For $t$ close to one, $\Omega_1(t)$ meets no box of $\pi_N^\delta$, so $v_t^\delta = 0$ on $\Omega_1(t)$, and dominated convergence with $z \sim \gamma$ gives
\begin{align*}
\|v_t - v_t^\delta\|_{L^2(\rho_N(\cdot,t))}^2 &\ge \int_{\Omega_1(t)} \|v_t\|^2 \rho_N\, dx \\ &= \frac{1}{N}\int_\Omega \Big(\frac{K(z)}{K(z) + N(\lambda(t) + \mu)(1-t)^D}\Big)^2 \|x_1 - z\|^2 K(z)\, dz \\
&\to \frac{1}{N}\,\mathbb{E}\,\|x_1 - z\|^2 \ge \frac{\operatorname{tr}\operatorname{Cov}(\gamma)}{N}.
\end{align*}
The last inequality holds because $\mathbb{E}\,\|a - z\|^2 = \|a - \mathbb{E}\,z\|^2 + \operatorname{tr}\operatorname{Cov}(\gamma)$ for every $a \in \R^D$.
However small $\delta$ is, the limit inferior as $t \to 1$ of the distance of the fields in $L^2(\rho_N(\cdot,t))$ is thus at least $(\operatorname{tr}\operatorname{Cov}(\gamma)/N)^{1/2}$.
For $2k > D$, unisolvent points and $\delta \le \delta_0$, in contrast, the remark after \cref{thm:stability} and~\cref{eq:holder} bound this distance by $c_H C\,(\delta/N)^\nu$, uniformly in $t$.
\end{example}

\section{Numerical Solution in One and Two Dimensions}\label{sec:numerics}
We discretize~\cref{eq:regfm} by tensor-product cubic splines and solve the problem at each time by a sparse direct method, so that the computed field is the minimizer of the discrete problem up to rounding; the only approximations are the spline space and the time integration of the flow, whose effect \cref{sec:discretization} measures.

\subsection{Spline Discretization and Solver}\label{sec:solver}

Since the problem decouples across time, we describe the solution of~\cref{eq:regfm} at fixed $t$.
Each component of the minimizer of~\cref{eq:regfm} satisfies, in the sense of distributions, the Euler--Lagrange equation
\begin{equation}\label{eq:euler-lagrange}
\big(\rho_N(\cdot,t) + \mu + \lambda(t)\, (-\Delta)^k\big)\, v_d(\cdot,t) = \rho_N(\cdot,t)\, v_{N,d}(\cdot,t) \quad \text{in } \Omega, \qquad d = 1,\ldots,D,
\end{equation}
since $|u|_{H^k}^2$ has the first variation $2\int_\Omega \chi\, (-\Delta)^k u\, dx$ for perturbations $\chi$ with compact support in $\Omega$.
We impose no boundary conditions, so for $k \ge 1$ the minimizer satisfies the natural boundary conditions of the energy, which the Galerkin discretization below imposes weakly.

We expand each component $v_d$ in a spline basis with the coefficient vector $\mathbf{v}_d$.
The basis consists of the tensor products of the $n = m+3$ cubic B-splines on $m$ uniform cells in each direction, which contain the affine functions and lie in $H^2(\Omega)$.
Minimizing~\cref{eq:regfm} over the spline space gives the normal equations
\begin{equation}\label{eq:normaleq}
\left(\mathbf{A} + \mu\mathbf{M} + \lambda(t)\, \mathbf{P}\right) \mathbf{v}_d = \mathbf{b}_d, \qquad d = 1, \ldots, D,
\end{equation}
where $\mathbf{A}$ and $\mathbf{b}_d$ discretize the data term, $\mathbf{P}$ the regularizer, and the mass matrix $\mathbf{M}$ the mass term.
This system is the Galerkin discretization of~\cref{eq:euler-lagrange} in weak form, which needs $k$ derivatives of the basis functions instead of the $2k$ of the strong form.
We compute the Gram matrices of the basis functions and their derivatives exactly; \cref{sec:splinedetails} gives them and the data matrices in one and two dimensions.
We integrate the data term to machine precision, which matters at late times, when the side length $2L(1-t)$ of the boxes of \cref{lem:boxes} falls below the mesh width.

Since the spline space is a subspace of $H^k(\Omega)^D$ and the problem at time $t$ is a coercive symmetric quadratic minimization (\cref{thm:existence}), the spline solution is, by Galerkin orthogonality, the best approximation of $v_t$ in the energy norm $a_t(v,v)^{1/2}$ of the objective, and hence quasi-optimal in $H^k(\Omega)^D$, with a constant that depends on $t$, by C\'ea's lemma~\cite[Thm.~2.4.1 and Rem.~2.4.1]{ciarlet2002finite}.

The matrix of~\cref{eq:normaleq} is sparse, symmetric positive definite, and shared by all components.
In one dimension, the matrix is banded, and the factorization costs $\mathcal{O}(n)$ operations.
In two dimensions, the regularizer and mass matrices are sums of Kronecker products of one-dimensional Gram matrices, and the data matrix $\mathbf{A}$ is a sum of $N$ Kronecker products, because the truncated Gaussian latent factorizes over the coordinates.
The matrix is sparse with 49 nonzeros per row and bandwidth $3n + 3$, and we factorize it by a sparse Cholesky method~\cite{chen2008cholmod}; since the sparsity pattern does not depend on $t$, we assemble $\mathbf{P}$ and its null space once and reuse the symbolic analysis across $t$ and $\lambda$.

Since the analysis requires $\lambda$ bounded below, we compute the limit $\lambda \to 0$ at the floor $\lambda = 10^{-14}\,\lambda_s$, with the scale $\lambda_s = \max_j \mathbf{A}_{jj} / \max_j \mathbf{P}_{jj}$ at which both terms balance.
For large $\lambda$, the rounding error of the component of the solution in the null space of $\mathbf{P}$ grows proportionally to $\lambda$.
So, for $\lambda \ge 100\,\lambda_s$, we compute this component separately, as in the transformation of general-form Tikhonov regularization to standard form (\cref{sec:splinedetails}); see, e.g.,~\cite[Sec.~9.2]{englRegularization1996}.

\subsection{Selection of the Regularization Parameter}\label{sec:cv}

Classical rules select the regularization parameter of linear inverse problems from a noise level or from the residual, among them the discrepancy principle, the L-curve and generalized cross-validation~\cite[Secs.~4.3 and 4.5]{englRegularization1996}, \cite{hansen1992lcurve}.
The discrepancy principle needs a noise level, which the empirical problem lacks, and at late times the residual of the empirical velocity vanishes (\cref{lem:boxes}(b)); we use an oracle and a hold-out rule instead.
Both select $\lambda$ independently at the time points $t_0 < \cdots < t_T$, so each selection is a static regression problem that needs no integration of the flow.

When the population velocity is known, as in our experiments, an oracle selects
\begin{equation}\label{eq:oracle}
\lambda^*(t) = \operatorname*{arg\,min}_{\lambda} \mathbb{E}_{x \sim \rho(\cdot,t)}\left[\|v_\lambda(x,t) - v_\pi(x,t)\|^2\right],
\end{equation}
where $v_\lambda$ solves~\cref{eq:regfm} for the parameter $\lambda$ and $\rho(\cdot,t)$ is the population path.
We estimate the expectation from a fixed sample of $x_t$ drawn from $\rho(\cdot,t)$.
The oracle is not available in practice and serves as a reference.
A small velocity error is neither necessary nor sufficient for an accurate generated distribution, because errors may cancel or accumulate during time integration.
Therefore, our numerical experiments measure the errors between the generated and the target distribution.

The hold-out rule is cross-validation with a single split: we divide the training samples into a fitting part and a held-out part and select the candidate whose fit to the fitting part has the smallest loss~\cref{eq:efm} on the held-out samples.
It estimates the prediction error of the fitted field from held-out samples.
By~\cref{eq:dist-vN}, this criterion is, up to a constant, the $L^2$ distance of the candidate to the empirical velocity of the held-out samples, weighted by their path $\rho_N(\cdot,t)$.
In the spline space, it is $\mathbf{v}_d^\top \mathbf{A}_{\rm val} \mathbf{v}_d - 2\,\mathbf{v}_d^\top \mathbf{b}_{d,\rm val}$, where $\mathbf{v}_d$ solves~\cref{eq:normaleq} with the matrices of the fitting part and $\mathbf{A}_{\rm val}$ and $\mathbf{b}_{d,\rm val}$ are the matrices~\cref{eq:datamatrix} of the held-out part.
With the selected $\lambda$, we solve again with all samples.

Between the time points, we interpolate $\log\lambda$ linearly in $\tau = -\log(1-t)$, and between two time points of which one selects $\lambda = 0$ we use the floor of \cref{sec:solver}; the last time point is the end time $t_{\rm end}$ of the flow, so any extension beyond it, for instance the constant $\lambda(t_{\rm end})$, satisfies the hypothesis of \cref{thm:latetime}.
To evaluate the velocity at any other time, we solve the normal equations at that time with the interpolated $\lambda$.
Interpolating the solution in time instead would add a smoothing that $\lambda$ does not control, which changed the total variation error by up to $0.005$ in a test on one training set each of the circle and the trimodal target.

\subsection{Sampling}\label{sec:sampling}

To sample, we integrate $\dot{x} = v(x,t)$ in the time variable $\tau = -\log(1-t)$, for which $t(\tau) = 1 - e^{-\tau}$ and
\begin{equation}
\frac{dx}{d\tau} = v\big(x, t(\tau)\big)e^{-\tau}, \qquad \tau \in [0, \tau_{\rm end}], \quad \tau_{\rm end} = -\log(1 - t_{\rm end}).
\end{equation}
We use the classical fourth-order Runge--Kutta method (RK4) with equidistant steps in $\tau$ and, except in \cref{sec:discretization}, the end time $t_{\rm end} = 1 - 10^{-4}$, so $\tau_{\rm end} \approx 9.2$.
Since $t = 1$ corresponds to $\tau = \infty$, the steps in $t$ shrink in proportion to $1 - t$ as the flow approaches the data, where the velocity changes fastest.

For $D \le 3$ and unisolvent training points, the field with the Hessian regularizer stays bounded in the supremum norm, uniformly in $t$ (\cref{eq:R,eq:holder}), so every solution of the flow equation moves by at most a constant times $1 - t_{\rm end}$ after $t_{\rm end}$.
The empirical velocity $v_N$ instead has the slope $-1/(1-t)$ on the boxes of \cref{lem:boxes}, so its flow keeps contracting the samples toward the training points.
Since the bound is qualitative, we measure the effect of stopping at $t_{\rm end}$ (\cref{sec:discretization}).
In two dimensions, $H^2(\Omega)$ embeds only into H\"older spaces with exponent below one, so the theory does not guarantee unique trajectories; the spline field is a piecewise polynomial in $x$ and hence Lipschitz for each fixed $m$.
Along each path, we also integrate $-e^{-\tau}\,\nabla\cdot v\big(x, t(\tau)\big)$ to obtain the density of the generated samples, which we use to estimate the total variation error.

\section{Numerical Experiments}\label{sec:experiments}

We test the analysis of \cref{sec:method} in one and two dimensions.
We first measure how well the flows learn the targets, with KDEs as a baseline (\cref{sec:flowaccuracy}), then show that the regularizer, and not the discretization, causes the generalization of the flows (\cref{sec:discretization}), test the stability with respect to the training data (\cref{sec:stability}), compare the regularizers of order zero, one and two (\cref{sec:operator}), and study the choice of $\lambda(t)$ (\cref{sec:lambda}).

\subsection{Datasets}\label{sec:data}
We use two one-dimensional and four two-dimensional targets, each chosen to test one difficulty for a density estimator.
The one-dimensional targets are mixtures of Beta distributions, mapped affinely from $[0,1]$ to $[-1,1]$:
\begin{itemize}
\item The trimodal density $\pi = 0.30\,\mathrm{B}(8,24) + 0.45\,\mathrm{B}(30,28) + 0.25\,\mathrm{B}(24,6)$ is smooth and its three modes have similar widths, so a single bandwidth suits the whole domain and a well-tuned fixed-bandwidth KDE is a strong baseline.
\item The inhomogeneous density $\pi = 0.8\,\mathrm{B}(4,5) + 0.2\,\mathrm{B}(1500,500)$ combines a broad component with standard deviation $0.31$ and a sharp mode at $x = 0.5$ with standard deviation $0.019$.
No single bandwidth resolves both length scales, so this example tests whether an estimator adapts to the local density.
\end{itemize}
As two-dimensional targets, we construct a noisy circle and use three toy datasets~\cite{Grathwohl2019}:
\begin{itemize}
\item The noisy circle concentrates the mass near a curve with a density that varies along it.
Its samples are $y = \varrho(\cos\theta, \sin\theta)$ with radius $\varrho \sim \mathcal{N}(2, 0.1^2)$ and, independently, an angle from the von Mises mixture $0.5\,\mathrm{vM}(0, 4) + 0.3\,\mathrm{vM}(2\pi/3, 16) + 0.2\,\mathrm{vM}(4\pi/3, 64)$.
\item The checkerboard is the uniform law on the eight dark squares of side $2$ of a $4 \times 4$ board on $[-4,4]^2$, so its density jumps at the square edges.
\item The eight Gaussians test well-separated modes.
They form the equal-weight mixture of $\mathcal{N}(\bar y_k, (0.5/1.414)^2 I)$ with $\bar y_k$ on a circle of radius $4/1.414$, where we keep the divisor $1.414$ of the reference implementation.
\item The two spirals concentrate the mass near two interleaved curves.
With $u = 3\pi\sqrt{U}$, $U$ uniform on $[0,1]$, and $\zeta(u) = (-u\cos u, u\sin u)$, its samples are $y = (\pm\zeta(u) + 0.15\,\eta)/3$ with $\eta \sim \mathcal{N}(0, I)$ and either sign with probability $1/2$.
The reference implementation instead adds noise uniform on $[0, 0.5]^2$ before and Gaussian noise with standard deviation $0.1$ after the division by $3$.
We replace both by the Gaussian noise above, so that the density is a one-dimensional integral along the curve, which we evaluate by quadrature to compute the total variation error.
The arms of our spirals are therefore about half as wide as those of the reference law.
\end{itemize}

We draw five independent training sets of each size $N \in \{100, 1000, 3000\}$ for each target.
All methods use the same training sets, and the flows share the latent samples, so differences between methods on one training set are not affected by the sampling of the training data.
We report means and standard errors over the five sets and call a difference unresolved when it is smaller than twice its Monte Carlo standard error or than its spread across the sets.

\subsection{Experimental Setup}\label{sec:setup}
The box $\Omega = [-L, L]^D$ of \cref{sec:formulation} has $L = 5$ in one and $L = 4.5$ in two dimensions, with the frames of \cref{sec:frames}, so the truncated latent differs from $\mathcal{N}(0, I)$ in total variation by $5.7 \cdot 10^{-7}$ and $1.4 \cdot 10^{-5}$.
We map a fixed box that contains the training samples into $\Omega$ by the frame $y \mapsto (y - c)/s$, where $c$ is its center and $s > 0$ maps its largest half-width to $L$, so the mapped training points satisfy (A3).
The computation is therefore the same for every shift and scalar rescaling of the training data and their box.
In the original coordinates, this is flow matching from the latent $\mathcal{N}(c, s^2 I)$ conditioned on the cube $c + s\Omega$.
The weight of the mass term is $\mu = 10^{-8}$ in this frame.
We discretize each flow with $m = 512$ cubic B-spline cells on $\Omega$ in one dimension and $m = 128$ cells per axis in two dimensions.
\Cref{sec:discretization} shows that the tuned errors do not change on finer grids.

We select $\lambda$ by the oracle~\cref{eq:oracle} and by hold-out with an $80/20$ split followed by a refit on all samples.
Each rule minimizes its criterion over $\lambda = 0$ and the candidates $\lambda = 10^j \lambda_s$ on a grid of integers $j$, with the scale $\lambda_s$ defined in \cref{sec:solver} at the same $t$, and then over the two half-decade neighbors of the best value (\cref{sec:splinedetails}).
Here $\lambda = 0$ stands for the floor $10^{-14}\lambda_s$ of \cref{sec:solver}, so every selected $\lambda(t)$ is positive.
Figures report $\lambda(t)/\lambda_s$; since $\lambda_s$ depends on $t$ through $\mathbf{A}$, a constant ratio is a time-dependent $\lambda(t)$.

The rules select $\lambda$ at $49$ time points, $19$ equidistant in $[0, 0.9]$ and $30$ with $1 - t$ equidistant in $\log$ from $10^{-1.1}$ to $10^{-4}$, and the sampler takes $96$ RK4 steps in $\tau$.
Integrated with the same scheme, the population velocity reaches a total variation error of $1.0 \cdot 10^{-4}$ for the trimodal and $2.5 \cdot 10^{-4}$ for the inhomogeneous target, of the same order with $48$ and $192$ steps, which is two orders of magnitude below the errors we report.

We measure the error of a generated density $p$ on $\R^D$ by the total variation (TV) distance
\begin{equation}\label{eq:tv}
\mathrm{TV}(p, \pi) = \int_{\R^D} \left(p(x) - \pi(x)\right)_+ dx = \mathbb{E}_{x \sim p}\left[\left(1 - \frac{\pi(x)}{p(x)}\right)_+\right],
\end{equation}
which we estimate by the second expression of~\cref{eq:tv}, using the density $p(x)$ that the flow transports along each path (\cref{sec:sampling}) and the density itself for the KDEs.
Let $F_L$ be the distribution function of the truncated latent in one dimension.
In one dimension, the flows use the images under $F_L^{-1}$ of the midpoints of $3000$ equal subintervals of $[0,1]$.
In two dimensions, they use the first $2^{12}$ points of the scrambled Sobol' sequence of SciPy~\cite{virtanen2020scipy}, mapped to the latent coordinatewise by $F_L^{-1}$.
The Sobol' sequence is a low-discrepancy sequence in the unit cube~\cite{sobol1967distribution}.
The KDEs use the same number of independent draws.
A Wasserstein distance would not detect memorization, since $\pi_N$ is close to $\pi$ in Wasserstein distance for large $N$ but at total variation distance one.

The positive part gives the total variation distance because $p - \pi$ integrates to zero, and its integrand $(1 - \pi/p)_+ \in [0,1]$ needs only samples of $p$ and the values of $p$ at them.

To compare the velocities, we evaluate $e(t) = \mathbb{E}_{x \sim \rho(\cdot,t)}\|v_\lambda(x,t) - v_\pi(x,t)\|^2$ with $2048$ samples of the population path in one and $4096$ in two dimensions.
All computations, including the assembly, the factorizations, the time integration, and the error measures, use double precision.

\subsection{Accuracy of the Regularized Flow}\label{sec:flowaccuracy}

To measure how well the regularized flow learns the target, we compare it with KDEs, an established method whose bandwidth rules are mature.
A comparison of trained diffusion models with a KDE whose bandwidth follows Silverman's rule of thumb does not separate the regularization from the architecture and the training~\cite{kwon2026factorizable}.
Here, the flow and the KDE both select their parameters from the data, and the flow involves neither a network nor its training.
We select the bandwidths by least-squares cross-validation (LSCV)~\cite{rudemo1982empirical,bowman1984alternative}, by the plug-in rule of Sheather and Jones~\cite{sheather1991reliable}, and by Abramson's adaptive rule~\cite{abramson1982bandwidth}, with second- and fourth-order kernels (\cref{sec:kde}); see, e.g.,~\cite[Sec.~1.4]{tsybakov2009introduction} for cross-validation.
For each target, we report the data-driven KDE with the lowest error averaged over the three values of $N$, which we select with the true density, so the baseline favors the KDEs.
\Cref{tab:main} lists the flow with oracle, hold-out and constant $\lambda$ and this KDE.
The hold-out flow is more accurate than the KDE on the two targets whose mass concentrates near a curve, on the noisy circle for every $N$ and on the spirals for $N \ge 1000$, with errors $23$ to $39\%$ lower.
On the other four targets and on the spirals with $N = 100$, its error is at most $14\%$ above that of the KDE.
Against the best KDE for each $N$ separately, its error is $18$ to $39\%$ lower on the noisy circle and on the spirals with $N \ge 1000$, and between $2\%$ lower and $26\%$ higher in the other cases.
Its error decreases with $N$ on every target, from $0.13$ to $0.57$ at $N = 100$ to $0.03$ to $0.16$ at $N = 3000$, far below the value one of a memorizing flow; \cref{fig:twod-ops} shows the two-dimensional densities at $N = 1000$.

\begin{table}[tbp]
\centering
\caption{Total variation error for all targets, mean $\pm$ standard error over five training sets, for the flows and for the data-driven KDE with the lowest mean error. Bold: the lowest data-driven mean and the data-driven entries not resolved from it (difference below twice the combined Monte Carlo or set-to-set standard error), per target and $N$. \Cref{tab:oned-full,tab:twod-full} list all rules, including the oracle KDEs. FM: flow matching with the Hessian regularizer ($k = 2$); hold-out constant: one $\lambda$ selected by hold-out for all times (\cref{sec:lambda}). KDE bandwidth rules: Sheather--Jones plug-in~\cite{sheather1991reliable}, least-squares cross-validation (LSCV)~\cite{rudemo1982empirical,bowman1984alternative} and Abramson's adaptive rule~\cite{abramson1982bandwidth}; the fourth-order and product kernels are defined in \cref{sec:kde}.}
\label{tab:main}
{\footnotesize\input{tables/tab_main.tex}}
\end{table}

\subsection{Excluding Implicit Regularization from Discretization}\label{sec:discretization}
To show that the discretization does not contribute to the generalization of the hold-out flows, we refine the mesh, compare the time integrator with a reference, and integrate closer to the terminal time.
Discretization can regularize ill-posed problems, which is the principle of regularization by projection~\cite[Sec.~3.3]{englRegularization1996}.

\begin{figure}[tbp]
\centering
\includegraphics[width=\textwidth]{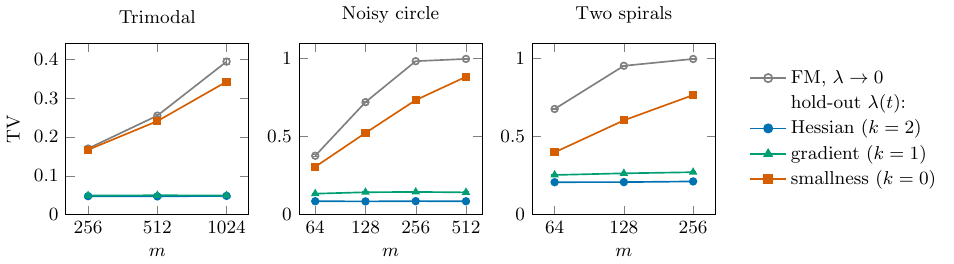}
\caption{Total variation error against the spline resolution $m$ for $N = 1000$, for the limit $\lambda \to 0$, computed at the floor $\lambda(t) = 10^{-14}\lambda_s$ of \cref{sec:solver}, and for the hold-out flows of the three regularizers, mean $\pm$ standard error over five training sets, for the trimodal target, the noisy circle and the two spirals; \cref{tab:refine-1d,tab:refine-2d} list all targets. The limit $\lambda \to 0$ and the flow with the smallness regularizer move toward memorization as $m$ grows, whereas the error of the Hessian flow changes by at most $0.007$.}
\label{fig:accuracy}
\end{figure}

As $\lambda \to 0$, the minimizer of~\cref{eq:regfm} tends to $\rho_N v_N/(\rho_N + \mu)$, which is close to $v_N$ if $\rho_N \gg \mu$.
Under mesh refinement, its error grows toward the value $1$ of a memorizing flow on every target.
The error of the flow with the smallness regularizer also grows toward $1$, albeit more slowly (\cref{fig:accuracy,tab:refine-1d,tab:refine-2d}).
The error of the Hessian flow changes by at most $0.007$ (\cref{fig:accuracy}).
The error of the gradient flow does not change in one dimension, where $2k > D$, and grows by up to $14\%$ from $m = 64$ to $m = 256$ in two dimensions, the borderline case $2k = D$ (\cref{tab:refine-1d,tab:refine-2d}).
Stopping at $t_{\rm end}<1$ changes the errors of the regularized flows by at most $4\cdot 10^{-4}$ (\cref{tab:tend}).
Compared with the adaptive eighth-order Runge--Kutta integrator DOP853, the $96$ RK4 steps change the total variation error of the hold-out flows by at most $2.6 \cdot 10^{-4}$ in one and $9 \cdot 10^{-4}$ in two dimensions (\cref{tab:integrator}).

\subsection{Stability with Respect to the Training Data}\label{sec:stability}

\begin{figure}[tbp]
\centering
\includegraphics[width=\textwidth]{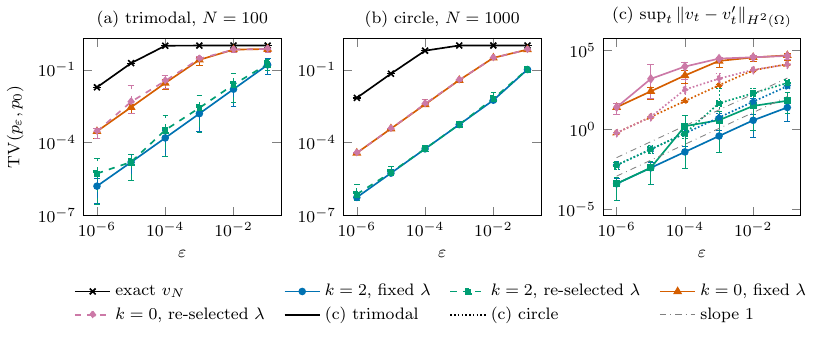}
\caption{Distance between the laws generated from the training set and from the training set with every point moved by $\varepsilon$, for (a) the trimodal target with $N = 100$ and (b) the noisy circle with $N = 1000$, and (c) the corresponding difference of the fields in $H^2(\Omega)$, mean over five training sets and two perturbation directions, with ranges, for the Hessian ($k = 2$) and the smallness ($k = 0$) regularizer with $\lambda(t)$ fixed to the unperturbed selection or re-selected by hold-out.}
\label{fig:stability}
\end{figure}

To test \cref{thm:stability}, we move every training point by $\varepsilon$ in a random direction, so that $W_1(\pi_N, \pi_N') \le \varepsilon$, and compare the fields and the generated laws for the perturbed and the unperturbed training sets, with $\lambda(t)$ fixed to its hold-out selection on the unperturbed set, as in the theorem, and re-selected by hold-out on the perturbed set (\cref{fig:stability,tab:stability}).
Since \cref{thm:stability} assumes $\lambda$ bounded below, hold-out in this test minimizes over the positive candidates only, which replaces the selection $\lambda = 0$ by the smallest candidate at $7$ (inhomogeneous target) and $15$ (noisy circle) of the $245$ pairs of training set and time.
The theorem concerns the field, not the generated law, so the distance of the laws tests whether the stability of the field carries over to the samples.
For the Hessian regularizer with fixed $\lambda(t)$, both $\mathrm{TV}(p_\varepsilon, p_0)$ and $\sup_t \|v_t - v_t'\|_{H^2(\Omega)}$ are proportional to $\varepsilon$, with fitted slopes $1.00$ on the trimodal target with $N = 100$ and on the noisy circle with $N = 1000$; on the circle, the slope of the field difference shows no logarithmic factor in this range.
With $\lambda(t)$ re-selected, the selected candidate changes at no more than $3$ of the $48$ times $t > 0$ for $\varepsilon \le 10^{-4}$, and the distances of the laws stay within a factor of $3.4$ of those with fixed $\lambda(t)$ for $\varepsilon \le 10^{-3}$, whereas the field differences can be larger by a factor of up to $42$.
The law of $v_N$ instead jumps: its total variation distance reaches $0.98$ at $\varepsilon = 10^{-4}$ in one dimension and $1.00$ at $\varepsilon = 10^{-3}$ on the circle, once $\varepsilon$ exceeds the width of its stopped law, and its field difference in $L^2(\rho_N(\cdot,t))$ grows tenfold per decade of $1-t$ for $1-t \le 10^{-2}$, as \cref{sec:intro} predicts.

The flow with the smallness regularizer is stable on each grid, but its stability constants grow under refinement.
From $m = 256$ to $m = 1024$, its constant $\mathrm{TV}(p_\varepsilon, p_0)/\varepsilon$ grows by a factor of $5$ to $9$ and that of its field difference in $H^2$ by a factor of $64$ to $109$, whereas the total variation and $L^2$ constants of the Hessian regularizer with fixed $\lambda(t)$ do not change and its $H^2$ constant grows by at most a factor of $6$ (\cref{tab:stability-refine}).
On the noisy circle with $m = 128$, the law and the field of the flow with the smallness regularizer change $68$ and $108$ times as much as those of the Hessian regularizer at $\varepsilon = 10^{-6}$; we computed its total variation on a grid with spacing $0.005$, since at the spacing $0.02$ of the other flows its density loses up to $1.1\%$ of its mass on the grid.

\subsection{Impact of the Regularization Operator}\label{sec:operator}

To test which property of the regularizer prevents memorization, and the distinction between $2k > D$ and $2k < D$ of \cref{thm:latetime,prop:localized,ex:tikhonov}, we replace the Hessian regularizer by the gradient and smallness regularizers and keep the discretization, selection rules and error measures (\cref{tab:reg-full}).
For the gradient regularizer in two dimensions, the borderline case $2k = D$, \cref{thm:existence} gives a unique minimizer, but no stability or late-time result of \cref{sec:wellposed} applies.
Each operator uses its own candidate grid (\cref{sec:splinedetails}), and hold-out puts the minimizer numerically in the null space of $|\cdot|_{H^k}$ for at most $5.4\%$ of the pairs of training set and time.

\begin{figure}[tbp]
\centering
\includegraphics{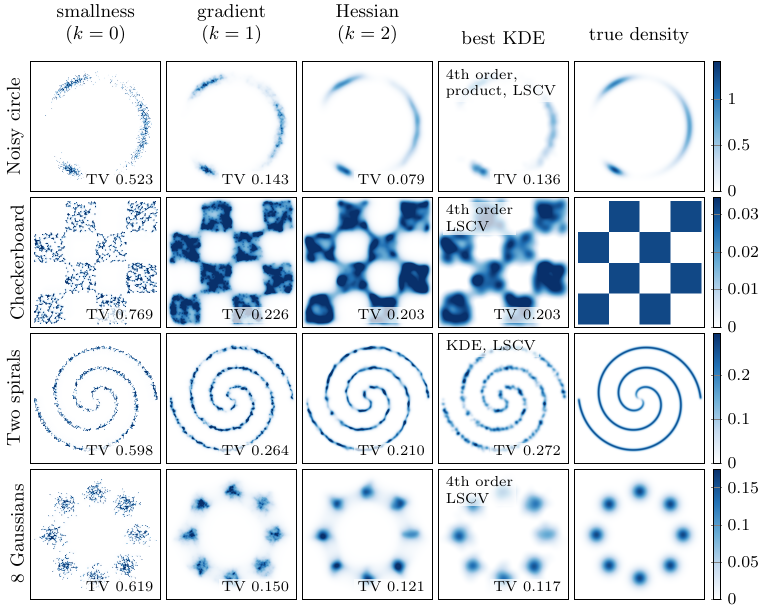}
\caption{Densities generated on the two-dimensional targets (rows) with $N = 1000$ by the flow with hold-out $\lambda(t)$ for the smallness ($k = 0$), gradient ($k = 1$) and Hessian ($k = 2$) regularizers, by the best data-driven KDE, and the true density (columns). Each flow uses the training set with its median error; the KDE uses the median set of the Hessian regularizer. Color scales are shared within a row, with values above $1.1$ times the maximum of the true density saturated. Panel tags give the total variation error.}
\label{fig:twod-ops}
\end{figure}

\Cref{tab:reg-full} shows that the Hessian regularizer gives the lowest hold-out error on every target for $N \ge 1000$ and is more accurate than the gradient regularizer there in $57$ of the $60$ training sets, and the smallness regularizer the highest on every target and $N$, by a factor of $1.5$ to $6.2$; at $N = 100$, the gradient regularizer ties the Hessian regularizer within two standard errors on the two one-dimensional targets and on the spirals.
On the trimodal target, the flow with the smallness regularizer places spikes at the training points, with $31\%$ of its samples within $10^{-3}$ of a training point against $15$ to $17\%$ for the other flows, because at late times hold-out selects $\lambda/\lambda_s \approx 3 \cdot 10^{6}$, which stops the samples where they are; the gradient regularizer gives a smooth density but flattens the modes.
In two dimensions, the flow with the smallness regularizer concentrates its density in isolated peaks that reach $13$ to $140$ times the largest value of the true density, while the gradient and Hessian flows spread the density over the support (\cref{fig:twod-ops}).

On both one-dimensional targets and for every $N$, the Hessian regularizer is more accurate than convolution of the velocity with a constant or a time-dependent bandwidth selected by hold-out (\cref{tab:oned-full}).
Convolution replaces the regularizer in kernel-smoothed score models~\cite{gabriel2025kernelsmoothed}.

\subsection{Choice of the Regularization Parameter}\label{sec:lambda}

In one dimension, the hold-out rule selects $\lambda(t)$ within one decade of the oracle at most late times.
Individual sets differ from the oracle by up to three decades, but the error of the generated law exceeds that of the oracle by at most $0.033$ at $N = 100$ and $0.007$ at $N \ge 1000$ (\cref{tab:main}).

One constant $\lambda(t) = \lambda_c$, chosen by the hold-out criterion of \cref{sec:cv} summed over the time points, changes the error of the Hessian flow by $-4$ to $10\%$ in one dimension and raises it by $3$ to $38\%$ in two dimensions, where it is less accurate than the hold-out $\lambda(t)$ on all $60$ training sets (\cref{tab:main}).
For the gradient regularizer at $N = 1000$, the changes are $-12$ to $8\%$ and $24$ to $34\%$ (\cref{tab:lam-sweep-full}).
Multiplying the hold-out selection at all times by one factor $\alpha$ raises the error of the Hessian flow by at most $20\%$ for $\alpha$ within half a decade on all six targets, with the minimum at $\alpha = 1$ on the two-dimensional targets and within one standard error of $\alpha = 1$ in one dimension.
On the two-dimensional targets, the gradient and smallness regularizers reach their lowest errors at larger $\alpha$; the hold-out criterion targets the velocity error, not the error of the generated law, and for these regularizers the minimizers of the two errors differ.
On the noisy circle, hold-out selects $\lambda$ below a tenth of its median over $t \ge 0.97$ at $77\%$ of the pairs of training set and time with $t < 0.9$, so a constant either smooths at early times, where hold-out does not, or smooths less than hold-out at late times.

\section{Summary and Open Problems}\label{sec:discussion}

We show that memorization and instability of the empirical flow matching problem can be avoided with explicit derivative-based regularization.
Our analysis shows that a derivative regularizer of order $2k>D$ renders the estimation of the velocity well posed uniformly in time if the training points are in general position (\cref{lem:uniform,thm:existence,thm:stability}).
In our experiments on one- and two-dimensional datasets, we exclude implicit regularization by the discretization, the optimization and the time integration.
The Hessian-regularized fields are stable under perturbations of the training points, as the theory predicts. The generated laws do not memorize, and their errors are comparable to or lower than those of well-tuned KDEs.

Our analysis requires a regularizer whose order grows with the dimension, $2k > D$, and for $2k < D$ the minimizer approaches the memorizing velocity.
With the smallness regularizer, the field either approaches the memorizing velocity or stays bounded by a constant that is small only for large weights (\cref{ex:tikhonov}), and the error of its flow grows toward that of memorization under mesh refinement.
In two dimensions, the borderline case $2k = D$, the error of the gradient flow grows by up to $14\%$ under refinement, whereas that of the Hessian flow changes by at most $0.007$ (\cref{sec:discretization}).

The well-posed formulation is a first step toward a better theoretical understanding and more efficient and reliable generative models, and our experiments in one and two dimensions are a proof of concept.
The following problems remain open for the analysis and the computation:
\begin{enumerate}
\item \emph{Stability and convergence of the generated law.} Although our theory covers only the velocity, our numerical experiments indicate stability of the generated law. A proof that the generated law depends continuously on the data, and converges to $\pi$, would prove that the regularized flow generalizes. In one dimension, $H^2(\Omega)$ embeds into $C^1(\overline\Omega)$~\cite[Thm.~4.12]{adams2003sobolev}. For $N \ge 2$, the Hessian-regularized field is then Lipschitz continuous in $x$, uniformly in $t$, which suggests starting there.

\item \emph{Consistency.} Regularization methods in the sense of Engl et al.\ come with a parameter choice that makes the regularized solution converge as the noise vanishes~\cite[Def.~3.1]{englRegularization1996}. For flow matching, this asks how $\lambda$ should depend on $N$, and under which source condition on the population velocity the regularized field converges to it as $N \to \infty$.
\item \emph{Higher dimensions.} The condition $2k > D$ asks for regularizers of higher order or, by the scaling of \cref{lem:boxes}(c), for a parameter that grows at least as fast as $(1-t)^{2k-D}$ as $t \to 1$, which excludes the localized fields of \cref{prop:localized} but is not covered by our results. For bounded $\lambda(t)$, a regularizer on the second derivatives, as in score smoothing~\cite{chen2026scoresmoothing}, has $2k < D$ for $D \ge 5$, and the case $D = 4$ is open.
\item \emph{The sampling equation.} For $2k > D$ with $k - D/2 \le 1$, as for the Hessian regularizer in two and three dimensions, the regularized field is H\"older but not necessarily Lipschitz continuous, so the uniqueness of its trajectories is not guaranteed by our results.
\item \emph{Scalable discretizations.} Whether the stability survives a discretization whose cost does not grow like $m^D$, such as a trained network, is open. One challenge in higher dimensions is to separate the effect of the regularizer from the implicit regularization provided by function approximators and early stopping in training.
\end{enumerate}

\section*{Acknowledgments}
We used Claude Opus 5.5 (Anthropic), through Claude Code, as a writing assistant, as a proof assistant, and to review intermediate drafts. Claude also searched the literature, checked the bibliographic entries against publisher records, wrote the code, and ran and managed the computational experiments. We have reviewed all code and the entire paper and take responsibility for its content.

\FloatBarrier
\bibliographystyle{siamplain}
\bibliography{references}

\FloatBarrier
\appendix
\section{Spline Discretization}\label{sec:splinedetails}

This section derives the matrices of the normal equations~\cref{eq:normaleq} of the main text.
The problem at each time is a penalized least-squares fit in a spline space, as for smoothing splines; see, e.g.,~\cite{wahba1990spline}.
The matrices of the regularizer and of the mass term are computed exactly, and those of the data term have a relative error below $5 \cdot 10^{-14}$.
They are banded in one dimension and sums of Kronecker products of banded matrices in two dimensions.

\subsection{One Dimension}

We represent the velocity in the cubic B-spline basis on uniform knots; see, e.g.,~\cite{deboor2001practical}.
To evaluate the basis, we map $\Omega$ to $[-1,1]$ by $x = L\xi$ and use the knots $\xi_j = -1 + j h_\xi$, $j = 0, \ldots, m$, with $h_\xi = 2/m$, so the mesh width is $h = L h_\xi$.
We write $v(x) = \sum_a \mathbf{v}_a b_a(\xi)$ with the coefficient vector $\mathbf{v} \in \R^n$ and the $n = m+3$ functions $b_a(\xi) = b((\xi - \xi_a)/h_\xi)$, $a = -1, \ldots, m+1$, restricted to $[-1,1]$, where $b$ is the cubic cardinal B-spline supported on $[-2,2]$.
The basis spans the twice continuously differentiable cubic splines on the knots, which contain the affine functions and lie in $H^2(\Omega)$.

The regularizer and the mass term need the Gram matrices of the derivatives of the basis functions, which we compute exactly.
For $k = 0, 1, 2$, the Gram matrix $(\mathbf{P}_k)_{aa'} = \int_\Omega b_a^{(k)} b_{a'}^{(k)}\, dx$ of the derivatives of order $k$ is symmetric with bandwidth three.
Except for the first and last three rows, which the restriction to $\Omega$ modifies, it is a Toeplitz matrix.
The matrices $\mathbf{P}_0$ and $\mathbf{P}_1$ are the mass and stiffness matrices of the Galerkin method with this basis; see, e.g.,~\cite{ciarlet2002finite}.
The regularizer of order $k$ is $\int_\Omega (v^{(k)})^2\, dx = \mathbf{v}^\top \mathbf{P}_k \mathbf{v}$.
The null space of $\mathbf{P}_2$ consists of the coefficients of the affine functions, the null space of $\mathbf{P}_1$ consists of the coefficients of the constants, and $\mathbf{P}_0$ is positive definite.

The data term needs the integrals of the basis functions against the kernels of the training points.
The matrix and vector of the data term in~\cref{eq:normaleq} are
\begin{equation}\label{eq:datamatrix}
\begin{aligned}
\mathbf{A} &= \frac{1}{N}\sum_{i=1}^N \mathbf{T}_i, & (\mathbf{T}_i)_{aa'} &= \int_\Omega b_a b_{a'}\, K_{1-t}(x - t x_i)\, dx,\\
\mathbf{b} &= \frac{1}{N}\sum_{i=1}^N \mathbf{g}_i, & (\mathbf{g}_i)_a &= \int_\Omega b_a\, K_{1-t}(x - t x_i)\, \frac{x_i - x}{1-t}\, dx.
\end{aligned}
\end{equation}
These follow from $\mathbf{A} = \int_\Omega \boldsymbol{\phi}\boldsymbol{\phi}^\top \rho_N\, dx$ and $\mathbf{b} = \int_\Omega \boldsymbol{\phi}\, \rho_N v_N\, dx$, where $\boldsymbol{\phi}$ collects the basis functions, because $\rho_N v_N = \frac{1}{N}\sum_{i=1}^N K_{1-t}(\cdot - t x_i)\,(x_i - \cdot)/(1-t)$ does not involve the weights $w_i$.
Each $\mathbf{T}_i$ has the same band structure as $\mathbf{P}_0$.
On each knot interval, the integrands are polynomials times the kernel, and Gauss--Legendre panels adapted to the decay of the kernel evaluate them with a relative error below $5 \cdot 10^{-14}$, measured against high-precision quadrature.
The truncated latent confines the kernel of $x_i$ to the box $t x_i + (1-t)\Omega$, so only the knot intervals that meet this box contribute, and the panels end at its edges.

The regularization matrix in~\cref{eq:normaleq} is $\mathbf{P} = \mathbf{P}_k$ for the regularizer of order $k$ and the mass matrix is $\mathbf{M} = \mathbf{P}_0$, so the matrix of the normal equations has bandwidth three.

\subsection{Two Dimensions}

In two dimensions, the basis is the tensor product of two one-dimensional bases, and every matrix is a sum of Kronecker products of the one-dimensional matrices; see, e.g.,~\cite[Chap.~XVII]{deboor2001practical}.
We map the box $\Omega$ to $[-1,1]^2$ by $x = L\xi$, with one-dimensional bases $b_a$ and $b_{a'}$ in the two directions as above.
We write each component as $v_d(x) = \sum_{a,a'} \mathbf{v}_{d,aa'}\, b_a(\xi_1) b_{a'}(\xi_2)$ and order the coefficients lexicographically into $\mathbf{v}_d \in \R^{n_1 n_2}$, with the second index running fastest.
Superscripts mark the direction of the one-dimensional matrices.

For the Hessian regularizer, the energy $\|\nabla^2 v_d\|_F^2 = (\partial_{11} v_d)^2 + 2(\partial_{12} v_d)^2 + (\partial_{22} v_d)^2$ separates into products of one-dimensional integrals, which gives
\begin{equation}\label{eq:regmatrix}
\int_\Omega \|\nabla^2 v_d\|_F^2\, dx = \mathbf{v}_d^\top \mathbf{P} \mathbf{v}_d, \qquad
\mathbf{P} = \mathbf{P}_2^{(1)} \otimes \mathbf{P}_0^{(2)} + 2\, \mathbf{P}_1^{(1)} \otimes \mathbf{P}_1^{(2)} + \mathbf{P}_0^{(1)} \otimes \mathbf{P}_2^{(2)},
\end{equation}
and the mass matrix is $\mathbf{M} = \mathbf{P}_0^{(1)} \otimes \mathbf{P}_0^{(2)}$.
The mixed partial derivative contributes the middle term, the only place where $\mathbf{P}_1$ enters.
It is also the term that restricts the null space of $\mathbf{P}$ to the affine functions: the first and third terms vanish on $x_1 x_2$, which the middle term penalizes.

The separation that gives~\cref{eq:regmatrix} also gives $\mathbf{P} = \mathbf{P}_1^{(1)} \otimes \mathbf{P}_0^{(2)} + \mathbf{P}_0^{(1)} \otimes \mathbf{P}_1^{(2)}$ for $k = 1$ and $\mathbf{P} = \mathbf{P}_0^{(1)} \otimes \mathbf{P}_0^{(2)}$ for $k = 0$.
All three matrices have the sparsity pattern of the data matrix below.

For a product kernel $K(x) = K^{(1)}(x_1) K^{(2)}(x_2)$, such as the standard Gaussian, each sample contributes a Kronecker product, and
\begin{equation*}
\mathbf{A} = \frac{1}{N}\sum_{i=1}^N \mathbf{T}_i^{(1)} \otimes \mathbf{T}_i^{(2)}, \qquad
\mathbf{b}_1 = \frac{1}{N}\sum_{i=1}^N \mathbf{g}_i^{(1)} \otimes \mathbf{f}_i^{(2)}, \qquad
\mathbf{b}_2 = \frac{1}{N}\sum_{i=1}^N \mathbf{f}_i^{(1)} \otimes \mathbf{g}_i^{(2)},
\end{equation*}
where $(\mathbf{f}_i)_a = \int_\Omega b_a\, K_{1-t}(x - t x_{i})\, dx$ and the one-dimensional matrices and vectors use the corresponding coordinate of $x_i$.

\subsection{Large Regularization Parameters}

For large regularization parameters, we compute the solution's component in the null space of $\mathbf{P}$ separately, because rounding would otherwise lose it.
For the Hessian regularizer and $\lambda \ge 100\,\lambda_s$, rounding in $\mathbf{A} + \mu\mathbf{M} + \lambda\mathbf{P}$ would remove the part of $\mathbf{A} + \mu\mathbf{M}$ that determines the affine component of the solution, on which $\mathbf{P}$ vanishes.
We therefore compute the affine component from $\mathbf{A} + \mu\mathbf{M}$ alone, through the $(D+1)$-dimensional null space of $\mathbf{P}$, and the remainder from the factorized complement.
This reduces the relative error at $\lambda = 10^8 \lambda_s$ from $10^{-8}$ to $10^{-15}$.
For the gradient regularizer, the same splitting uses the one-dimensional null space of the constants.
For the smallness regularizer, no splitting is needed, because the condition number of $\mathbf{P}$ does not grow with the mesh; it is $1.4 \cdot 10^3$ in one and $2.1 \cdot 10^6$ in two dimensions.

For every regularizer, the candidates are $\lambda = 0$ and $\lambda = 10^{j}\lambda_s$ for integers $j$ in a fixed range per regularizer and dimension, with $\lambda_s$ of the normal equations at the same time.
The range extends at least two decades beyond the selections of test runs.
In the final runs, selections at the largest candidate occur; for the Hessian and smallness regularizers they lie at the limit $\lambda \to \infty$, and for the gradient regularizer in one dimension ten oracle and hold-out selections at $t \le 0.25$ lie at the end of the range.
At the other end, for the Hessian and gradient regularizers, at most $17$ of the $2940$ selections in one dimension and $16$ of the $5880$ in two dimensions lie at the smallest positive candidate or its lower half-decade neighbor, all at $t \le 0.75$, where $\lambda = 0$ is also a candidate.
Since $\lambda_s$ and $\lambda$ transform alike under a change of the frame of \cref{sec:frames}, the selection does not depend on the frame.

\subsection{Domains, Frames and Latents}\label{sec:frames}
The frame is the map $y \mapsto (y - c)/s$ of \cref{sec:setup}, which shifts the data box by its center $c$ and scales it by $1/s$ into $\Omega$.
In one dimension, the latent is $\mathcal{N}(0, 1)$ conditioned on $\Omega = [-5, 5]$, which differs from $\mathcal{N}(0, 1)$ in total variation by $5.7 \cdot 10^{-7}$, and the frame has $c = 0$ and $s = 0.21$, which maps $[-1.05, 1.05]$ to $\Omega$. In two dimensions, the latent is $\mathcal{N}(0, I)$ conditioned on $\Omega = [-4.5, 4.5]^2$, at a total variation distance of $1.4 \cdot 10^{-5}$, and the frames have $c = 0$ and $s = 0.58$, $0.93$, $0.74$ and $0.94$ for the circle, the checkerboard, the spirals and the eight Gaussians.
We sample the truncated latent coordinatewise by the inverse of its distribution function, $z = F_L^{-1}(F_0(g))$ for $g \sim \mathcal{N}(0, 1)$, where $F_0$ is the standard normal distribution function and $F_L$ that of the truncated law, so every sample lies in $\Omega$.
Intermediate stages of the time integrator can leave $\Omega$, for at most $2.8\%$ of the trajectories on the checkerboard; there, we continue each spline by the cubic of its boundary cell up to one cell beyond $\Omega$ and by a constant beyond that.

\section{Benchmark Methods}\label{sec:kde}

We compare the flows with KDEs of the training data.
Let $G$ denote the standard normal density on $\R^D$, scaled as $G_\sigma(x) = \sigma^{-D} G(x/\sigma)$.
The KDE with bandwidth $\sigma > 0$ is $\hat\pi_\sigma = \frac{1}{N}\sum_{i=1}^N G_\sigma(\cdot - x_i)$, and $\hat\pi_{\sigma,-i}$ denotes the KDE without $x_i$.
Least-squares cross-validation (LSCV) selects the bandwidth that minimizes
\begin{equation*}
\mathrm{CV}(\sigma) = \int \hat\pi_\sigma^2\, dx - \frac{2}{N}\sum_{i=1}^N \hat\pi_{\sigma,-i}(x_i).
\end{equation*}
The criterion is an unbiased estimate of $\mathbb{E}\int (\hat\pi_\sigma - \pi)^2\, dx - \int \pi^2\, dx$, whose last term does not depend on $\sigma$; see, e.g.,~\cite[Sec.~1.4]{tsybakov2009introduction}.
The oracle selects the bandwidth that minimizes the total variation error.
Both minimize over a geometric grid of bandwidths: $241$ values in $[2\cdot 10^{-3}, 0.3]$ for the trimodal and $281$ values in $[5 \cdot 10^{-4}, 0.3]$ for the inhomogeneous target, $61$ values in $[0.02, 0.6]$ for the noisy circle, and $71$ values in $[0.04, 2]$, $[0.01, 0.6]$ and $[0.02, 1.5]$ for the checkerboard, the spirals and the eight Gaussians.

In one dimension, we also use two rules derived from the bandwidth
\begin{equation*}
\sigma^* = \Big(\frac{\int G^2\, du}{N \int (\pi'')^2\, dx}\Big)^{1/5}
\end{equation*}
that minimizes the asymptotic mean integrated squared error of the KDE~\cite{silverman1986density}.
For a normal density with standard deviation $q$, this bandwidth is $1.06\, q N^{-1/5}$.
Silverman's rule of thumb lowers the factor to $0.9$, so that it smooths multimodal densities less, and uses $\sigma = 0.9\, q N^{-1/5}$~\cite[Eq.~(3.31)]{silverman1986density}.
Here $q$ is the smaller of the sample standard deviation and the distance between the first and third sample quartiles divided by $1.349$, its value for a normal density.
The plug-in rule of Sheather and Jones estimates $\int (\pi'')^2\, dx$ by a KDE of $\pi''$ whose bandwidth depends on $\sigma$, and solves the resulting equation $\sigma = \sigma^*$ for $\sigma$~\cite{sheather1991reliable}.
The adaptive KDE of Abramson~\cite{abramson1982bandwidth} replaces $\sigma$ at the training point $x_i$ by
\begin{equation*}
\sigma_i = \sigma \Big(\frac{\tilde\pi(x_i)}{g}\Big)^{-1/2},
\end{equation*}
where $\tilde\pi$ is a preliminary KDE and $g$ is the geometric mean of $\tilde\pi(x_1), \ldots, \tilde\pi(x_N)$.
The preliminary KDE uses the Sheather--Jones bandwidth in one dimension and the LSCV bandwidth in two dimensions, and LSCV or the oracle selects $\sigma$.

Since the cubic smoothing spline in one dimension acts like a kernel smoother with a fourth-order kernel~\cite[Thm.~A and Eq.~(3.3)]{silverman1984spline}, we also use the fourth-order kernel $G^{[4]}(u) = \frac12(3 - u^2)\,G(u)$, whose second moment vanishes; see, e.g.,~\cite{wand1995kernel}.
In one dimension, we combine it with a fixed bandwidth and with the Abramson bandwidths $\sigma_i$.
In two dimensions, we use the radial kernel $G^{[4]}(u) = (2 - \|u\|^2/2)\,G(u)$, whose bias does not depend on the coordinate axes, and the product of two one-dimensional kernels.
The fourth-order KDE can be negative, so we set it to zero where it is negative and renormalize it on the evaluation grid.
LSCV and the oracle select its bandwidth as above, LSCV from the estimate before clipping.
The fourth-order KDEs of \cref{tab:main} clip at most $0.3\%$ of their mass in one dimension and at most $6.0\%$ in two dimensions, the maximum occurring on the noisy circle with $N = 100$.

For the noisy circle and the spirals, the anisotropic KDE places at $x_i$ the Gaussian with covariance $\sigma_T^2 e_i e_i^\top + \sigma_N^2 n_i n_i^\top$, where $e_i$ and $n_i$ are the exact unit tangent and normal of the curve.
LSCV or the oracle selects $(\sigma_T, \sigma_N)$ among the $138$ pairs $\sigma_T = 0.02 \cdot 1.28^{j}$ and $\sigma_N = 0.02 \cdot 1.28^{l}$ with $0 \le l \le \min(j, 11)$ and $j \le 16$.
The fourth-order anisotropic KDE uses the radial fourth-order kernel in the same coordinates and searches the same pairs locally, starting from the oracle bandwidth of the radial fourth-order KDE.
Since both use the exact tangent, we count them among the oracles.

Kernel-smoothed score models replace the regularizer by a convolution~\cite{gabriel2025kernelsmoothed}.
In one dimension, we apply this smoothing to the empirical velocity, $v_h(\cdot,t) = G_{t\sqrt{h}} * v_N(\cdot,t)$, with a constant $h$ or a time-dependent $h(t)$, and between the time points $\sqrt{h}$ is linear in $\log(1-t)$.
The hold-out rule of \cref{sec:cv}, with the same split, selects $h(t)$ at each time point, and a constant $h$ by its loss summed over the time points with trapezoidal weights; the candidates are $h = 0$ and $25$ geometric values in $[10^{-6}, 1]$.
The oracle selects a constant $h$ that minimizes the total variation error, over $21$ geometric values in $[10^{-5}, 1]$ refined by a bounded scalar search in $\log h$, and $h(t)$ that minimizes the velocity error of~\cref{eq:oracle}.

In the main text, the best oracle and the best data-driven KDE of each target are the rules with the lowest error averaged over the three values of $N$ among the oracle and among the data-driven rules of \cref{tab:oned-full,tab:twod-full}.

\section{Supplementary Tables}\label{sec:tables}

The tables follow the order of the main text.
\begin{itemize}
\item \Cref{tab:oned-full,tab:twod-full} list the estimators of \cref{sec:flowaccuracy}, from which \cref{tab:main} selects its rows, and \cref{tab:oned-full} also the convolution of \cref{sec:operator}.
\item \Cref{tab:tend,tab:refine-1d,tab:refine-2d,tab:integrator} show that stopping the flow, the mesh and the time integrator do not explain the accuracy of the regularized flows (\cref{sec:discretization}).
\item \Cref{tab:stability,tab:stability-refine} test the stability in the training data of \cref{thm:stability} and its dependence on the mesh (\cref{sec:stability}).
\item \Cref{tab:reg-full} compares the three regularizers (\cref{sec:operator}).
\item \Cref{tab:lam-sweep-full} compares the time-dependent and the constant regularization parameters (\cref{sec:lambda}).
\end{itemize}

\begin{table}[tbp]
\centering
\caption{Extended version of \cref{tab:main} for the one-dimensional targets, with the estimators of \cref{sec:flowaccuracy}, from which \cref{tab:main} selects the best data-driven KDE, and the convolution of the velocity that \cref{sec:operator} compares with the Hessian regularizer. Total variation error of the generated densities for the flows with oracle and hold-out $\lambda(t)$, all KDEs, and the velocity smoothed by convolution~\cite{gabriel2025kernelsmoothed} with a constant or a time-dependent bandwidth selected by hold-out or by the oracle (\cref{sec:kde}), mean $\pm$ standard error over five training sets. Shaded: rows of \cref{tab:main}.
Bold: the lowest data-driven mean and the data-driven entries not resolved from it (difference below twice the combined Monte Carlo or set-to-set standard error), in each column.}
\label{tab:oned-full}
\input{tables/tab_oned_full.tex}
\end{table}

\begin{table}[tbp]
\centering
\caption{Extended version of \cref{tab:main} for the two-dimensional targets, with the estimators of \cref{sec:flowaccuracy}, from which \cref{tab:main} selects the best data-driven KDE. Total variation error of the generated densities, mean $\pm$ standard error over five training sets. The anisotropic KDEs use the exact tangent of the curve and therefore exist only for the noisy circle and the two spirals; their data-driven column uses LSCV with the exact tangent and counts as an oracle. The Gaussian mixture has eight components with full covariances, fitted by EM, best training likelihood of $21$ starts. Shaded: the entries of \cref{tab:main} and the best oracle KDE. Bold: the lowest data-driven mean and the data-driven entries not resolved from it (difference below twice the combined Monte Carlo or set-to-set standard error), per column. The flow of the exact velocity $v_N$ is integrated by RK4 with $96$ steps. The limit $\lambda \to 0$ is computed at the floor $\lambda(t) = 10^{-14}\lambda_s$.}
\label{tab:twod-full}
\input{tables/tab_twod_full.tex}
\end{table}

\begin{table}[tbp]
\centering
\caption{Stopping the flow at $t_{\mathrm{end}} < 1$ regularizes the empirical velocity $v_N$, whose flow generates nearly the stopped law $\rho_N(\cdot, t_{\mathrm{end}})$ (\cref{sec:discretization}). As $t_{\mathrm{end}} \to 1$, the error of $v_N$ tends to the value $1$ of a memorizing flow, whereas the errors of the regularized flows change by at most $4 \cdot 10^{-4}$, below their Monte Carlo standard error, so stopping does not explain their accuracy. Total variation error against the end time $t_{\mathrm{end}}$ for $v_N$, the stopped law and the regularized flows, mean $\pm$ standard error over five training sets. The errors use independent latent samples and therefore differ from those of \cref{tab:main} by up to $0.004$. Beyond $t = 1 - 10^{-4}$, the regularized flows select $\lambda$ by the same rule at $40$ added times, with $1 - t$ from $10^{-4.1}$ to $10^{-8}$.}
\label{tab:tend}
\input{tables/tab_tend.tex}
\end{table}

\begin{table}[tbp]
\centering
\caption{Mesh refinement separates the regularization by the discretization from the explicit regularization (\cref{sec:discretization}); one-dimensional targets. Without explicit regularization and with the smallness regularizer, the error grows toward the value $1$ of a memorizing flow as $m$ grows, whereas that of the Hessian flow does not change. Total variation error against the spline resolution $m$, mean $\pm$ standard error over five training sets. The blocks $\nabla^k$ regularizer show the hold-out flows of the Hessian ($k = 2$), gradient ($k = 1$) and smallness ($k = 0$) regularizers. The blocks $\lambda \to 0$ show the limit computed at the floor $\lambda(t) = 10^{-14}\lambda_s$ of \cref{sec:solver} with independent latent samples, since the production latents alias with the spikes of a memorizing flow.}
\label{tab:refine-1d}
\input{tables/tab_refine_1d.tex}
\end{table}

\begin{table}[tbp]
\centering
\caption{As \cref{tab:refine-1d}, for the two-dimensional targets.}
\label{tab:refine-2d}
\input{tables/tab_refine_2d.tex}
\end{table}

\begin{table}[tbp]
\centering
\caption{This table checks that the RK4 integrator of the experiments solves the flow equation accurately, so that the integration error does not regularize the flows (\cref{sec:discretization}). Deviation of the RK4 samples $x$ and log-densities $\log f$ with $96$ steps at $t_{\mathrm{end}}$ from a reference, on $1000$ latent samples per training set. For the hold-out flow, the reference is the adaptive eighth-order Runge--Kutta method DOP853 of Dormand and Prince; for the exact velocity $v_N$, which jumps at the faces of the boxes of \cref{lem:boxes}, it is RK4 with the number of steps doubled from $96$ until two successive solutions differ by at most $10^{-6}$ in $x$ and by at most $10^{-4}$ in the mean of $\log f$ over the samples; for one training set of the two spirals with $N = 100$, the doubling stopped before reaching this tolerance. The total variation errors of the hold-out flows with RK4 and DOP853 differ by at most $2.6 \cdot 10^{-4}$ in one and $9 \cdot 10^{-4}$ in two dimensions. The deviation $\max|\Delta x|$ is the maximum over training sets and samples, in two dimensions of the Euclidean norm, and $|\Delta \log f|$ is the largest mean over the samples among the training sets. The two-dimensional flows use the Hessian regularizer with $m = 128$.}
\label{tab:integrator}
\input{tables/tab_integrator.tex}
\end{table}

\begin{table}[tbp]
\centering
\caption{This table tests the stability in the training data of \cref{thm:stability} (\cref{sec:stability}). A row with constant entries, and slope one, belongs to a distance proportional to $\varepsilon$. Entries: (a) $\mathrm{TV}(p_\varepsilon, p_0)/\varepsilon$ and (b) $\sup_t \|v_t - v_t'\|_{H^2(\Omega)}/\varepsilon$ against the perturbation $\varepsilon$ of every training point, mean over five training sets and two perturbation directions. Slope: least-squares fit of the logarithm of the unscaled distance against $\log\varepsilon$ for $\varepsilon \le 10^{-3}$.}
\label{tab:stability}
\input{tables/tab_stability.tex}
\end{table}

\begin{table}[tbp]
\centering
\caption{This table tests whether the stability constants of \cref{tab:stability} depend on the mesh (\cref{sec:stability}). For the smallness regularizer they grow under refinement, whereas the total variation constant of the Hessian regularizer with fixed $\lambda$ does not change. Stability constants in one dimension, $N = 100$, $\varepsilon \le 10^{-4}$: mean and range over five training sets, two perturbation directions and three values of $\varepsilon$, and the growth factor from $m = 256$ to $m = 1024$. Fixed $\lambda$: the hold-out selection at $m = 512$.}
\label{tab:stability-refine}
\input{tables/tab_stability_refine.tex}
\end{table}

\begin{table}[tbp]
\centering
\caption{This table compares the regularizers of order $k = 2$, $1$ and $0$ on every target, to test which property of the regularizer prevents memorization (\cref{sec:operator}). Total variation error of the flow for the Hessian ($k = 2$), gradient ($k = 1$) and smallness ($k = 0$) regularizers with oracle and hold-out $\lambda(t)$, mean $\pm$ standard error over five training sets. The Hessian rows are those of \cref{tab:main}.}
\label{tab:reg-full}
\input{tables/tab_reg_full.tex}
\end{table}

\begin{table}[tbp]
\centering
\caption{This table tests whether the time dependence of the regularization parameter matters, by comparing the hold-out selection $\lambda(t)$ with its best multiple and with constant parameters (\cref{sec:lambda}). Total variation error with $N = 1000$ for the three regularizers, mean $\pm$ standard error over five training sets: hold-out selection $\lambda(t)$, its best multiple $\alpha\,\lambda(t)$, and the best and the hold-out constant $\lambda(t) = \lambda_c$.}
\label{tab:lam-sweep-full}
\input{tables/tab_lam_sweep_full.tex}
\end{table}

\FloatBarrier

\end{document}

%% file: tables/tab_main.tex
\renewcommand{\arraystretch}{1.30}
\begin{tabular}{@{}clccc@{}}
\toprule
 & Estimator & $N = 100$ & $N = 1000$ & $N = 3000$ \\
 &  & ($10^{-2}$) & ($10^{-2}$) & ($10^{-2}$) \\
\midrule
\multirow{4}{*}{\rotatebox[origin=c]{90}{\begin{tabular}{@{}c@{}}Trimodal\\(1D)\end{tabular}}} & FM, oracle $\lambda$ & $10.0 \pm 0.8$ & $4.1 \pm 0.3$ & $2.5 \pm 0.2$ \\
 & FM, hold-out & $\mathbf{13.4 \pm 1.7}$ & $\mathbf{4.7 \pm 0.3}$ & $\mathbf{2.9 \pm 0.4}$ \\
 & FM, hold-out constant $\lambda$ & $\mathbf{13.0 \pm 1.3}$ & $4.9 \pm 0.2$ & $3.2 \pm 0.3$ \\
 & KDE, Sheather--Jones & $\mathbf{11.8 \pm 1.1}$ & $\mathbf{4.3 \pm 0.4}$ & $\mathbf{2.7 \pm 0.2}$ \\
\midrule
\multirow{4}{*}{\rotatebox[origin=c]{90}{\begin{tabular}{@{}c@{}}Inhom.\\(1D)\end{tabular}}} & FM, oracle $\lambda$ & $14.6 \pm 0.7$ & $5.9 \pm 0.3$ & $4.0 \pm 0.1$ \\
 & FM, hold-out & $\mathbf{16.9 \pm 1.5}$ & $\mathbf{6.7 \pm 0.3}$ & $\mathbf{4.3 \pm 0.1}$ \\
 & FM, hold-out constant $\lambda$ & $\mathbf{17.3 \pm 1.5}$ & $\mathbf{6.5 \pm 0.4}$ & $\mathbf{4.2 \pm 0.1}$ \\
 & 4th-order Abramson KDE, LSCV & $\mathbf{17.6 \pm 1.6}$ & $\mathbf{6.1 \pm 0.3}$ & $\mathbf{4.1 \pm 0.1}$ \\
\midrule
\multirow{4}{*}{\rotatebox[origin=c]{90}{\begin{tabular}{@{}c@{}}Noisy circle\\(2D)\end{tabular}}} & FM, oracle $\lambda$ & $18.2 \pm 0.8$ & $8.5 \pm 0.3$ & $5.5 \pm 0.3$ \\
 & FM, hold-out & $\mathbf{18.9 \pm 0.5}$ & $\mathbf{8.4 \pm 0.5}$ & $\mathbf{5.4 \pm 0.3}$ \\
 & FM, hold-out constant $\lambda$ & $22.2 \pm 0.6$ & $11.0 \pm 0.4$ & $7.4 \pm 0.3$ \\
 & 4th-order product KDE, LSCV & $27.2 \pm 0.7$ & $12.6 \pm 0.5$ & $8.9 \pm 0.2$ \\
\midrule
\multirow{4}{*}{\rotatebox[origin=c]{90}{\begin{tabular}{@{}c@{}}Checkerb.\\(2D)\end{tabular}}} & FM, oracle $\lambda$ & $36.7 \pm 0.4$ & $20.6 \pm 0.4$ & $16.1 \pm 0.2$ \\
 & FM, hold-out & $37.3 \pm 0.6$ & $\mathbf{20.4 \pm 0.3}$ & $\mathbf{16.2 \pm 0.2}$ \\
 & FM, hold-out constant $\lambda$ & $39.0 \pm 0.9$ & $22.0 \pm 0.3$ & $\mathbf{16.6 \pm 0.2}$ \\
 & 4th-order KDE, LSCV & $\mathbf{33.7 \pm 0.4}$ & $\mathbf{20.3 \pm 0.3}$ & $\mathbf{16.1 \pm 0.2}$ \\
\midrule
\multirow{4}{*}{\rotatebox[origin=c]{90}{\begin{tabular}{@{}c@{}}Two spirals\\(2D)\end{tabular}}} & FM, oracle $\lambda$ & $50.9 \pm 3.0$ & $20.9 \pm 0.3$ & $13.9 \pm 0.3$ \\
 & FM, hold-out & $\mathbf{57.2 \pm 5.1}$ & $\mathbf{20.7 \pm 0.3}$ & $\mathbf{13.6 \pm 0.3}$ \\
 & FM, hold-out constant $\lambda$ & $\mathbf{59.1 \pm 5.2}$ & $23.9 \pm 0.1$ & $17.1 \pm 0.2$ \\
 & KDE, LSCV & $\mathbf{54.5 \pm 1.0}$ & $26.8 \pm 0.3$ & $18.6 \pm 0.3$ \\
\midrule
\multirow{4}{*}{\rotatebox[origin=c]{90}{\begin{tabular}{@{}c@{}}8 Gaussians\\(2D)\end{tabular}}} & FM, oracle $\lambda$ & $25.6 \pm 1.2$ & $11.7 \pm 0.2$ & $8.4 \pm 0.2$ \\
 & FM, hold-out & $\mathbf{26.7 \pm 1.2}$ & $\mathbf{12.4 \pm 0.5}$ & $8.4 \pm 0.2$ \\
 & FM, hold-out constant $\lambda$ & $29.6 \pm 1.1$ & $15.3 \pm 0.7$ & $10.5 \pm 0.2$ \\
 & 4th-order KDE, LSCV & $\mathbf{25.2 \pm 1.8}$ & $\mathbf{11.9 \pm 0.1}$ & $\mathbf{7.9 \pm 0.2}$ \\
\bottomrule
\end{tabular}

%% file: tables/tab_oned_full.tex
\begin{tabular}{@{}clccc@{}}
\toprule
 & Estimator & $N = 100$ & $N = 1000$ & $N = 3000$ \\
 &  & ($10^{-2}$) & ($10^{-2}$) & ($10^{-2}$) \\
\midrule
\rowcolor{black!8}
\cellcolor{white} & FM, oracle $\lambda$ & $10.0 \pm 0.8$ & $4.1 \pm 0.3$ & $2.5 \pm 0.2$ \\
\rowcolor{black!8}
\cellcolor{white} & FM, hold-out & $\mathbf{13.4 \pm 1.7}$ & $\mathbf{4.7 \pm 0.3}$ & $\mathbf{2.9 \pm 0.4}$ \\
\addlinespace
 & KDE, oracle $\sigma$ & $10.8 \pm 1.1$ & $4.2 \pm 0.4$ & $2.6 \pm 0.2$ \\
 & Abramson, oracle $\sigma$ & $10.7 \pm 0.9$ & $4.5 \pm 0.4$ & $2.6 \pm 0.2$ \\
\rowcolor{black!8}
\cellcolor{white} & KDE, Sheather--Jones & $\mathbf{11.8 \pm 1.1}$ & $\mathbf{4.3 \pm 0.4}$ & $2.7 \pm 0.2$ \\
 & KDE, LSCV & $\mathbf{12.9 \pm 1.1}$ & $4.4 \pm 0.4$ & $2.7 \pm 0.2$ \\
 & Abramson, LSCV & $\mathbf{12.6 \pm 1.0}$ & $5.1 \pm 0.4$ & $2.7 \pm 0.1$ \\
 & KDE, Silverman & $16.6 \pm 0.9$ & $8.9 \pm 0.3$ & $6.3 \pm 0.2$ \\
\rowcolor{black!8}
\cellcolor{white} & 4th-order KDE, oracle $\sigma$ & $9.9 \pm 1.0$ & $3.8 \pm 0.6$ & $2.1 \pm 0.2$ \\
 & 4th-order KDE, LSCV & $\mathbf{12.6 \pm 1.4}$ & $\mathbf{3.9 \pm 0.6}$ & $\mathbf{2.3 \pm 0.2}$ \\
 & 4th-order Abramson KDE, oracle $\sigma$ & $10.0 \pm 0.8$ & $4.5 \pm 0.5$ & $2.6 \pm 0.1$ \\
 & 4th-order Abramson KDE, LSCV & $\mathbf{12.7 \pm 0.8}$ & $5.6 \pm 0.4$ & $3.1 \pm 0.1$ \\
\addlinespace
 & Gabriel et al., constant $h$, hold-out & $24.8 \pm 4.6$ & $5.9 \pm 0.5$ & $4.3 \pm 0.6$ \\
\rowcolor{black!8}
\cellcolor{white} & Gabriel et al., $h(t)$, hold-out & $23.4 \pm 4.6$ & $5.6 \pm 0.4$ & $3.6 \pm 0.4$ \\
\rowcolor{black!8}
\cellcolor{white} & Gabriel et al., constant $h$, oracle TV & $15.6 \pm 1.5$ & $4.6 \pm 0.2$ & $3.0 \pm 0.1$ \\
 & Gabriel et al., $h(t)$, oracle velocity error & $20.3 \pm 1.7$ & $4.9 \pm 0.3$ & $3.1 \pm 0.1$ \\
\addlinespace
 & exact $v_N$, reference & $98.8 \pm 0.0$ & $91.1 \pm 0.1$ & $79.7 \pm 0.1$ \\
 & exact $v_N$, RK4 (96 steps) & $98.8 \pm 0.0$ & $91.2 \pm 0.1$ & $79.7 \pm 0.1$ \\
\rowcolor{black!8}
\cellcolor{white}\multirow{-19}{*}[1.5\defaultaddspace]{\rotatebox[origin=c]{90}{\begin{tabular}{@{}c@{}}\rowcolor{white}Trimodal\\\rowcolor{white}(1D)\end{tabular}}} & FM, $\lambda \to 0$, $m = 512$ & $87.7 \pm 1.1$ & $25.5 \pm 0.3$ & $13.9 \pm 0.3$ \\
\midrule
\rowcolor{black!8}
\cellcolor{white} & FM, oracle $\lambda$ & $14.6 \pm 0.7$ & $5.9 \pm 0.3$ & $4.0 \pm 0.1$ \\
\rowcolor{black!8}
\cellcolor{white} & FM, hold-out & $\mathbf{16.9 \pm 1.5}$ & $\mathbf{6.7 \pm 0.3}$ & $\mathbf{4.3 \pm 0.1}$ \\
\addlinespace
 & KDE, oracle $\sigma$ & $17.9 \pm 1.5$ & $7.8 \pm 0.2$ & $5.2 \pm 0.2$ \\
 & Abramson, oracle $\sigma$ & $16.6 \pm 1.4$ & $6.3 \pm 0.2$ & $4.0 \pm 0.2$ \\
 & KDE, Sheather--Jones & $19.1 \pm 1.1$ & $8.4 \pm 0.2$ & $5.3 \pm 0.2$ \\
 & KDE, LSCV & $19.1 \pm 1.9$ & $8.3 \pm 0.2$ & $5.6 \pm 0.2$ \\
 & Abramson, LSCV & $\mathbf{17.2 \pm 1.6}$ & $\mathbf{6.8 \pm 0.4}$ & $\mathbf{4.2 \pm 0.1}$ \\
 & KDE, Silverman & $21.3 \pm 1.0$ & $14.7 \pm 0.4$ & $12.8 \pm 0.5$ \\
 & 4th-order KDE, oracle $\sigma$ & $18.2 \pm 1.4$ & $7.7 \pm 0.2$ & $4.8 \pm 0.2$ \\
 & 4th-order KDE, LSCV & $19.5 \pm 1.9$ & $7.9 \pm 0.3$ & $5.3 \pm 0.2$ \\
\rowcolor{black!8}
\cellcolor{white} & 4th-order Abramson KDE, oracle $\sigma$ & $17.1 \pm 1.4$ & $6.0 \pm 0.3$ & $3.7 \pm 0.1$ \\
\rowcolor{black!8}
\cellcolor{white} & 4th-order Abramson KDE, LSCV & $\mathbf{17.6 \pm 1.6}$ & $\mathbf{6.1 \pm 0.3}$ & $\mathbf{4.1 \pm 0.1}$ \\
\addlinespace
\rowcolor{black!8}
\cellcolor{white} & Gabriel et al., constant $h$, hold-out & $18.4 \pm 1.8$ & $6.9 \pm 0.4$ & $4.9 \pm 0.3$ \\
 & Gabriel et al., $h(t)$, hold-out & $18.8 \pm 1.4$ & $7.5 \pm 0.3$ & $5.0 \pm 0.3$ \\
\rowcolor{black!8}
\cellcolor{white} & Gabriel et al., constant $h$, oracle TV & $16.3 \pm 0.9$ & $6.7 \pm 0.4$ & $4.5 \pm 0.1$ \\
 & Gabriel et al., $h(t)$, oracle velocity error & $18.0 \pm 0.5$ & $7.0 \pm 0.3$ & $4.6 \pm 0.1$ \\
\addlinespace
 & exact $v_N$, reference & $98.3 \pm 0.0$ & $88.4 \pm 0.2$ & $74.3 \pm 0.2$ \\
 & exact $v_N$, RK4 (96 steps) & $98.3 \pm 0.0$ & $88.4 \pm 0.2$ & $74.3 \pm 0.2$ \\
\rowcolor{black!8}
\cellcolor{white}\multirow{-19}{*}[1.5\defaultaddspace]{\rotatebox[origin=c]{90}{\begin{tabular}{@{}c@{}}\rowcolor{white}Inhom.\\\rowcolor{white}(1D)\end{tabular}}} & FM, $\lambda \to 0$, $m = 512$ & $81.4 \pm 1.1$ & $24.0 \pm 0.6$ & $13.3 \pm 0.2$ \\
\bottomrule
\end{tabular}

%% file: tables/tab_twod_full.tex
\setlength{\tabcolsep}{3.5pt}%
\begin{tabular}{@{}lcccccc@{}}
\toprule
 & \multicolumn{2}{c}{$N = 100$} & \multicolumn{2}{c}{$N = 1000$} & \multicolumn{2}{c}{$N = 3000$} \\
\cmidrule(lr){2-3}\cmidrule(lr){4-5}\cmidrule(lr){6-7}
TV ($10^{-2}$) & data-driven & oracle & data-driven & oracle & data-driven & oracle \\
\midrule
\multicolumn{7}{l}{\emph{Noisy circle}} \\
FM & \cellcolor{black!8}$\mathbf{18.9 \pm 0.5}$ & \cellcolor{black!8}$18.2 \pm 0.8$ & \cellcolor{black!8}$\mathbf{8.4 \pm 0.5}$ & \cellcolor{black!8}$8.5 \pm 0.3$ & \cellcolor{black!8}$\mathbf{5.4 \pm 0.3}$ & \cellcolor{black!8}$5.5 \pm 0.3$ \\
\addlinespace
KDE & $28.1 \pm 0.8$ & $26.4 \pm 0.3$ & $13.7 \pm 0.5$ & $13.7 \pm 0.3$ & $9.7 \pm 0.3$ & $9.8 \pm 0.2$ \\
Abramson & $29.3 \pm 1.3$ & $26.4 \pm 0.6$ & $15.3 \pm 0.3$ & $14.3 \pm 0.3$ & $11.0 \pm 0.2$ & $10.2 \pm 0.3$ \\
4th-order KDE & $27.0 \pm 0.7$ & $25.8 \pm 0.5$ & $13.1 \pm 0.6$ & $12.5 \pm 0.4$ & $8.8 \pm 0.2$ & $8.4 \pm 0.2$ \\
4th-order product KDE & \cellcolor{black!8}$27.2 \pm 0.7$ & $25.8 \pm 0.4$ & \cellcolor{black!8}$12.6 \pm 0.5$ & $12.6 \pm 0.4$ & \cellcolor{black!8}$8.9 \pm 0.2$ & $8.7 \pm 0.3$ \\
aniso. KDE & $21.0 \pm 0.7$ & $19.6 \pm 0.4$ & $10.8 \pm 0.3$ & $10.5 \pm 0.3$ & $7.6 \pm 0.3$ & $7.6 \pm 0.3$ \\
4th-order aniso. KDE & $19.7 \pm 0.6$ & \cellcolor{black!8}$19.1 \pm 0.3$ & $9.8 \pm 0.2$ & \cellcolor{black!8}$9.2 \pm 0.4$ & $6.6 \pm 0.4$ & \cellcolor{black!8}$6.4 \pm 0.3$ \\
\addlinespace
exact $v_N$ & \multicolumn{2}{c}{$100.0 \pm 0.0$} & \multicolumn{2}{c}{$100.0 \pm 0.0$} & \multicolumn{2}{c}{$100.0 \pm 0.0$} \\
FM, $\lambda \to 0$, $m = 128$ & \multicolumn{2}{c}{\cellcolor{black!8}$99.8 \pm 0.0$} & \multicolumn{2}{c}{\cellcolor{black!8}$72.1 \pm 0.4$} & \multicolumn{2}{c}{\cellcolor{black!8}$40.5 \pm 0.2$} \\
\midrule
\multicolumn{7}{l}{\emph{Checkerboard}} \\
FM & \cellcolor{black!8}$37.3 \pm 0.6$ & \cellcolor{black!8}$36.7 \pm 0.4$ & \cellcolor{black!8}$\mathbf{20.4 \pm 0.3}$ & \cellcolor{black!8}$20.6 \pm 0.4$ & \cellcolor{black!8}$\mathbf{16.2 \pm 0.2}$ & \cellcolor{black!8}$16.1 \pm 0.2$ \\
\addlinespace
KDE & $35.5 \pm 0.6$ & $33.7 \pm 0.6$ & $\mathbf{20.9 \pm 0.4}$ & $20.5 \pm 0.3$ & $\mathbf{16.2 \pm 0.1}$ & $16.3 \pm 0.2$ \\
Abramson & $38.7 \pm 0.8$ & $36.8 \pm 0.2$ & $23.8 \pm 0.5$ & $23.2 \pm 0.2$ & $19.1 \pm 0.1$ & $18.1 \pm 0.4$ \\
4th-order KDE & \cellcolor{black!8}$\mathbf{33.7 \pm 0.4}$ & \cellcolor{black!8}$32.1 \pm 0.2$ & \cellcolor{black!8}$\mathbf{20.3 \pm 0.3}$ & \cellcolor{black!8}$20.0 \pm 0.2$ & \cellcolor{black!8}$\mathbf{16.1 \pm 0.2}$ & \cellcolor{black!8}$15.9 \pm 0.2$ \\
4th-order product KDE & $\mathbf{33.1 \pm 0.5}$ & $32.1 \pm 0.2$ & $21.2 \pm 0.1$ & $20.2 \pm 0.4$ & $\mathbf{16.6 \pm 0.2}$ & $16.2 \pm 0.3$ \\
\addlinespace
exact $v_N$ & \multicolumn{2}{c}{$100.0 \pm 0.0$} & \multicolumn{2}{c}{$100.0 \pm 0.0$} & \multicolumn{2}{c}{$100.0 \pm 0.0$} \\
FM, $\lambda \to 0$, $m = 128$ & \multicolumn{2}{c}{\cellcolor{black!8}$99.9 \pm 0.0$} & \multicolumn{2}{c}{\cellcolor{black!8}$99.6 \pm 0.0$} & \multicolumn{2}{c}{\cellcolor{black!8}$94.1 \pm 0.1$} \\
\midrule
\multicolumn{7}{l}{\emph{Two spirals}} \\
FM & \cellcolor{black!8}$\mathbf{57.2 \pm 5.1}$ & \cellcolor{black!8}$50.9 \pm 3.0$ & \cellcolor{black!8}$\mathbf{20.7 \pm 0.3}$ & \cellcolor{black!8}$20.9 \pm 0.3$ & \cellcolor{black!8}$\mathbf{13.6 \pm 0.3}$ & \cellcolor{black!8}$13.9 \pm 0.3$ \\
\addlinespace
KDE & \cellcolor{black!8}$\mathbf{54.5 \pm 1.0}$ & $49.2 \pm 1.0$ & \cellcolor{black!8}$26.8 \pm 0.3$ & $25.3 \pm 0.3$ & \cellcolor{black!8}$18.6 \pm 0.3$ & $18.0 \pm 0.2$ \\
Abramson & $56.4 \pm 1.3$ & $51.4 \pm 1.2$ & $28.7 \pm 0.4$ & $26.9 \pm 0.4$ & $19.7 \pm 0.3$ & $19.3 \pm 0.2$ \\
4th-order KDE & $57.2 \pm 0.9$ & $49.4 \pm 0.7$ & $25.8 \pm 0.3$ & $24.9 \pm 0.2$ & $17.6 \pm 0.3$ & $17.2 \pm 0.2$ \\
4th-order product KDE & $57.4 \pm 0.7$ & $49.6 \pm 1.1$ & $25.5 \pm 0.4$ & $25.4 \pm 0.3$ & $17.5 \pm 0.3$ & $17.2 \pm 0.3$ \\
aniso. KDE & $38.9 \pm 1.1$ & $36.0 \pm 1.2$ & $17.2 \pm 0.5$ & $17.1 \pm 0.3$ & $11.5 \pm 0.1$ & $11.4 \pm 0.1$ \\
4th-order aniso. KDE & $39.5 \pm 1.5$ & \cellcolor{black!8}$35.4 \pm 1.1$ & $16.8 \pm 0.4$ & \cellcolor{black!8}$16.4 \pm 0.3$ & $10.8 \pm 0.1$ & \cellcolor{black!8}$10.8 \pm 0.1$ \\
\addlinespace
exact $v_N$ & \multicolumn{2}{c}{$100.0 \pm 0.0$} & \multicolumn{2}{c}{$100.0 \pm 0.0$} & \multicolumn{2}{c}{$100.0 \pm 0.0$} \\
FM, $\lambda \to 0$, $m = 128$ & \multicolumn{2}{c}{\cellcolor{black!8}$99.9 \pm 0.0$} & \multicolumn{2}{c}{\cellcolor{black!8}$95.0 \pm 0.1$} & \multicolumn{2}{c}{\cellcolor{black!8}$56.9 \pm 0.5$} \\
\midrule
\multicolumn{7}{l}{\emph{8 Gaussians}} \\
FM & \cellcolor{black!8}$\mathbf{26.7 \pm 1.2}$ & \cellcolor{black!8}$25.6 \pm 1.2$ & \cellcolor{black!8}$\mathbf{12.4 \pm 0.5}$ & \cellcolor{black!8}$11.7 \pm 0.2$ & \cellcolor{black!8}$8.4 \pm 0.2$ & \cellcolor{black!8}$8.4 \pm 0.2$ \\
\addlinespace
KDE & $26.6 \pm 1.6$ & $25.0 \pm 1.5$ & $12.8 \pm 0.1$ & $12.9 \pm 0.4$ & $9.3 \pm 0.2$ & $9.1 \pm 0.2$ \\
Abramson & $27.3 \pm 1.6$ & $25.0 \pm 1.7$ & $13.7 \pm 0.1$ & $12.9 \pm 0.3$ & $9.7 \pm 0.2$ & $9.2 \pm 0.3$ \\
4th-order KDE & \cellcolor{black!8}$\mathbf{25.2 \pm 1.8}$ & $24.0 \pm 1.4$ & \cellcolor{black!8}$\mathbf{11.9 \pm 0.1}$ & $11.5 \pm 0.2$ & \cellcolor{black!8}$\mathbf{7.9 \pm 0.2}$ & $8.0 \pm 0.2$ \\
4th-order product KDE & $\mathbf{25.5 \pm 1.5}$ & \cellcolor{black!8}$24.0 \pm 1.7$ & $\mathbf{11.9 \pm 0.1}$ & \cellcolor{black!8}$11.6 \pm 0.1$ & $\mathbf{7.8 \pm 0.2}$ & \cellcolor{black!8}$7.8 \pm 0.2$ \\
\addlinespace
Gaussian mixture, EM & \multicolumn{2}{c}{$24.3 \pm 2.1$} & \multicolumn{2}{c}{$7.6 \pm 0.3$} & \multicolumn{2}{c}{$4.7 \pm 0.2$} \\
\addlinespace
exact $v_N$ & \multicolumn{2}{c}{$100.0 \pm 0.0$} & \multicolumn{2}{c}{$100.0 \pm 0.0$} & \multicolumn{2}{c}{$100.0 \pm 0.0$} \\
FM, $\lambda \to 0$, $m = 128$ & \multicolumn{2}{c}{\cellcolor{black!8}$99.9 \pm 0.0$} & \multicolumn{2}{c}{\cellcolor{black!8}$93.5 \pm 0.3$} & \multicolumn{2}{c}{\cellcolor{black!8}$58.1 \pm 0.4$} \\
\bottomrule
\end{tabular}

%% file: tables/tab_tend.tex
\begin{tabular}{lcccccc}
\toprule
Estimator & \multicolumn{2}{c}{$N = 100$} & \multicolumn{2}{c}{$N = 1000$} & \multicolumn{2}{c}{$N = 3000$} \\
\cmidrule(lr){2-3}\cmidrule(lr){4-5}\cmidrule(lr){6-7}
$1 - t_{\mathrm{end}}$ & $10^{-4}$ & $10^{-8}$ & $10^{-4}$ & $10^{-8}$ & $10^{-4}$ & $10^{-8}$ \\
 & ($10^{-2}$) & ($10^{-2}$) & ($10^{-2}$) & ($10^{-2}$) & ($10^{-2}$) & ($10^{-2}$) \\
\midrule
\multicolumn{7}{l}{\emph{Trimodal}} \\
exact $v_N$, reference & $98.8 \pm 0.0$ & $100.0 \pm 0.0$ & $91.1 \pm 0.1$ & $100.0 \pm 0.0$ & $79.7 \pm 0.1$ & $100.0 \pm 0.0$ \\
law $\rho_N(\cdot, t_{\mathrm{end}})$ & $98.9 \pm 0.0$ & $100.0 \pm 0.0$ & $91.3 \pm 0.1$ & $100.0 \pm 0.0$ & $79.7 \pm 0.0$ & $100.0 \pm 0.0$ \\
FM, oracle $\lambda(t)$ & $9.9 \pm 0.8$ & $9.9 \pm 0.8$ & $4.2 \pm 0.3$ & $4.2 \pm 0.3$ & $2.5 \pm 0.2$ & $2.5 \pm 0.2$ \\
FM, hold-out $\lambda(t)$ & $13.2 \pm 1.7$ & $13.2 \pm 1.7$ & $4.7 \pm 0.3$ & $4.7 \pm 0.3$ & $2.9 \pm 0.4$ & $2.9 \pm 0.4$ \\
\midrule
\multicolumn{7}{l}{\emph{Inhomogeneous}} \\
exact $v_N$, reference & $98.3 \pm 0.0$ & $100.0 \pm 0.0$ & $88.4 \pm 0.2$ & $100.0 \pm 0.0$ & $74.3 \pm 0.2$ & $100.0 \pm 0.0$ \\
law $\rho_N(\cdot, t_{\mathrm{end}})$ & $98.3 \pm 0.0$ & $100.0 \pm 0.0$ & $88.4 \pm 0.2$ & $100.0 \pm 0.0$ & $74.1 \pm 0.1$ & $100.0 \pm 0.0$ \\
FM, oracle $\lambda(t)$ & $14.8 \pm 0.7$ & $14.8 \pm 0.7$ & $5.9 \pm 0.3$ & $5.9 \pm 0.3$ & $4.0 \pm 0.1$ & $4.0 \pm 0.1$ \\
FM, hold-out $\lambda(t)$ & $16.8 \pm 1.5$ & $16.8 \pm 1.5$ & $6.7 \pm 0.3$ & $6.7 \pm 0.3$ & $4.3 \pm 0.1$ & $4.3 \pm 0.1$ \\
\midrule
\multicolumn{7}{l}{\emph{Noisy circle (2D)}} \\
exact $v_N$, reference & $100.0 \pm 0.0$ & $100.0 \pm 0.0$ & $100.0 \pm 0.0$ & $100.0 \pm 0.0$ & $100.0 \pm 0.0$ & $100.0 \pm 0.0$ \\
law $\rho_N(\cdot, t_{\mathrm{end}})$ & $100.0 \pm 0.0$ & $100.0 \pm 0.0$ & $100.0 \pm 0.0$ & $100.0 \pm 0.0$ & $100.0 \pm 0.0$ & $100.0 \pm 0.0$ \\
FM, oracle $\lambda(t)$ & $18.3 \pm 0.7$ & $18.3 \pm 0.7$ & $8.5 \pm 0.3$ & $8.5 \pm 0.3$ & $5.5 \pm 0.2$ & $5.5 \pm 0.2$ \\
FM, hold-out $\lambda(t)$ & $18.9 \pm 0.4$ & $18.9 \pm 0.4$ & $8.4 \pm 0.5$ & $8.5 \pm 0.5$ & $5.4 \pm 0.2$ & $5.4 \pm 0.2$ \\
\midrule
\multicolumn{7}{l}{\emph{Checkerboard (2D)}} \\
exact $v_N$, reference & $100.0 \pm 0.0$ & $100.0 \pm 0.0$ & $100.0 \pm 0.0$ & $100.0 \pm 0.0$ & $100.0 \pm 0.0$ & $100.0 \pm 0.0$ \\
law $\rho_N(\cdot, t_{\mathrm{end}})$ & $100.0 \pm 0.0$ & $100.0 \pm 0.0$ & $100.0 \pm 0.0$ & $100.0 \pm 0.0$ & $100.0 \pm 0.0$ & $100.0 \pm 0.0$ \\
FM, oracle $\lambda(t)$ & $37.1 \pm 0.5$ & $37.1 \pm 0.5$ & $20.4 \pm 0.3$ & $20.4 \pm 0.3$ & $16.1 \pm 0.1$ & $16.2 \pm 0.1$ \\
FM, hold-out $\lambda(t)$ & $37.3 \pm 0.6$ & $37.3 \pm 0.6$ & $20.4 \pm 0.3$ & $20.4 \pm 0.3$ & $16.2 \pm 0.2$ & $16.2 \pm 0.2$ \\
\midrule
\multicolumn{7}{l}{\emph{Two spirals (2D)}} \\
exact $v_N$, reference & $100.0 \pm 0.0$ & $100.0 \pm 0.0$ & $100.0 \pm 0.0$ & $100.0 \pm 0.0$ & $100.0 \pm 0.0$ & $100.0 \pm 0.0$ \\
law $\rho_N(\cdot, t_{\mathrm{end}})$ & $100.0 \pm 0.0$ & $100.0 \pm 0.0$ & $100.0 \pm 0.0$ & $100.0 \pm 0.0$ & $100.0 \pm 0.0$ & $100.0 \pm 0.0$ \\
FM, oracle $\lambda(t)$ & $51.1 \pm 2.9$ & $51.1 \pm 2.9$ & $20.7 \pm 0.2$ & $20.7 \pm 0.2$ & $13.8 \pm 0.4$ & $13.8 \pm 0.4$ \\
FM, hold-out $\lambda(t)$ & $56.9 \pm 4.8$ & $56.9 \pm 4.8$ & $20.4 \pm 0.3$ & $20.4 \pm 0.3$ & $13.5 \pm 0.4$ & $13.5 \pm 0.4$ \\
\midrule
\multicolumn{7}{l}{\emph{8 Gaussians (2D)}} \\
exact $v_N$, reference & $100.0 \pm 0.0$ & $100.0 \pm 0.0$ & $100.0 \pm 0.0$ & $100.0 \pm 0.0$ & $100.0 \pm 0.0$ & $100.0 \pm 0.0$ \\
law $\rho_N(\cdot, t_{\mathrm{end}})$ & $100.0 \pm 0.0$ & $100.0 \pm 0.0$ & $100.0 \pm 0.0$ & $100.0 \pm 0.0$ & $100.0 \pm 0.0$ & $100.0 \pm 0.0$ \\
FM, oracle $\lambda(t)$ & $25.6 \pm 1.1$ & $25.6 \pm 1.1$ & $11.6 \pm 0.2$ & $11.6 \pm 0.2$ & $8.4 \pm 0.2$ & $8.4 \pm 0.2$ \\
FM, hold-out $\lambda(t)$ & $26.7 \pm 1.2$ & $26.7 \pm 1.2$ & $12.3 \pm 0.5$ & $12.3 \pm 0.5$ & $8.4 \pm 0.2$ & $8.4 \pm 0.2$ \\
\bottomrule
\end{tabular}

%% file: tables/tab_refine_1d.tex
\setlength{\tabcolsep}{5.5pt}%
\begin{tabular}{@{}llccc@{\hspace{1.5em}}llccc@{}}
\toprule
$k$ & $m$ & $N = 100$ & $N = 1000$ & $N = 3000$ & $k$ & $m$ & $N = 100$ & $N = 1000$ & $N = 3000$ \\
 &  & ($10^{-2}$) & ($10^{-2}$) & ($10^{-2}$) &  &  & ($10^{-2}$) & ($10^{-2}$) & ($10^{-2}$) \\
\midrule
\multicolumn{5}{l}{\emph{Trimodal, $\nabla^k$ regularizer}} & \multicolumn{5}{l}{\emph{Inhomogeneous, $\nabla^k$ regularizer}} \\
$2$ & 256 & $13.0 \pm 1.8$ & $4.7 \pm 0.4$ & $3.0 \pm 0.5$ & $2$ & 256 & $16.8 \pm 1.5$ & $6.6 \pm 0.3$ & $4.3 \pm 0.1$ \\
 & 512 & $13.4 \pm 1.7$ & $4.7 \pm 0.3$ & $2.9 \pm 0.4$ &  & 512 & $16.9 \pm 1.5$ & $6.7 \pm 0.3$ & $4.3 \pm 0.1$ \\
 & 1024 & $13.0 \pm 1.7$ & $4.8 \pm 0.4$ & $2.9 \pm 0.4$ &  & 1024 & $16.8 \pm 1.4$ & $6.7 \pm 0.3$ & $4.3 \pm 0.1$ \\
\addlinespace
$1$ & 256 & $15.4 \pm 1.9$ & $4.9 \pm 0.3$ & $3.5 \pm 0.3$ & $1$ & 256 & $16.0 \pm 1.3$ & $7.5 \pm 0.4$ & $4.6 \pm 0.2$ \\
 & 512 & $15.1 \pm 1.9$ & $5.0 \pm 0.2$ & $3.6 \pm 0.3$ &  & 512 & $15.9 \pm 1.3$ & $7.5 \pm 0.4$ & $4.6 \pm 0.1$ \\
 & 1024 & $14.8 \pm 1.8$ & $4.9 \pm 0.2$ & $3.6 \pm 0.3$ &  & 1024 & $15.8 \pm 1.3$ & $7.5 \pm 0.4$ & $4.7 \pm 0.1$ \\
\addlinespace
$0$ & 256 & $50.2 \pm 1.7$ & $16.7 \pm 0.3$ & $9.7 \pm 0.2$ & $0$ & 256 & $38.8 \pm 2.5$ & $16.3 \pm 0.6$ & $9.5 \pm 0.3$ \\
 & 512 & $57.5 \pm 2.7$ & $24.1 \pm 0.2$ & $13.9 \pm 0.2$ &  & 512 & $51.6 \pm 1.5$ & $23.0 \pm 0.4$ & $13.0 \pm 0.2$ \\
 & 1024 & $59.9 \pm 3.2$ & $34.2 \pm 0.5$ & $20.0 \pm 0.2$ &  & 1024 & $59.4 \pm 1.5$ & $30.6 \pm 0.6$ & $18.7 \pm 0.2$ \\
\addlinespace
\multicolumn{5}{l}{\emph{Trimodal, $\lambda \to 0$}} & \multicolumn{5}{l}{\emph{Inhomogeneous, $\lambda \to 0$}} \\
 & 256 & $71.0 \pm 1.3$ & $17.0 \pm 0.4$ & $9.7 \pm 0.2$ &  & 256 & $65.3 \pm 1.1$ & $16.6 \pm 0.8$ & $9.7 \pm 0.3$ \\
 & 512 & $87.7 \pm 1.1$ & $25.5 \pm 0.3$ & $13.9 \pm 0.3$ &  & 512 & $81.4 \pm 1.1$ & $24.0 \pm 0.6$ & $13.3 \pm 0.2$ \\
 & 1024 & $94.8 \pm 0.6$ & $39.4 \pm 0.7$ & $20.3 \pm 0.2$ &  & 1024 & $89.5 \pm 1.0$ & $36.0 \pm 0.7$ & $19.4 \pm 0.2$ \\
\bottomrule
\end{tabular}

%% file: tables/tab_refine_2d.tex
\setlength{\tabcolsep}{5.5pt}%
\begin{tabular}{@{}llccc@{\hspace{1.5em}}llccc@{}}
\toprule
$k$ & $m$ & $N = 100$ & $N = 1000$ & $N = 3000$ & $k$ & $m$ & $N = 100$ & $N = 1000$ & $N = 3000$ \\
 &  & ($10^{-2}$) & ($10^{-2}$) & ($10^{-2}$) &  &  & ($10^{-2}$) & ($10^{-2}$) & ($10^{-2}$) \\
\midrule
\multicolumn{5}{l}{\emph{Noisy circle, $\nabla^k$ regularizer}} & \multicolumn{5}{l}{\emph{Checkerboard, $\nabla^k$ regularizer}} \\
$2$ & 64 & $19.4 \pm 0.6$ & $8.5 \pm 0.5$ & $5.5 \pm 0.3$ & $2$ & 64 & $37.3 \pm 0.5$ & $20.4 \pm 0.4$ & $16.6 \pm 0.3$ \\
 & 128 & $18.9 \pm 0.5$ & $8.4 \pm 0.5$ & $5.4 \pm 0.3$ &  & 128 & $37.3 \pm 0.6$ & $20.4 \pm 0.3$ & $16.2 \pm 0.2$ \\
 & 256 & $18.7 \pm 0.5$ & $8.6 \pm 0.6$ & $5.4 \pm 0.3$ &  & 256 & $37.4 \pm 0.6$ & $20.5 \pm 0.4$ & $16.4 \pm 0.2$ \\
 & 512 & $19.0 \pm 0.6$ & $8.5 \pm 0.5$ & $5.5 \pm 0.3$ &  & 512 & $37.1 \pm 0.5$ & $20.5 \pm 0.4$ & $16.2 \pm 0.2$ \\
\addlinespace
$1$ & 64 & $26.3 \pm 0.5$ & $13.4 \pm 0.4$ & $9.0 \pm 0.3$ & $1$ & 64 & $38.4 \pm 0.6$ & $22.4 \pm 0.3$ & $17.9 \pm 0.3$ \\
 & 128 & $27.3 \pm 0.5$ & $14.3 \pm 0.3$ & $9.4 \pm 0.3$ &  & 128 & $38.6 \pm 0.6$ & $22.5 \pm 0.5$ & $18.0 \pm 0.4$ \\
 & 256 & $27.7 \pm 0.4$ & $14.5 \pm 0.5$ & $9.7 \pm 0.3$ &  & 256 & $38.6 \pm 0.5$ & $22.5 \pm 0.5$ & $17.8 \pm 0.2$ \\
 & 512 & $27.9 \pm 0.7$ & $14.2 \pm 0.6$ & $9.9 \pm 0.4$ &  & 512 & $38.8 \pm 0.3$ & $22.4 \pm 0.5$ & $18.0 \pm 0.2$ \\
\addlinespace
$0$ & 64 & $69.3 \pm 0.6$ & $30.6 \pm 0.3$ & $18.4 \pm 0.3$ & $0$ & 64 & $86.7 \pm 1.1$ & $56.0 \pm 0.6$ & $34.2 \pm 0.3$ \\
 & 128 & $85.1 \pm 0.9$ & $52.3 \pm 0.1$ & $33.5 \pm 0.2$ &  & 128 & $90.7 \pm 0.9$ & $77.9 \pm 1.0$ & $61.2 \pm 0.4$ \\
 & 256 & $91.7 \pm 1.0$ & $73.4 \pm 0.2$ & $57.8 \pm 0.2$ &  & 256 & $92.3 \pm 1.5$ & $89.5 \pm 0.6$ & $80.7 \pm 0.3$ \\
 & 512 & $94.8 \pm 0.3$ & $88.3 \pm 0.5$ & $77.4 \pm 0.5$ &  & 512 & $93.2 \pm 1.8$ & $93.6 \pm 0.8$ & $90.7 \pm 0.4$ \\
\addlinespace
\multicolumn{5}{l}{\emph{Noisy circle, $\lambda \to 0$}} & \multicolumn{5}{l}{\emph{Checkerboard, $\lambda \to 0$}} \\
 & 64 & $98.5 \pm 0.2$ & $37.7 \pm 0.6$ & $20.8 \pm 0.2$ &  & 64 & $99.9 \pm 0.0$ & $89.4 \pm 0.1$ & $40.1 \pm 0.2$ \\
 & 128 & $99.8 \pm 0.0$ & $72.1 \pm 0.4$ & $40.5 \pm 0.2$ &  & 128 & $99.9 \pm 0.0$ & $99.6 \pm 0.0$ & $94.1 \pm 0.1$ \\
 & 256 & $99.9 \pm 0.0$ & $98.3 \pm 0.1$ & $79.0 \pm 0.2$ &  & 256 & $100.0 \pm 0.0$ & $99.8 \pm 0.0$ & $99.5 \pm 0.0$ \\
 & 512 & $100.0 \pm 0.0$ & $99.7 \pm 0.0$ & $98.5 \pm 0.1$ &  & 512 & $100.0 \pm 0.0$ & $99.9 \pm 0.0$ & $99.8 \pm 0.0$ \\
\midrule
\multicolumn{5}{l}{\emph{Two spirals, $\nabla^k$ regularizer}} & \multicolumn{5}{l}{\emph{8 Gaussians, $\nabla^k$ regularizer}} \\
$2$ & 64 & $57.3 \pm 5.1$ & $20.7 \pm 0.3$ & $13.6 \pm 0.3$ & $2$ & 64 & $26.7 \pm 1.2$ & $12.4 \pm 0.4$ & $8.4 \pm 0.3$ \\
 & 128 & $57.2 \pm 5.1$ & $20.7 \pm 0.3$ & $13.6 \pm 0.3$ &  & 128 & $26.7 \pm 1.2$ & $12.4 \pm 0.5$ & $8.4 \pm 0.2$ \\
 & 256 & $57.3 \pm 5.1$ & $21.2 \pm 0.3$ & $13.5 \pm 0.2$ &  & 256 & $26.9 \pm 1.0$ & $12.3 \pm 0.5$ & $8.4 \pm 0.3$ \\
 & 512 & $57.0 \pm 5.0$ & $21.4 \pm 0.3$ & $13.7 \pm 0.2$ &  & 512 & $26.6 \pm 1.3$ & $12.3 \pm 0.5$ & $8.5 \pm 0.3$ \\
\addlinespace
$1$ & 64 & $55.9 \pm 1.3$ & $25.3 \pm 0.4$ & $16.8 \pm 0.2$ & $1$ & 64 & $31.0 \pm 1.1$ & $14.3 \pm 0.3$ & $10.3 \pm 0.3$ \\
 & 128 & $57.5 \pm 0.9$ & $26.3 \pm 0.2$ & $18.6 \pm 0.2$ &  & 128 & $30.9 \pm 1.5$ & $14.8 \pm 0.3$ & $10.5 \pm 0.3$ \\
 & 256 & $57.4 \pm 1.2$ & $27.1 \pm 0.4$ & $19.1 \pm 0.2$ &  & 256 & $31.2 \pm 1.6$ & $15.4 \pm 0.3$ & $10.3 \pm 0.2$ \\
 & 512 & $57.2 \pm 1.1$ & $26.8 \pm 0.5$ & $19.2 \pm 0.4$ &  & 512 & $30.9 \pm 1.4$ & $15.2 \pm 0.4$ & $10.6 \pm 0.1$ \\
\addlinespace
$0$ & 64 & $79.8 \pm 0.9$ & $39.8 \pm 0.3$ & $23.9 \pm 0.4$ & $0$ & 64 & $74.0 \pm 1.0$ & $40.9 \pm 0.5$ & $24.8 \pm 0.3$ \\
 & 128 & $86.8 \pm 1.2$ & $60.2 \pm 0.8$ & $40.5 \pm 0.3$ &  & 128 & $83.5 \pm 1.0$ & $62.0 \pm 0.8$ & $45.4 \pm 0.4$ \\
 & 256 & $91.9 \pm 1.3$ & $76.3 \pm 0.3$ & $60.8 \pm 0.4$ &  & 256 & $87.3 \pm 1.4$ & $79.3 \pm 0.7$ & $66.1 \pm 0.6$ \\
 & 512 & $93.0 \pm 1.7$ & $86.7 \pm 0.8$ & $78.1 \pm 0.3$ &  & 512 & $88.3 \pm 1.6$ & $86.7 \pm 0.3$ & $81.9 \pm 0.6$ \\
\addlinespace
\multicolumn{5}{l}{\emph{Two spirals, $\lambda \to 0$}} & \multicolumn{5}{l}{\emph{8 Gaussians, $\lambda \to 0$}} \\
 & 64 & $99.8 \pm 0.0$ & $67.4 \pm 0.5$ & $30.1 \pm 0.3$ &  & 64 & $99.6 \pm 0.0$ & $53.1 \pm 0.7$ & $29.0 \pm 0.3$ \\
 & 128 & $99.9 \pm 0.0$ & $95.0 \pm 0.1$ & $56.9 \pm 0.5$ &  & 128 & $99.9 \pm 0.0$ & $93.5 \pm 0.3$ & $58.1 \pm 0.4$ \\
 & 256 & $100.0 \pm 0.0$ & $99.4 \pm 0.0$ & $94.2 \pm 0.2$ &  & 256 & $100.0 \pm 0.0$ & $99.6 \pm 0.0$ & $95.7 \pm 0.2$ \\
 & 512 & $100.0 \pm 0.0$ & $99.8 \pm 0.0$ & $99.2 \pm 0.0$ &  & 512 & $100.0 \pm 0.0$ & $99.9 \pm 0.0$ & $99.5 \pm 0.1$ \\
\bottomrule
\end{tabular}

%% file: tables/tab_integrator.tex
\begin{tabular}{llcccccc}
\toprule
 & & \multicolumn{2}{c}{$N = 100$} & \multicolumn{2}{c}{$N = 1000$} & \multicolumn{2}{c}{$N = 3000$} \\
\cmidrule(lr){3-4}\cmidrule(lr){5-6}\cmidrule(lr){7-8}
Flow & RK4 steps & $\max |\Delta x|$ & $|\Delta \log f|$ & $\max |\Delta x|$ & $|\Delta \log f|$ & $\max |\Delta x|$ & $|\Delta \log f|$ \\
\midrule
\multicolumn{8}{l}{\emph{Trimodal (1D)}} \\
exact $v_N$ & 96 & $4.3 \cdot 10^{-6}$ & $1.2 \cdot 10^{-3}$ & $4.3 \cdot 10^{-3}$ & $6.6 \cdot 10^{-3}$ & $1.5 \cdot 10^{-3}$ & $8.9 \cdot 10^{-3}$ \\
 & 384 & $5.2 \cdot 10^{-8}$ & $1.9 \cdot 10^{-5}$ & $1.5 \cdot 10^{-7}$ & $4.6 \cdot 10^{-5}$ & $1.2 \cdot 10^{-7}$ & $5.6 \cdot 10^{-5}$ \\
 & 1536 & $1.3 \cdot 10^{-8}$ & $7.2 \cdot 10^{-6}$ & $6.1 \cdot 10^{-8}$ & $1.9 \cdot 10^{-5}$ & $4.5 \cdot 10^{-8}$ & $1.8 \cdot 10^{-5}$ \\
\addlinespace
FM, hold-out & 96 & $3.5 \cdot 10^{-3}$ & $7.0 \cdot 10^{-3}$ & $4.0 \cdot 10^{-4}$ & $1.5 \cdot 10^{-3}$ & $2.1 \cdot 10^{-4}$ & $3.6 \cdot 10^{-4}$ \\
 & 384 & $3.1 \cdot 10^{-4}$ & $5.3 \cdot 10^{-4}$ & $2.0 \cdot 10^{-5}$ & $5.0 \cdot 10^{-5}$ & $2.4 \cdot 10^{-6}$ & $5.8 \cdot 10^{-6}$ \\
 & 1536 & $1.5 \cdot 10^{-5}$ & $2.7 \cdot 10^{-5}$ & $7.9 \cdot 10^{-6}$ & $8.4 \cdot 10^{-6}$ & $4.2 \cdot 10^{-7}$ & $3.9 \cdot 10^{-7}$ \\
\midrule
\multicolumn{8}{l}{\emph{Inhomogeneous (1D)}} \\
exact $v_N$ & 96 & $2.3 \cdot 10^{-6}$ & $7.7 \cdot 10^{-4}$ & $1.1 \cdot 10^{-4}$ & $3.0 \cdot 10^{-3}$ & $1.5 \cdot 10^{-3}$ & $6.4 \cdot 10^{-3}$ \\
 & 384 & $0$ & $0$ & $7.9 \cdot 10^{-8}$ & $3.2 \cdot 10^{-5}$ & $7.0 \cdot 10^{-8}$ & $3.2 \cdot 10^{-5}$ \\
 & 1536 & $7.1 \cdot 10^{-8}$ & $2.5 \cdot 10^{-5}$ & $5.2 \cdot 10^{-8}$ & $1.4 \cdot 10^{-5}$ & $4.2 \cdot 10^{-8}$ & $1.4 \cdot 10^{-5}$ \\
\addlinespace
FM, hold-out & 96 & $2.8 \cdot 10^{-4}$ & $9.9 \cdot 10^{-4}$ & $5.8 \cdot 10^{-4}$ & $1.3 \cdot 10^{-3}$ & $1.4 \cdot 10^{-4}$ & $3.4 \cdot 10^{-4}$ \\
 & 384 & $1.9 \cdot 10^{-4}$ & $2.0 \cdot 10^{-4}$ & $4.3 \cdot 10^{-5}$ & $1.3 \cdot 10^{-4}$ & $2.4 \cdot 10^{-5}$ & $4.7 \cdot 10^{-5}$ \\
 & 1536 & $1.2 \cdot 10^{-5}$ & $1.4 \cdot 10^{-5}$ & $2.9 \cdot 10^{-6}$ & $8.4 \cdot 10^{-6}$ & $8.7 \cdot 10^{-7}$ & $2.5 \cdot 10^{-6}$ \\
\midrule
\multicolumn{8}{l}{\emph{Circle (2D)}} \\
exact $v_N$ & 96 & $2.3 \cdot 10^{-5}$ & $4.1 \cdot 10^{-3}$ & $9.0 \cdot 10^{-2}$ & $1.2 \cdot 10^{-2}$ & $1.4 \cdot 10^{-1}$ & $1.7 \cdot 10^{-2}$ \\
 & 384 & $1.8 \cdot 10^{-7}$ & $5.3 \cdot 10^{-5}$ & $9.1 \cdot 10^{-7}$ & $9.4 \cdot 10^{-5}$ & $1.1 \cdot 10^{-6}$ & $1.8 \cdot 10^{-4}$ \\
 & 1536 & $1.0 \cdot 10^{-7}$ & $2.5 \cdot 10^{-5}$ & $7.3 \cdot 10^{-8}$ & $2.2 \cdot 10^{-5}$ & $1.5 \cdot 10^{-7}$ & $3.0 \cdot 10^{-5}$ \\
\addlinespace
FM, hold-out & 96 & $1.5 \cdot 10^{-2}$ & $1.7 \cdot 10^{-3}$ & $5.3 \cdot 10^{-3}$ & $8.3 \cdot 10^{-4}$ & $6.4 \cdot 10^{-3}$ & $4.5 \cdot 10^{-4}$ \\
 & 384 & $3.7 \cdot 10^{-4}$ & $7.1 \cdot 10^{-5}$ & $1.2 \cdot 10^{-4}$ & $2.5 \cdot 10^{-5}$ & $5.7 \cdot 10^{-5}$ & $1.1 \cdot 10^{-5}$ \\
 & 1536 & $2.3 \cdot 10^{-4}$ & $1.9 \cdot 10^{-5}$ & $7.6 \cdot 10^{-6}$ & $1.8 \cdot 10^{-6}$ & $3.4 \cdot 10^{-6}$ & $8.0 \cdot 10^{-7}$ \\
\midrule
\multicolumn{8}{l}{\emph{Checkerboard (2D)}} \\
exact $v_N$ & 96 & $7.3 \cdot 10^{-6}$ & $8.1 \cdot 10^{-4}$ & $2.6 \cdot 10^{-1}$ & $1.8 \cdot 10^{-3}$ & $4.2 \cdot 10^{-2}$ & $4.8 \cdot 10^{-3}$ \\
 & 384 & $3.8 \cdot 10^{-7}$ & $4.8 \cdot 10^{-5}$ & $2.3 \cdot 10^{-6}$ & $1.4 \cdot 10^{-4}$ & $2.5 \cdot 10^{-6}$ & $1.6 \cdot 10^{-4}$ \\
 & 1536 & $3.4 \cdot 10^{-7}$ & $5.6 \cdot 10^{-5}$ & $2.8 \cdot 10^{-7}$ & $3.8 \cdot 10^{-5}$ & $1.8 \cdot 10^{-7}$ & $5.1 \cdot 10^{-5}$ \\
\addlinespace
FM, hold-out & 96 & $5.0 \cdot 10^{-3}$ & $1.5 \cdot 10^{-3}$ & $3.9 \cdot 10^{-2}$ & $4.5 \cdot 10^{-3}$ & $1.2 \cdot 10^{-2}$ & $1.0 \cdot 10^{-3}$ \\
 & 384 & $1.6 \cdot 10^{-3}$ & $6.9 \cdot 10^{-4}$ & $8.1 \cdot 10^{-3}$ & $4.1 \cdot 10^{-4}$ & $3.1 \cdot 10^{-4}$ & $3.1 \cdot 10^{-5}$ \\
 & 1536 & $5.4 \cdot 10^{-5}$ & $5.1 \cdot 10^{-6}$ & $1.4 \cdot 10^{-4}$ & $1.8 \cdot 10^{-5}$ & $5.4 \cdot 10^{-6}$ & $1.6 \cdot 10^{-6}$ \\
\midrule
\multicolumn{8}{l}{\emph{Two spirals (2D)}} \\
exact $v_N$ & 96 & $2.4 \cdot 10^{-1}$ & $1.1 \cdot 10^{-3}$ & $7.2 \cdot 10^{-6}$ & $9.1 \cdot 10^{-4}$ & $7.3 \cdot 10^{-6}$ & $9.9 \cdot 10^{-4}$ \\
 & 384 & $2.4 \cdot 10^{-1}$ & $8.0 \cdot 10^{-5}$ & $2.9 \cdot 10^{-7}$ & $6.5 \cdot 10^{-5}$ & $3.1 \cdot 10^{-7}$ & $6.0 \cdot 10^{-5}$ \\
 & 1536 & $8.8 \cdot 10^{-7}$ & $3.7 \cdot 10^{-5}$ & $2.5 \cdot 10^{-7}$ & $5.1 \cdot 10^{-5}$ & $2.9 \cdot 10^{-7}$ & $3.9 \cdot 10^{-5}$ \\
\addlinespace
FM, hold-out & 96 & $2.9 \cdot 10^{-3}$ & $2.2 \cdot 10^{-3}$ & $4.0 \cdot 10^{-1}$ & $1.4 \cdot 10^{-2}$ & $1.1 \cdot 10^{-2}$ & $1.1 \cdot 10^{-3}$ \\
 & 384 & $7.9 \cdot 10^{-4}$ & $3.4 \cdot 10^{-4}$ & $4.1 \cdot 10^{-3}$ & $3.8 \cdot 10^{-4}$ & $1.8 \cdot 10^{-4}$ & $5.1 \cdot 10^{-5}$ \\
 & 1536 & $2.5 \cdot 10^{-5}$ & $1.1 \cdot 10^{-5}$ & $7.1 \cdot 10^{-4}$ & $3.4 \cdot 10^{-5}$ & $2.2 \cdot 10^{-5}$ & $5.7 \cdot 10^{-6}$ \\
\midrule
\multicolumn{8}{l}{\emph{8 Gaussians (2D)}} \\
exact $v_N$ & 96 & $1.7 \cdot 10^{-5}$ & $1.0 \cdot 10^{-3}$ & $4.2 \cdot 10^{-2}$ & $3.3 \cdot 10^{-3}$ & $1.3 \cdot 10^{-1}$ & $2.3 \cdot 10^{-3}$ \\
 & 384 & $5.5 \cdot 10^{-7}$ & $6.5 \cdot 10^{-5}$ & $5.6 \cdot 10^{-7}$ & $7.9 \cdot 10^{-5}$ & $2.4 \cdot 10^{-6}$ & $1.2 \cdot 10^{-4}$ \\
 & 1536 & $1.3 \cdot 10^{-6}$ & $7.6 \cdot 10^{-5}$ & $6.0 \cdot 10^{-7}$ & $8.1 \cdot 10^{-5}$ & $2.8 \cdot 10^{-7}$ & $5.1 \cdot 10^{-5}$ \\
\addlinespace
FM, hold-out & 96 & $9.0 \cdot 10^{-3}$ & $3.1 \cdot 10^{-3}$ & $2.1 \cdot 10^{-3}$ & $3.7 \cdot 10^{-4}$ & $2.7 \cdot 10^{-3}$ & $1.2 \cdot 10^{-4}$ \\
 & 384 & $9.2 \cdot 10^{-4}$ & $2.6 \cdot 10^{-4}$ & $2.0 \cdot 10^{-4}$ & $1.6 \cdot 10^{-5}$ & $3.4 \cdot 10^{-5}$ & $8.9 \cdot 10^{-6}$ \\
 & 1536 & $1.8 \cdot 10^{-5}$ & $3.1 \cdot 10^{-6}$ & $1.2 \cdot 10^{-5}$ & $1.9 \cdot 10^{-6}$ & $2.7 \cdot 10^{-6}$ & $2.8 \cdot 10^{-7}$ \\
\bottomrule
\end{tabular}

%% file: tables/tab_stability.tex
\begin{tabular}{clrrrrrrr}
\toprule
 & Estimator & \multicolumn{6}{c}{$\varepsilon$} & slope \\
 & & $10^{-6}$ & $10^{-5}$ & $10^{-4}$ & $10^{-3}$ & $10^{-2}$ & $10^{-1}$ & \\
\midrule
\multicolumn{9}{l}{(a) $\mathrm{TV}(p_\varepsilon, p_0)/\varepsilon$} \\
\multirow{5}{*}{\smash{\rotatebox[origin=c]{90}{\begin{tabular}{@{}c@{}}Trimodal\\$N = 100$\end{tabular}}}} & exact $v_N$ & $1.9 \cdot 10^{4}$ & $1.9 \cdot 10^{4}$ & $9.8 \cdot 10^{3}$ & $1.0 \cdot 10^{3}$ & $9.9 \cdot 10^{1}$ & $1.0 \cdot 10^{1}$ & $0.59$ \\
 & $k = 2$, fixed $\lambda$ & $1.6 \cdot 10^{0}$ & $1.6 \cdot 10^{0}$ & $1.6 \cdot 10^{0}$ & $1.6 \cdot 10^{0}$ & $1.6 \cdot 10^{0}$ & $1.6 \cdot 10^{0}$ & $1.00$ \\
 & $k = 2$, re-selected $\lambda$ & $5.3 \cdot 10^{0}$ & $1.6 \cdot 10^{0}$ & $3.3 \cdot 10^{0}$ & $2.7 \cdot 10^{0}$ & $2.4 \cdot 10^{0}$ & $1.7 \cdot 10^{0}$ & $0.94$ \\
 & $k = 0$, fixed $\lambda$ & $2.9 \cdot 10^{2}$ & $2.9 \cdot 10^{2}$ & $2.9 \cdot 10^{2}$ & $2.6 \cdot 10^{2}$ & $6.8 \cdot 10^{1}$ & $7.2 \cdot 10^{0}$ & $0.98$ \\
 & $k = 0$, re-selected $\lambda$ & $2.9 \cdot 10^{2}$ & $4.8 \cdot 10^{2}$ & $3.5 \cdot 10^{2}$ & $2.8 \cdot 10^{2}$ & $6.7 \cdot 10^{1}$ & $7.3 \cdot 10^{0}$ & $0.98$ \\
\midrule
\multirow{5}{*}{\smash{\rotatebox[origin=c]{90}{\begin{tabular}{@{}c@{}}Inhom.\\$N = 100$\end{tabular}}}} & exact $v_N$ & $1.9 \cdot 10^{4}$ & $1.9 \cdot 10^{4}$ & $9.7 \cdot 10^{3}$ & $1.0 \cdot 10^{3}$ & $9.9 \cdot 10^{1}$ & $1.0 \cdot 10^{1}$ & $0.59$ \\
 & $k = 2$, fixed $\lambda$ & $1.4 \cdot 10^{0}$ & $1.4 \cdot 10^{0}$ & $1.4 \cdot 10^{0}$ & $1.4 \cdot 10^{0}$ & $1.4 \cdot 10^{0}$ & $1.5 \cdot 10^{0}$ & $1.00$ \\
 & $k = 2$, re-selected $\lambda$ & $1.4 \cdot 10^{0}$ & $1.3 \cdot 10^{0}$ & $2.7 \cdot 10^{0}$ & $3.4 \cdot 10^{0}$ & $2.5 \cdot 10^{0}$ & $1.8 \cdot 10^{0}$ & $1.15$ \\
 & $k = 0$, fixed $\lambda$ & $2.9 \cdot 10^{2}$ & $2.9 \cdot 10^{2}$ & $2.9 \cdot 10^{2}$ & $2.6 \cdot 10^{2}$ & $6.2 \cdot 10^{1}$ & $7.2 \cdot 10^{0}$ & $0.98$ \\
 & $k = 0$, re-selected $\lambda$ & $3.0 \cdot 10^{2}$ & $3.0 \cdot 10^{2}$ & $3.1 \cdot 10^{2}$ & $2.7 \cdot 10^{2}$ & $6.2 \cdot 10^{1}$ & $7.4 \cdot 10^{0}$ & $0.99$ \\
\midrule
\multirow{5}{*}{\smash{\rotatebox[origin=c]{90}{\begin{tabular}{@{}c@{}}Circle\\$N = 1000$\end{tabular}}}} & exact $v_N$ & $7.0 \cdot 10^{3}$ & $6.9 \cdot 10^{3}$ & $6.1 \cdot 10^{3}$ & $1.0 \cdot 10^{3}$ & $1.0 \cdot 10^{2}$ & $1.0 \cdot 10^{1}$ & $0.74$ \\
 & $k = 2$, fixed $\lambda$ & $5.5 \cdot 10^{-1}$ & $5.5 \cdot 10^{-1}$ & $5.5 \cdot 10^{-1}$ & $5.5 \cdot 10^{-1}$ & $5.6 \cdot 10^{-1}$ & $1.0 \cdot 10^{0}$ & $1.00$ \\
 & $k = 2$, re-selected $\lambda$ & $7.4 \cdot 10^{-1}$ & $5.9 \cdot 10^{-1}$ & $5.6 \cdot 10^{-1}$ & $5.5 \cdot 10^{-1}$ & $6.5 \cdot 10^{-1}$ & $1.0 \cdot 10^{0}$ & $0.96$ \\
 & $k = 0$, fixed $\lambda$ & $3.7 \cdot 10^{1}$ & $3.7 \cdot 10^{1}$ & $3.7 \cdot 10^{1}$ & $3.7 \cdot 10^{1}$ & $3.1 \cdot 10^{1}$ & $6.8 \cdot 10^{0}$ & $1.00$ \\
 & $k = 0$, re-selected $\lambda$ & $3.7 \cdot 10^{1}$ & $3.7 \cdot 10^{1}$ & $4.0 \cdot 10^{1}$ & $3.8 \cdot 10^{1}$ & $3.1 \cdot 10^{1}$ & $6.8 \cdot 10^{0}$ & $1.00$ \\
\midrule[\heavyrulewidth]
\multicolumn{9}{l}{(b) $\sup_t \|v_t - v_t'\|_{H^2(\Omega)}/\varepsilon$} \\
\multirow{4}{*}{\smash{\rotatebox[origin=c]{90}{\begin{tabular}{@{}c@{}}Trimodal\\$N = 100$\end{tabular}}}} & $k = 2$, fixed $\lambda$ & $4.0 \cdot 10^{2}$ & $4.0 \cdot 10^{2}$ & $4.0 \cdot 10^{2}$ & $4.0 \cdot 10^{2}$ & $3.9 \cdot 10^{2}$ & $2.5 \cdot 10^{2}$ & $1.00$ \\
 & $k = 2$, re-selected $\lambda$ & $4.0 \cdot 10^{2}$ & $4.0 \cdot 10^{2}$ & $1.7 \cdot 10^{4}$ & $4.0 \cdot 10^{3}$ & $3.1 \cdot 10^{3}$ & $6.6 \cdot 10^{2}$ & $1.46$ \\
 & $k = 0$, fixed $\lambda$ & $2.6 \cdot 10^{7}$ & $2.6 \cdot 10^{7}$ & $2.6 \cdot 10^{7}$ & $2.1 \cdot 10^{7}$ & $3.5 \cdot 10^{6}$ & $4.7 \cdot 10^{5}$ & $0.97$ \\
 & $k = 0$, re-selected $\lambda$ & $2.6 \cdot 10^{7}$ & $1.6 \cdot 10^{8}$ & $9.3 \cdot 10^{7}$ & $2.9 \cdot 10^{7}$ & $3.6 \cdot 10^{6}$ & $4.3 \cdot 10^{5}$ & $0.99$ \\
\midrule
\multirow{4}{*}{\smash{\rotatebox[origin=c]{90}{\begin{tabular}{@{}c@{}}Inhom.\\$N = 100$\end{tabular}}}} & $k = 2$, fixed $\lambda$ & $7.6 \cdot 10^{4}$ & $7.6 \cdot 10^{4}$ & $7.6 \cdot 10^{4}$ & $7.3 \cdot 10^{4}$ & $3.3 \cdot 10^{4}$ & $3.3 \cdot 10^{3}$ & $1.00$ \\
 & $k = 2$, re-selected $\lambda$ & $7.6 \cdot 10^{4}$ & $7.6 \cdot 10^{4}$ & $1.2 \cdot 10^{6}$ & $1.7 \cdot 10^{5}$ & $3.1 \cdot 10^{4}$ & $3.3 \cdot 10^{3}$ & $1.22$ \\
 & $k = 0$, fixed $\lambda$ & $2.4 \cdot 10^{7}$ & $2.4 \cdot 10^{7}$ & $2.4 \cdot 10^{7}$ & $2.2 \cdot 10^{7}$ & $3.9 \cdot 10^{6}$ & $5.2 \cdot 10^{5}$ & $0.99$ \\
 & $k = 0$, re-selected $\lambda$ & $2.4 \cdot 10^{7}$ & $4.2 \cdot 10^{8}$ & $1.1 \cdot 10^{8}$ & $3.4 \cdot 10^{7}$ & $3.8 \cdot 10^{6}$ & $4.8 \cdot 10^{5}$ & $0.99$ \\
\midrule
\addlinespace[1pt]
\multirow{4}{*}{\smash{\rotatebox[origin=c]{90}{\begin{tabular}{@{}c@{}}Circle\\$N = 1000$\end{tabular}}}} & $k = 2$, fixed $\lambda$ & $5.8 \cdot 10^{3}$ & $5.8 \cdot 10^{3}$ & $5.8 \cdot 10^{3}$ & $5.8 \cdot 10^{3}$ & $5.8 \cdot 10^{3}$ & $5.4 \cdot 10^{3}$ & $1.00$ \\
 & $k = 2$, re-selected $\lambda$ & $5.8 \cdot 10^{3}$ & $5.8 \cdot 10^{3}$ & $5.8 \cdot 10^{3}$ & $4.6 \cdot 10^{4}$ & $2.0 \cdot 10^{4}$ & $8.7 \cdot 10^{3}$ & $1.27$ \\
 & $k = 0$, fixed $\lambda$ & $6.3 \cdot 10^{5}$ & $6.3 \cdot 10^{5}$ & $6.3 \cdot 10^{5}$ & $6.3 \cdot 10^{5}$ & $5.4 \cdot 10^{5}$ & $1.3 \cdot 10^{5}$ & $1.00$ \\
 & $k = 0$, re-selected $\lambda$ & $6.3 \cdot 10^{5}$ & $6.3 \cdot 10^{5}$ & $3.3 \cdot 10^{6}$ & $1.6 \cdot 10^{6}$ & $5.8 \cdot 10^{5}$ & $1.3 \cdot 10^{5}$ & $1.19$ \\
\addlinespace[1pt]
\bottomrule
\end{tabular}

%% file: tables/tab_stability_refine.tex
\begin{tabular}{clrccc}
\toprule
 & Estimator & $m$ & TV$/\varepsilon$ & $L^2$ field$/\varepsilon$ & $H^2$ field$/\varepsilon$ \\
\midrule
\multirow{16}{*}[-1.5\defaultaddspace]{\smash{\rotatebox[origin=c]{90}{\begin{tabular}{@{}c@{}}Trimodal\\$N = 100$\end{tabular}}}} & $k = 0$ & 256 & $1.6 \cdot 10^{2}$ [$1.3 \cdot 10^{2}$, $2.0 \cdot 10^{2}$] & $2.2 \cdot 10^{3}$ [$7.9 \cdot 10^{2}$, $1.4 \cdot 10^{4}$] & $7.0 \cdot 10^{6}$ [$3.2 \cdot 10^{6}$, $5.1 \cdot 10^{7}$] \\
 & re-selected $\lambda$ & 512 & $3.7 \cdot 10^{2}$ [$1.5 \cdot 10^{2}$, $2.3 \cdot 10^{3}$] & $6.7 \cdot 10^{3}$ [$7.3 \cdot 10^{2}$, $9.7 \cdot 10^{4}$] & $9.2 \cdot 10^{7}$ [$8.6 \cdot 10^{6}$, $1.3 \cdot 10^{9}$] \\
 &  & 1024 & $8.3 \cdot 10^{2}$ [$1.7 \cdot 10^{2}$, $2.9 \cdot 10^{3}$] & $2.4 \cdot 10^{4}$ [$1.5 \cdot 10^{3}$, $1.2 \cdot 10^{5}$] & $7.4 \cdot 10^{8}$ [$9.3 \cdot 10^{7}$, $4.2 \cdot 10^{9}$] \\
 &  & growth & $5.26$ & $11.03$ & $104.71$ \\
\addlinespace
 & $k = 0$ & 256 & $1.3 \cdot 10^{2}$ [$9.1 \cdot 10^{1}$, $1.5 \cdot 10^{2}$] & $1.2 \cdot 10^{3}$ [$4.7 \cdot 10^{2}$, $2.9 \cdot 10^{3}$] & $4.1 \cdot 10^{6}$ [$1.8 \cdot 10^{6}$, $8.7 \cdot 10^{6}$] \\
 & fixed $\lambda$ & 512 & $2.9 \cdot 10^{2}$ [$1.5 \cdot 10^{2}$, $3.9 \cdot 10^{2}$] & $1.6 \cdot 10^{3}$ [$7.1 \cdot 10^{2}$, $2.9 \cdot 10^{3}$] & $2.6 \cdot 10^{7}$ [$8.5 \cdot 10^{6}$, $5.4 \cdot 10^{7}$] \\
 &  & 1024 & $7.9 \cdot 10^{2}$ [$1.7 \cdot 10^{2}$, $1.3 \cdot 10^{3}$] & $4.0 \cdot 10^{3}$ [$7.7 \cdot 10^{2}$, $8.9 \cdot 10^{3}$] & $2.6 \cdot 10^{8}$ [$2.0 \cdot 10^{7}$, $6.8 \cdot 10^{8}$] \\
 &  & growth & $6.21$ & $3.34$ & $64.42$ \\
\addlinespace
 & $k = 2$ & 256 & $2.9 \cdot 10^{0}$ [$2.9 \cdot 10^{-1}$, $1.6 \cdot 10^{1}$] & $6.1 \cdot 10^{3}$ [$3.0 \cdot 10^{1}$, $8.1 \cdot 10^{4}$] & $2.5 \cdot 10^{4}$ [$4.5 \cdot 10^{1}$, $3.1 \cdot 10^{5}$] \\
 & re-selected $\lambda$ & 512 & $3.4 \cdot 10^{0}$ [$2.7 \cdot 10^{-1}$, $2.2 \cdot 10^{1}$] & $8.9 \cdot 10^{2}$ [$2.5 \cdot 10^{1}$, $1.3 \cdot 10^{4}$] & $5.8 \cdot 10^{3}$ [$3.7 \cdot 10^{1}$, $8.2 \cdot 10^{4}$] \\
 &  & 1024 & $3.7 \cdot 10^{0}$ [$2.9 \cdot 10^{-1}$, $5.4 \cdot 10^{1}$] & $3.8 \cdot 10^{3}$ [$2.9 \cdot 10^{1}$, $9.9 \cdot 10^{4}$] & $1.6 \cdot 10^{4}$ [$4.2 \cdot 10^{1}$, $4.0 \cdot 10^{5}$] \\
 &  & growth & $1.28$ & $0.62$ & $0.64$ \\
\addlinespace
 & $k = 2$ & 256 & $1.6 \cdot 10^{0}$ [$2.8 \cdot 10^{-1}$, $3.3 \cdot 10^{0}$] & $6.0 \cdot 10^{1}$ [$2.2 \cdot 10^{1}$, $1.6 \cdot 10^{2}$] & $4.0 \cdot 10^{2}$ [$3.5 \cdot 10^{1}$, $9.3 \cdot 10^{2}$] \\
 & fixed $\lambda$ & 512 & $1.6 \cdot 10^{0}$ [$2.8 \cdot 10^{-1}$, $3.3 \cdot 10^{0}$] & $6.0 \cdot 10^{1}$ [$2.2 \cdot 10^{1}$, $1.6 \cdot 10^{2}$] & $4.0 \cdot 10^{2}$ [$3.5 \cdot 10^{1}$, $9.3 \cdot 10^{2}$] \\
 &  & 1024 & $1.6 \cdot 10^{0}$ [$2.8 \cdot 10^{-1}$, $3.3 \cdot 10^{0}$] & $6.0 \cdot 10^{1}$ [$2.2 \cdot 10^{1}$, $1.6 \cdot 10^{2}$] & $4.0 \cdot 10^{2}$ [$3.5 \cdot 10^{1}$, $9.3 \cdot 10^{2}$] \\
 &  & growth & $1.00$ & $1.00$ & $1.01$ \\
\midrule
\multirow{16}{*}[-1.5\defaultaddspace]{\smash{\rotatebox[origin=c]{90}{\begin{tabular}{@{}c@{}}Inhom.\\$N = 100$\end{tabular}}}} & $k = 0$ & 256 & $9.9 \cdot 10^{1}$ [$8.0 \cdot 10^{1}$, $1.6 \cdot 10^{2}$] & $2.3 \cdot 10^{3}$ [$5.8 \cdot 10^{2}$, $9.2 \cdot 10^{3}$] & $7.0 \cdot 10^{6}$ [$2.3 \cdot 10^{6}$, $2.9 \cdot 10^{7}$] \\
 & re-selected $\lambda$ & 512 & $3.0 \cdot 10^{2}$ [$2.0 \cdot 10^{2}$, $3.8 \cdot 10^{2}$] & $1.1 \cdot 10^{4}$ [$8.9 \cdot 10^{2}$, $2.2 \cdot 10^{5}$] & $1.8 \cdot 10^{8}$ [$1.5 \cdot 10^{7}$, $4.0 \cdot 10^{9}$] \\
 &  & 1024 & $7.9 \cdot 10^{2}$ [$6.1 \cdot 10^{2}$, $1.9 \cdot 10^{3}$] & $1.3 \cdot 10^{4}$ [$1.7 \cdot 10^{3}$, $1.3 \cdot 10^{5}$] & $7.6 \cdot 10^{8}$ [$1.1 \cdot 10^{8}$, $8.8 \cdot 10^{9}$] \\
 &  & growth & $7.92$ & $5.46$ & $109.23$ \\
\addlinespace
 & $k = 0$ & 256 & $1.0 \cdot 10^{2}$ [$7.8 \cdot 10^{1}$, $1.4 \cdot 10^{2}$] & $9.0 \cdot 10^{2}$ [$5.6 \cdot 10^{2}$, $1.4 \cdot 10^{3}$] & $3.6 \cdot 10^{6}$ [$2.4 \cdot 10^{6}$, $5.3 \cdot 10^{6}$] \\
 & fixed $\lambda$ & 512 & $2.9 \cdot 10^{2}$ [$2.0 \cdot 10^{2}$, $3.7 \cdot 10^{2}$] & $1.6 \cdot 10^{3}$ [$9.2 \cdot 10^{2}$, $2.7 \cdot 10^{3}$] & $2.4 \cdot 10^{7}$ [$1.5 \cdot 10^{7}$, $4.7 \cdot 10^{7}$] \\
 &  & 1024 & $9.4 \cdot 10^{2}$ [$6.7 \cdot 10^{2}$, $1.6 \cdot 10^{3}$] & $4.0 \cdot 10^{3}$ [$2.1 \cdot 10^{3}$, $9.7 \cdot 10^{3}$] & $2.8 \cdot 10^{8}$ [$1.5 \cdot 10^{8}$, $6.9 \cdot 10^{8}$] \\
 &  & growth & $9.30$ & $4.44$ & $78.92$ \\
\addlinespace
 & $k = 2$ & 256 & $1.5 \cdot 10^{0}$ [$3.7 \cdot 10^{-1}$, $5.5 \cdot 10^{0}$] & $4.6 \cdot 10^{2}$ [$1.4 \cdot 10^{1}$, $5.3 \cdot 10^{3}$] & $1.1 \cdot 10^{4}$ [$8.6 \cdot 10^{1}$, $5.0 \cdot 10^{4}$] \\
 & re-selected $\lambda$ & 512 & $1.8 \cdot 10^{0}$ [$3.5 \cdot 10^{-1}$, $1.2 \cdot 10^{1}$] & $4.9 \cdot 10^{2}$ [$1.6 \cdot 10^{1}$, $5.9 \cdot 10^{3}$] & $4.6 \cdot 10^{5}$ [$1.1 \cdot 10^{2}$, $7.4 \cdot 10^{6}$] \\
 &  & 1024 & $1.3 \cdot 10^{0}$ [$3.8 \cdot 10^{-1}$, $3.8 \cdot 10^{0}$] & $1.3 \cdot 10^{2}$ [$2.0 \cdot 10^{1}$, $2.8 \cdot 10^{2}$] & $3.2 \cdot 10^{5}$ [$9.4 \cdot 10^{1}$, $1.6 \cdot 10^{6}$] \\
 &  & growth & $0.90$ & $0.29$ & $29.75$ \\
\addlinespace
 & $k = 2$ & 256 & $1.4 \cdot 10^{0}$ [$3.5 \cdot 10^{-1}$, $3.8 \cdot 10^{0}$] & $1.3 \cdot 10^{2}$ [$1.9 \cdot 10^{1}$, $2.9 \cdot 10^{2}$] & $1.9 \cdot 10^{4}$ [$7.2 \cdot 10^{1}$, $9.6 \cdot 10^{4}$] \\
 & fixed $\lambda$ & 512 & $1.4 \cdot 10^{0}$ [$3.5 \cdot 10^{-1}$, $3.8 \cdot 10^{0}$] & $1.3 \cdot 10^{2}$ [$1.9 \cdot 10^{1}$, $2.9 \cdot 10^{2}$] & $7.6 \cdot 10^{4}$ [$7.2 \cdot 10^{1}$, $3.8 \cdot 10^{5}$] \\
 &  & 1024 & $1.4 \cdot 10^{0}$ [$3.5 \cdot 10^{-1}$, $3.8 \cdot 10^{0}$] & $1.3 \cdot 10^{2}$ [$1.9 \cdot 10^{1}$, $2.9 \cdot 10^{2}$] & $1.2 \cdot 10^{5}$ [$7.2 \cdot 10^{1}$, $5.9 \cdot 10^{5}$] \\
 &  & growth & $1.00$ & $1.00$ & $6.05$ \\
\bottomrule
\end{tabular}

%% file: tables/tab_reg_full.tex
\begin{tabular}{clccc@{\hspace{1.5em}}clccc}
\toprule
$k$ & $\lambda(t)$ & $N = 100$ & $N = 1000$ & $N = 3000$ & $k$ & $\lambda(t)$ & $N = 100$ & $N = 1000$ & $N = 3000$ \\
 &  & ($10^{-2}$) & ($10^{-2}$) & ($10^{-2}$) &  &  & ($10^{-2}$) & ($10^{-2}$) & ($10^{-2}$) \\
\midrule
\multicolumn{5}{l}{\emph{Trimodal (1D)}} & \multicolumn{5}{l}{\emph{Checkerboard (2D)}} \\
$2$ & oracle & $10.0 \pm 0.8$ & $4.1 \pm 0.3$ & $2.5 \pm 0.2$ & $2$ & oracle & $36.7 \pm 0.4$ & $20.6 \pm 0.4$ & $16.1 \pm 0.2$ \\
 & hold-out & $13.4 \pm 1.7$ & $4.7 \pm 0.3$ & $2.9 \pm 0.4$ &  & hold-out & $37.3 \pm 0.6$ & $20.4 \pm 0.3$ & $16.2 \pm 0.2$ \\
\addlinespace
$1$ & oracle & $11.3 \pm 0.7$ & $4.7 \pm 0.3$ & $2.9 \pm 0.1$ & $1$ & oracle & $38.8 \pm 0.5$ & $23.0 \pm 0.2$ & $18.2 \pm 0.2$ \\
 & hold-out & $15.1 \pm 1.9$ & $5.0 \pm 0.2$ & $3.6 \pm 0.3$ &  & hold-out & $38.6 \pm 0.6$ & $22.5 \pm 0.5$ & $18.0 \pm 0.4$ \\
\addlinespace
$0$ & oracle & $60.1 \pm 0.8$ & $24.4 \pm 0.1$ & $13.9 \pm 0.2$ & $0$ & oracle & $93.2 \pm 0.3$ & $77.9 \pm 0.4$ & $62.5 \pm 0.4$ \\
 & hold-out & $57.5 \pm 2.7$ & $24.1 \pm 0.2$ & $13.9 \pm 0.2$ &  & hold-out & $90.7 \pm 0.9$ & $77.9 \pm 1.0$ & $61.2 \pm 0.4$ \\
\midrule
\multicolumn{5}{l}{\emph{Inhomogeneous (1D)}} & \multicolumn{5}{l}{\emph{Two spirals (2D)}} \\
$2$ & oracle & $14.6 \pm 0.7$ & $5.9 \pm 0.3$ & $4.0 \pm 0.1$ & $2$ & oracle & $50.9 \pm 3.0$ & $20.9 \pm 0.3$ & $13.9 \pm 0.3$ \\
 & hold-out & $16.9 \pm 1.5$ & $6.7 \pm 0.3$ & $4.3 \pm 0.1$ &  & hold-out & $57.2 \pm 5.1$ & $20.7 \pm 0.3$ & $13.6 \pm 0.3$ \\
\addlinespace
$1$ & oracle & $13.9 \pm 1.0$ & $6.7 \pm 0.3$ & $4.5 \pm 0.1$ & $1$ & oracle & $54.8 \pm 1.2$ & $27.1 \pm 0.3$ & $18.9 \pm 0.2$ \\
 & hold-out & $15.9 \pm 1.3$ & $7.5 \pm 0.4$ & $4.6 \pm 0.1$ &  & hold-out & $57.5 \pm 0.9$ & $26.3 \pm 0.2$ & $18.6 \pm 0.2$ \\
\addlinespace
$0$ & oracle & $51.6 \pm 0.8$ & $23.1 \pm 0.4$ & $13.1 \pm 0.1$ & $0$ & oracle & $87.9 \pm 0.6$ & $61.1 \pm 0.5$ & $40.8 \pm 0.3$ \\
 & hold-out & $51.6 \pm 1.5$ & $23.0 \pm 0.4$ & $13.0 \pm 0.2$ &  & hold-out & $86.8 \pm 1.2$ & $60.2 \pm 0.8$ & $40.5 \pm 0.3$ \\
\midrule
\multicolumn{5}{l}{\emph{Noisy circle (2D)}} & \multicolumn{5}{l}{\emph{8 Gaussians (2D)}} \\
$2$ & oracle & $18.2 \pm 0.8$ & $8.5 \pm 0.3$ & $5.5 \pm 0.3$ & $2$ & oracle & $25.6 \pm 1.2$ & $11.7 \pm 0.2$ & $8.4 \pm 0.2$ \\
 & hold-out & $18.9 \pm 0.5$ & $8.4 \pm 0.5$ & $5.4 \pm 0.3$ &  & hold-out & $26.7 \pm 1.2$ & $12.4 \pm 0.5$ & $8.4 \pm 0.2$ \\
\addlinespace
$1$ & oracle & $27.7 \pm 1.1$ & $14.6 \pm 0.3$ & $9.9 \pm 0.2$ & $1$ & oracle & $31.2 \pm 1.3$ & $15.2 \pm 0.3$ & $10.8 \pm 0.1$ \\
 & hold-out & $27.3 \pm 0.5$ & $14.3 \pm 0.3$ & $9.4 \pm 0.3$ &  & hold-out & $30.9 \pm 1.5$ & $14.8 \pm 0.3$ & $10.5 \pm 0.3$ \\
\addlinespace
$0$ & oracle & $85.0 \pm 0.2$ & $53.7 \pm 0.3$ & $33.8 \pm 0.2$ & $0$ & oracle & $85.7 \pm 0.3$ & $64.6 \pm 0.4$ & $46.3 \pm 0.3$ \\
 & hold-out & $85.1 \pm 0.9$ & $52.3 \pm 0.1$ & $33.5 \pm 0.2$ &  & hold-out & $83.5 \pm 1.0$ & $62.0 \pm 0.8$ & $45.4 \pm 0.4$ \\
\bottomrule
\end{tabular}

%% file: tables/tab_lam_sweep_full.tex
\begin{tabular}{lccccc}
\toprule
Regularizer & hold-out $\lambda(t)$ & best scale $\alpha$ & TV at best $\alpha$ & best constant & hold-out constant \\
 & ($10^{-2}$) &  & ($10^{-2}$) & ($10^{-2}$) & ($10^{-2}$) \\
\midrule
\multicolumn{6}{l}{\emph{Trimodal (1D)}} \\
Hessian ($k = 2$) & $4.7 \pm 0.3$ & $10^{-0.5}$ & $4.5 \pm 0.3$ & $4.6 \pm 0.2$ & $4.9 \pm 0.2$ \\
gradient ($k = 1$) & $5.0 \pm 0.2$ & $1$ & $5.0 \pm 0.2$ & $5.3 \pm 0.2$ & $5.4 \pm 0.2$ \\
smallness ($k = 0$) & $24.1 \pm 0.2$ & $10^{0.5}$ & $23.7 \pm 0.1$ & $22.2 \pm 0.1$ & $24.1 \pm 0.1$ \\
\midrule
\multicolumn{6}{l}{\emph{Inhomogeneous (1D)}} \\
Hessian ($k = 2$) & $6.7 \pm 0.3$ & $10^{-0.5}$ & $6.5 \pm 0.4$ & $5.9 \pm 0.3$ & $6.5 \pm 0.4$ \\
gradient ($k = 1$) & $7.5 \pm 0.4$ & $1$ & $7.5 \pm 0.4$ & $6.6 \pm 0.4$ & $6.6 \pm 0.4$ \\
smallness ($k = 0$) & $23.0 \pm 0.4$ & $10^{0.5}$ & $22.8 \pm 0.4$ & $19.8 \pm 0.3$ & $22.8 \pm 0.5$ \\
\midrule
\multicolumn{6}{l}{\emph{Noisy circle (2D)}} \\
Hessian ($k = 2$) & $8.4 \pm 0.5$ & $1$ & $8.4 \pm 0.5$ & $10.5 \pm 0.4$ & $11.0 \pm 0.4$ \\
gradient ($k = 1$) & $14.3 \pm 0.3$ & $10^{0.5}$ & $11.7 \pm 0.3$ & $18.1 \pm 0.2$ & $18.1 \pm 0.2$ \\
smallness ($k = 0$) & $52.3 \pm 0.1$ & $10^{2}$ & $39.5 \pm 0.3$ & $55.6 \pm 0.2$ & $55.6 \pm 0.2$ \\
\midrule
\multicolumn{6}{l}{\emph{Checkerboard (2D)}} \\
Hessian ($k = 2$) & $20.4 \pm 0.3$ & $1$ & $20.4 \pm 0.3$ & $21.9 \pm 0.3$ & $22.0 \pm 0.3$ \\
gradient ($k = 1$) & $22.5 \pm 0.5$ & $10^{0.5}$ & $21.1 \pm 0.3$ & $25.9 \pm 0.2$ & $28.7 \pm 0.2$ \\
smallness ($k = 0$) & $77.9 \pm 1.0$ & $10^{2.5}$ & $43.7 \pm 1.5$ & $62.0 \pm 0.3$ & $77.9 \pm 0.2$ \\
\midrule
\multicolumn{6}{l}{\emph{Two spirals (2D)}} \\
Hessian ($k = 2$) & $20.7 \pm 0.3$ & $1$ & $20.7 \pm 0.3$ & $23.9 \pm 0.1$ & $23.9 \pm 0.1$ \\
gradient ($k = 1$) & $26.3 \pm 0.2$ & $10^{0.5}$ & $26.1 \pm 0.1$ & $32.6 \pm 0.2$ & $32.6 \pm 0.2$ \\
smallness ($k = 0$) & $60.2 \pm 0.8$ & $10^{1}$ & $52.7 \pm 0.3$ & $65.9 \pm 0.4$ & $66.2 \pm 0.4$ \\
\midrule
\multicolumn{6}{l}{\emph{8 Gaussians (2D)}} \\
Hessian ($k = 2$) & $12.4 \pm 0.5$ & $1$ & $12.4 \pm 0.5$ & $14.7 \pm 0.1$ & $15.3 \pm 0.7$ \\
gradient ($k = 1$) & $14.8 \pm 0.3$ & $10^{0.5}$ & $13.8 \pm 0.6$ & $18.8 \pm 0.1$ & $19.9 \pm 0.7$ \\
smallness ($k = 0$) & $62.0 \pm 0.8$ & $10^{2}$ & $41.8 \pm 2.4$ & $60.2 \pm 0.2$ & $61.2 \pm 1.1$ \\
\bottomrule
\end{tabular}